\documentclass[lettersize,journal]{IEEEtran}
\usepackage{amsmath,amsfonts}
\usepackage{algorithmic}
\usepackage{algorithm}
\usepackage{amssymb}
\usepackage{array}
\usepackage[caption=false,font=normalsize,labelfont=sf,textfont=sf]{subfig}
\usepackage{textcomp}
\usepackage{stfloats}
\usepackage{url}
\usepackage{verbatim}
\usepackage{graphicx}
\usepackage{cite}
\usepackage{multicol}
\usepackage{multirow}
\usepackage{tabularx}
\usepackage{booktabs}
\usepackage{tikz}
\usepackage{soul, color, xcolor}
\usetikzlibrary{snakes}
\usepackage{rotating}
\usetikzlibrary{fit, calc, decorations.markings}
\usepackage[edges]{forest}
\usepackage{xcolor}
\usepackage{color}
\usepackage{rotating}
\usepackage{tabularray}
\usepackage[colorlinks,linkcolor=blue]{hyperref}
\definecolor{hidden-draw}{RGB}{255,255,255}
\definecolor{mygray}{RGB}{200,200,200}
\definecolor{myred}{RGB}{255,191,182}
\definecolor{myblue}{RGB}{218,233,255}
\definecolor{myyellow}{RGB}{254,201,148}
\definecolor{bluelight2}{RGB}{220,220,255}
\definecolor{greenlight}{RGB}{220,255,220}
\usepackage[numbers]{natbib}
\newcommand{\etal}{\textit{et al.}}
\usepackage{CJKutf8}

\begin{document}

% \title{Generative Models for Robotic Manipulation: \\ A Survey}
\title{Generative Artificial Intelligence in Robotic Manipulation: A Survey}
\author{Kun Zhang$^{*}$, Peng Yun$^{*}$, Jun Cen, Junhao Cai, Didi Zhu, Hangjie Yuan, Chao Zhao, Tao Feng, \\ Michael Yu Wang, Qifeng Chen, Jia Pan, Wei Zhang, Bo Yang$^{\dagger}$, Hua Chen$^{\dagger}$ % <-this % stops a space
% \thanks{*This work was not supported by any organization}% <-this % stops a space
\thanks{$*$ \text { denotes equal contribution, } $\dagger$ \text {denotes corresponding author. }}

\thanks{Kun Zhang, Jia Pan, Wei Zhang, and Hua Chen are with LimX Dynamics, Shenzhen, China. Jia Pan is also with the Department of Computer Science, University of Hong Kong, Hong Kong, China. Wei Zhang is also with the School of System Design and Intelligent Manufacturing (SDIM), Southern University of Science and Technology, Shenzhen, China. Hua Chen is also with the ZJU-UIUC Institute, Zhejiang University, Haining, Zhejiang, 314400, China.  Emails: {\tt kun.zhang@connect.ust.hk, jpan@cs.hku.hk, zhangw3@sustech.edu.cn, huachen@intl.zju.edu.cn}}%
\thanks{Peng YUN and Bo YANG are with vLAR Group, The Hong Kong Polytechnic University, Hong Kong SAR, China. Emails: {\tt \{peng-aae.yun;bo.yang\}@polyu.edu.hk}}%
\thanks{Jun Cen, Junhao Cai, Chao Zhao, and Qifeng Chen are with The Hong Kong University of Science and Technology, Hong Kong, China. Emails: {\tt \{jcenaa;jcaiaq;czhaobb\}@connect.ust.hk, cqf@ust.hk}}% Email: {\tt jcenaa@connect.ust.hk}}%
\thanks{Didi Zhu, and Hangjie Yuan are with Zhejiang University, Zhejiang, China. Emails: {\tt \{didi\_zhu;hj.yuan\}@zju.edu.cn}}%

\thanks{Tao Feng is with the Department of Computer Science and Technology, Tsinghua University, Beijing, China. Emails: {\tt fengtao.hi@gmail.com}}%
\thanks{Michael Yu Wang is with the School of Engineering, Great Bay University, Songshan Lake, Dongguan, Guangdong, China. Email: {\tt mywang@gbu.edu.cn}}%
% \thanks{Jia Pan is with the Department of Computer Science, University of Hong Kong, Hong Kong, China. Email: {\tt jpan@cs.hku.hk}}%
% \thanks{Hua Chen is with ZJU-UIUC Institute, Zhejiang University, Haining, Zhejiang, 314400, China, and also with LimX Dynamics, shenzhen, China. Email: {\tt huachen@intl.zju.edu.cn}}%
}


\maketitle

\begin{abstract}
This survey provides a comprehensive review on recent advancements of generative learning models in robotic manipulation, addressing key challenges in the field. Robotic manipulation faces critical bottlenecks, including significant challenges in insufficient data and inefficient data acquisition, long-horizon and complex task planning, and the multi-modality reasoning ability for robust policy learning performance across diverse environments. To tackle these challenges, this survey introduces several generative model paradigms, including Generative Adversarial Networks (GANs), Variational Autoencoders (VAEs), diffusion models, probabilistic flow models, and autoregressive models, highlighting their strengths and limitations. The applications of these models are categorized into three hierarchical layers: the Foundation Layer, focusing on data generation and reward generation; the Intermediate Layer, covering language, code, visual, and state generation; and the Policy Layer, emphasizing grasp generation and trajectory generation. Each layer is explored in detail, along with notable works that have advanced the state of the art. Finally, the survey outlines future research directions and challenges, emphasizing the need for improved efficiency in data utilization, better handling of long-horizon tasks, and enhanced generalization across diverse robotic scenarios. All
the related resources, including research papers,
open-source data, and projects, are collected for the community
in \href{https://github.com/GAI4Manipulation/AwesomeGAIManipulation}{https://github.com/GAI4Manipulation/AwesomeGAIManipulation}
\end{abstract}

% \begin{IEEEkeywords}
% Article submission, IEEE, IEEEtran, journal, \LaTeX, paper, template, typesetting.
% \end{IEEEkeywords}
\begin{figure*}
  \includegraphics[width=\textwidth]{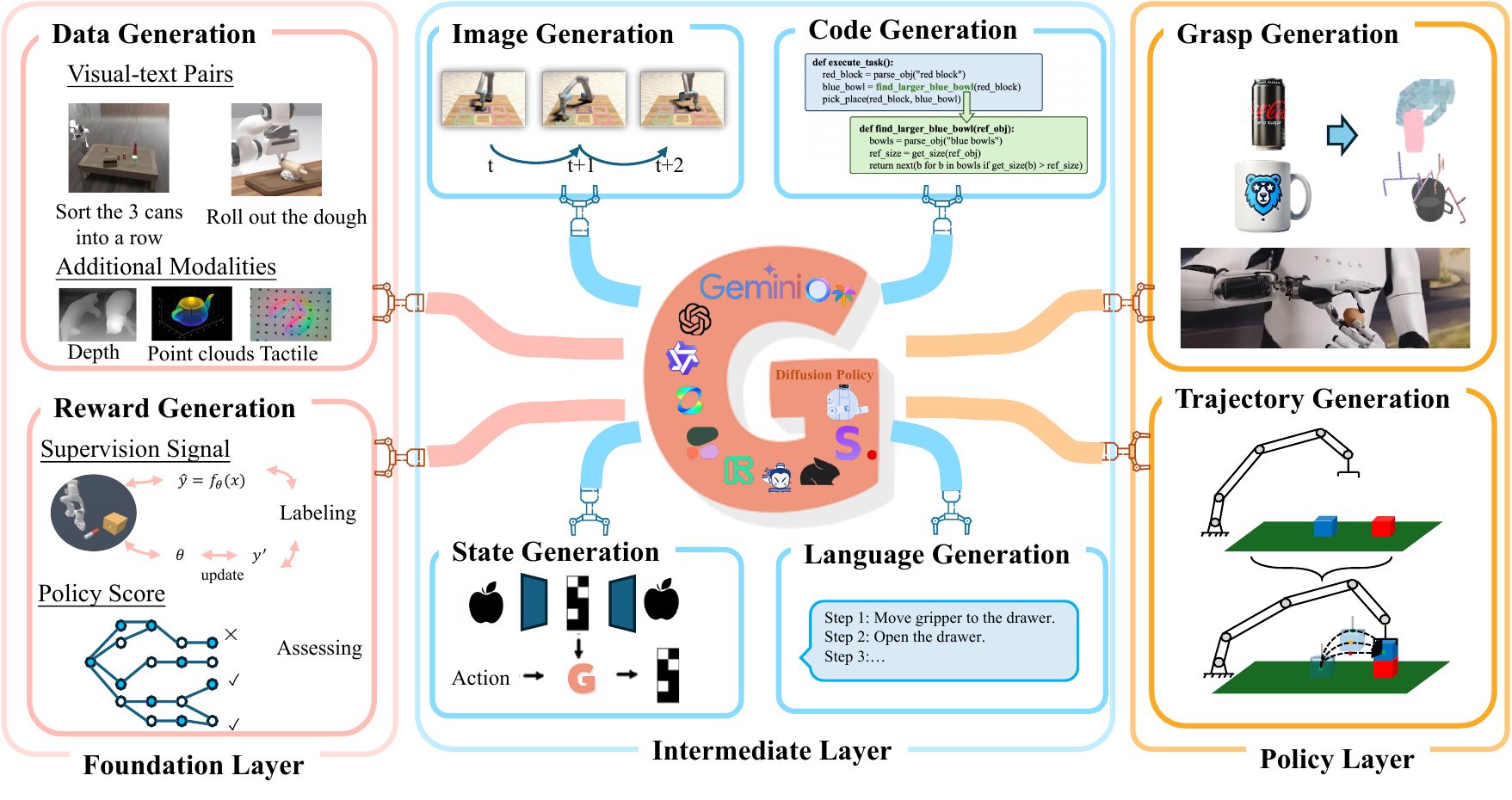}
  \caption{Overview of this survey. Versatile generative models in robotic manipulation.}
  \label{figure:overview}
\end{figure*}

\section{Introduction}
\label{section:introduction}

Robotic manipulation is pivotal in empowering machines to physically interact with and modify their surrounding environment, which is a fundamental step in achieving intelligent autonomy. Ranging from assembling delicate electronic devices in factories to performing assistive care in homes, robotic manipulation plays a crucial role in applications that significantly impact our society \cite{Billard2019, Mason2018}. Manipulation, as one of the most important problems in robotics, has long faced significant challenges in complex environments, especially in scenarios involving nontrivial interactions and intricate long-horizon decision making and planning~\cite{Billard2019, firoozi2023foundation}. Such challenges prevent robotic systems from performing reliable and robust manipulation tasks across different scenarios, leaving a huge gap. 

In recent years, there has been a growing emphasis on data-driven approaches in robotic manipulation, which leverage large-scale data and machine learning techniques to enable robots to better perceive, adapt to, and interact with diverse environments. Thanks to these explosive advances, the aforementioned gap has been drastically closed. In particular, through exploiting the extraordinary abilities of generative learning models in scene understanding, reasoning, task planning and policy synthesis, manipulation skills including operating deformable materials and executing long-horizon task sequences that are widely considered formidably difficult before have been demonstrated.

Generative learning models, as one of the most important classes of learning models in modern AI, have addressed several previously unresolved challenges in robotic manipulation, particularly in grasping tasks. First, their ability to synthesize diverse and high-quality data has significantly alleviated the reliance on extensive real-world datasets. By generating synthetic grasping scenarios and object variations, these models enable robots to train efficiently and handle a wider range of objects, even in data-scarce environments~\cite{Tobin2017DomainRandomization,wang2023robogen}. Second, their capacity to model high-dimensional action and object spaces allows robots to predict feasible grasp configurations and trajectories for complex or unseen objects~\cite{chi2023diffusion,3d_diffuser_actor,Sundermeyer2021ContactGraspNet}. This improves the robot's ability to adapt to novel tasks and environments, enhancing the robustness of grasp planning. Third, their strength in learning latent representations that capture object structure and interaction dynamics enables robots to generalize across diverse shapes, textures, and physical properties~\cite{hafnerdream,bauer_doughnet_nodate}. This ensures more reliable performance in tasks requiring precise manipulation, even in unstructured or dynamic settings. These breakthroughs highlight the transformative potential of generative models in advancing robotic grasping and manipulation.

In this survey, we focus on generative models due to their potential to address long-standing challenges in manipulation. Generative models offer promising solutions, such as improving scene understanding, reasoning, and task planning, which effectively mitigate these issues. In the following paragraphs, we list the key challenges in manipulation and discuss the potential mechanisms through which generative models can overcome these obstacles.

\subsection{Major Challenges in Modern Manipulation}
% Key Challenges
First, \textbf{insufficient data and inefficient data acquisition remain critical bottlenecks}. Data-driven methods have been gradually becoming one of the dominant methodologies in solving manipulation problems. The data-driven methods, such as Reinforcement Learning (RL) and Imitation Learning (IL), are well known to be data hungry, requiring vast amounts of high-quality data to train effective models \cite{Levine2016EndToEnd, Zhu2018Reinforcement}. Collecting high-quality data often necessitates either human intervention or extensive real-world robot experiments, which are time-consuming and difficult to scale to a large amount\cite{Pinto2016Supersizing}.  To simplify the data generation problem, some researchers have explored transfer learning from other tasks or domains \cite{james2019sim, Rusu2017SimToReal, Zhu2017Target}, and techniques like domain randomization to mitigate data scarcity \cite{Tobin2017DomainRandomization}. However, the reliance on high-quality, task-specific data continues to hinder performance and scalability. Addressing these issues is essential for unlocking the full potential of data-driven robotic manipulation.

% Generative models can take effects through mechanisms such as data generation, reward generation, task planning, and policy modeling to overcome these obstacles.
% Data is the foundation in artificial intelligence.
Generative models like Stable Diffusion \cite{Rombach2022HighResolution} and large-scale pre-trained language models \cite{brown2020language} have demonstrated remarkable capabilities in producing high-quality synthetic images, videos, annotations, and reward signals. These models enable the creation of abundant and diverse datasets, significantly alleviating the data insufficiency problem by providing scalable and efficient data generation pipelines. The synthetic data can be utilized to train and validate robotic manipulation models, enhancing their performance and generalization abilities. Additionally, the ability to generate rich reward functions facilitates more effective reinforcement learning by offering detailed feedback and enabling exploration in complex environments. This focus on \underline{data and reward generation} lays the foundation for overcoming data scarcity and inefficient data acquisition, thus propelling the field of robotic manipulation forward.

Second, \textbf{long-horizon task and complex task planning pose significant challenges}. Complex tasks, such as multi-step assembly operations, object rearrangement in cluttered environments, and collaborative tasks with humans \cite{Nikolaidis2015Efficient}, require robots to plan and execute extended sequences of interdependent actions. Effective planning demands sophisticated modeling techniques and often assumes full observability of the environment \cite{Kaelbling2011Hierarchical}. However, in real-world scenarios, full observation is rarely feasible, necessitating that agents develop an intrinsic understanding of tasks, including causal relationships and the effects of their actions on the environment \cite{Pearl2009Causality, hafnerdream}. Traditional deterministic models struggle to capture this complexity due to their inability to adequately represent the uncertainties and dynamic interactions inherent in long-horizon tasks \cite{Toussaint2015Logic}.

%% Task Layer, language generation, code generation
% Hierarchical planning strategies facilitate completing complex tasks.
Generative models contribute significantly to addressing long-horizon task planning by enabling the decomposition of complex tasks into manageable sub-goals through techniques like Chain-of-Thought reasoning~\cite{wei2022chain}. Leveraging the capabilities of \underline{language generation and code generation}, large-scale generative models assist robots in planning intricate sequences of actions by breaking them down into simpler, sequential steps \cite{Ahn2022DoAsICan, huang2022language}. This approach allows agents to generate explicit chains of thought and action plans, enhancing their understanding and execution of complex tasks. By incorporating these generative techniques, robots can better handle the uncertainties and dynamic interactions inherent in long-horizon tasks, thereby improving their overall performance in manipulation scenarios.

%% Task Layer, visual generation, state generation
Furthermore, generative models enhance robots' understanding of the physical world by enabling the development of world models and facilitating dynamics learning. By generating intermediate states-explicitly as \underline{visual representations}, like consequence images \cite{Finn2016Unsupervised, Ebert2018VisualForesight}, or implicitly through \underline{latent states}~\cite{Ha2018WorldModels}, these models allow robots to predict and plan for future events in their environment. The capability of visual generation of potential future states improves planning and decision-making processes in manipulation tasks. State generation captures the underlying dynamics essential for accurate task execution, addressing uncertainties and variability in complex environments. This empowers robots to anticipate and adapt to changes during manipulation tasks, enhancing their performance in dynamic settings.

Third, \textbf{policy learning requires multi-modality reasoning abilities}. In robotic manipulation, the current state can correspond to multiple valid actions and outcomes due to task complexity and environmental variability. For example, a mug can be grasped either by the handle or the body, with the optimal choice depending on the subsequent task: grasping the handle is preferable for filling it with water, while grasping the body is better for handing it over to others. Deterministic models often map input observations to single outputs, failing to capture the inherent multi-modality present in many manipulation tasks. This limitation restricts adaptability and hampers performance in diverse situations. By relying on one-to-one mappings, these models struggle to represent the full spectrum of possible actions, thereby hindering the development of more flexible and generalizable robotic systems.

%% Policy Layer, grasp generation, policy generation
Generative models have shown significant potential in policy learning, particularly in \underline{grasp generation and trajectory generation} for robotic manipulation tasks~\cite{urain2023se,singh2024constrained,chi2023diffusion,fu2024mobile}. By modeling the action sequences of entire trajectories, generative models enable joint optimization of control policies. For instance, diffusion models have been applied to policy learning, allowing for the generation of smooth and feasible motion trajectories \cite{urain2023se}. These models can incorporate constraints inherent in the robot's operational space, such as SE(3) constraints for generating valid grasp poses in three-dimensional space \cite{Sundermeyer2021ContactGraspNet}. This capability enhances the robot's ability to perform precise and complex manipulation tasks by generating policies that are both efficient and physically plausible.Moreover, their ability to model multi-modal distributions allows them to capture the diverse possible grasp poses and motion trajectories essential for complex manipulation tasks.

%% 承上 (Generative Model) 启下 (Related Work)
\subsection{Structural Overview of the Survey}
In summary, generative models offer solutions across multiple layers of robotic manipulation: from foundational data and reward generation to advanced task planning and policy modeling. By addressing the key challenges of data insufficiency, complex task planning, low-level control, and representation learning, generative models pave the way for more autonomous, efficient, and capable robotic systems. 
Several surveys have explored topics related to robotics and generative models \cite{hu_toward_2024, firoozi2023foundation, mccarthy_towards_2024}. These works have examined foundational models in robotics and the progression towards general artificial intelligence. However, none have specifically concentrated on how generative models can address the key challenges in robotic manipulation. Our survey focuses on the application of generative models in manipulation tasks, trying to provide a unified yet specific perspective on the roles of generative models in robotic manipulation at different layers. In our survey, by emphasizing the benefits that generative models bring to these specific areas, we aim to fill the existing gap in the literature.
Figure~\ref{figure:overview} shows the overall structure of our surveyed approaches.

To systematically understand the role of generative models in robotic manipulation, we categorize their applications into three hierarchical layers: \textbf{Foundation Layer}, \textbf{Intermediate Layer}, and \textbf{Policy Layer}. This structure reflects the progressive flow from fundamental data synthesis to high-level decision-making to low-level control. The foundation layer focuses on generating essential resources, such as synthetic data to augment limited datasets and reward signals to guide reinforcement learning, forming the backbone for model training and evaluation. Building on this, the intermediate layer encompasses tasks like language, code, visual, and state generation, which enable robots to interpret instructions, process sensory data, and reason about their environment, bridging perception and action. Finally, the policy layer directly addresses the core robotic manipulation problem, including grasp generation and trajectory planning, translating insights from the lower layers into actionable control strategies. This layered framework highlights the interdependence of these components, ensuring a comprehensive and scalable approach to robotic learning and control.

\section{Background in Generative Learning Models}
\label{section:background}
Generative models have emerged as a class of powerful tools in machine learning. Focusing on the problem of synthesizing complex data distributions, generative models offer core advantages such as capturing underlying patterns, generating diverse outputs, and enabling adaptive solutions across tasks. In the field of robotic manipulation, these models have been adopted to address various crucial challenges, such as generating realistic data, planning precise trajectories, and adapting to dynamic and unstructured environments. By leveraging their ability to model uncertainty and handle high-dimensional data, generative models enhance robots’ capacity for generalization, decision-making, and task execution. In this section, we provide an overview of several foundational generative model paradigms, highlighting their principles and significance in robotic manipulation.
% \begin{figure}
%     \centering
%     \includegraphics[width=\linewidth]{figures/generative_models.png}
%     \caption{Overview of generative models. \hl{We have to change this figure. This figure is just a placeholder copy from another paper.}}
%     \label{fig:generative_models}
% \end{figure}
\begin{figure*}
  \includegraphics[width=\textwidth]{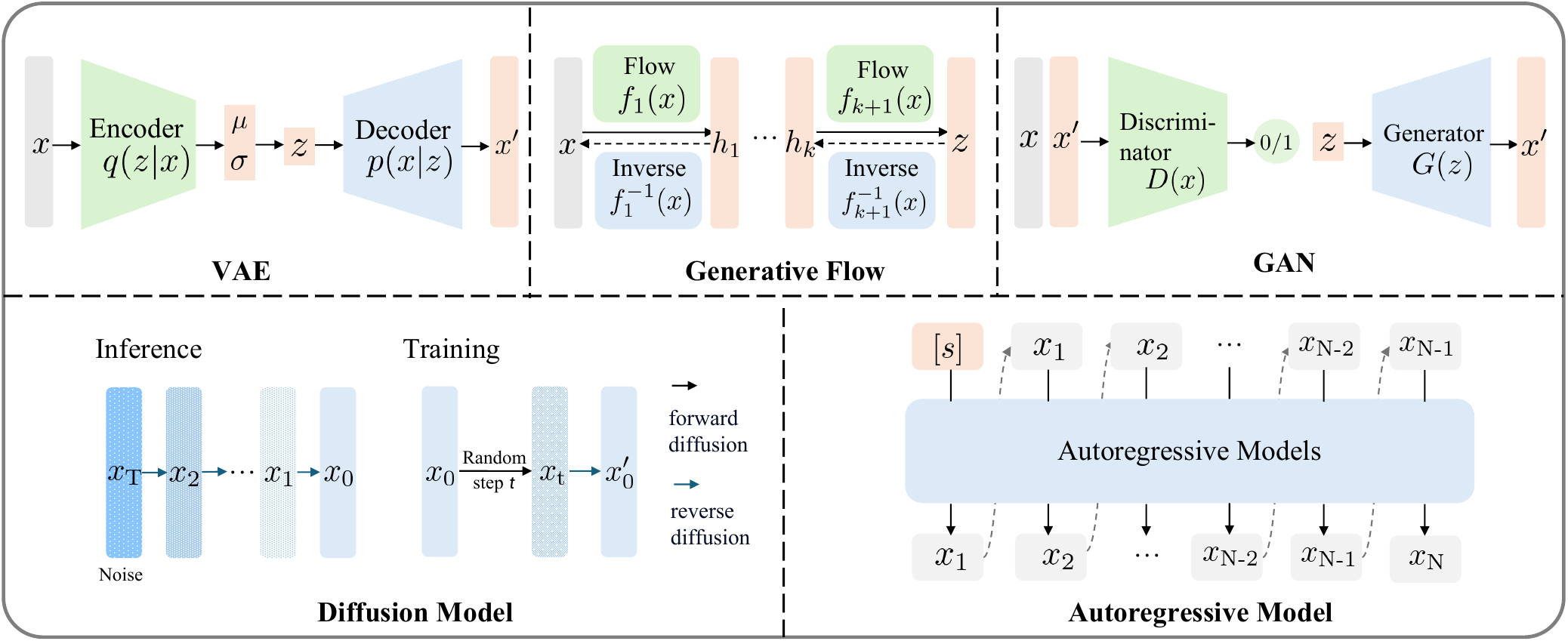}
  \caption{Overview of generative models.}
  \label{figure:generative_models}
\end{figure*}

\subsection{Generative Adversarial Network (GAN)}
Generative Adversarial Networks (GANs) consist of two components: a generator and a discriminator. The generator aims to create realistic data samples from random noise, while the discriminator attempts to differentiate between real and generated samples. The two networks are trained simultaneously in a zero-sum game, where the generator tries to fool the discriminator and the discriminator improves at distinguishing real data from fake ones. This adversarial training process leads to high-quality generated data over time \cite{goodfellow2014generative}.

Mathematically, the objective function for GANs can be written as follows:
\begin{align}
\min_G \max_D V(D, G) = &\ 
\mathbb{E}_{x \sim p_{\text{data}}(x)}[\log D(x)] \notag \\
 & + \mathbb{E}_{z \sim p_z(z)}[\log(1 - D(G(z)))] \notag
\end{align}
where \( p_{\text{data}}(x) \) represents the distribution of real data and \( p_z(z) \) is the prior distribution of the noise input. The generator \( G \) aims to minimize the loss by producing data that can deceive the discriminator \( D \), which maximizes its ability to distinguish between real and fake data.

In robotic manipulation, GAN can be leveraged to enhance a robot's ability to perform complex tasks. By learning the distribution of successful manipulation strategies from real-world data, the generator can produce novel and feasible action sequences for tasks such as grasping, object handling, and tool use \cite{chen2023graspada}. The discriminator ensures these generated actions are realistic by comparing them against actual manipulation examples. These approaches allow robots to adapt to new and unstructured environments by generating manipulation policies that are not only effective but also generalizable, thus improving their performance in tasks requiring fine motor skills and adaptability.

\subsection{Variational Autoencoder (VAE)}
Variational Autoencoder (VAE)~\cite{kingma2013auto} aims to learn a probabilistic mapping from a lower-dimensional latent space to the data space. By modeling the data distribution using an encoder-decoder structure, VAEs learn to generate new data by sampling from the learned latent space. One of the key features of VAEs is their ability to perform efficient inference and generation using variational inference.

The training objective for VAEs is to maximize the Evidence Lower Bound (ELBO), which approximates the log-likelihood of the data:
\[
\mathcal{L}_{\text{VAE}} = \mathbb{E}_{q(z|x)}[\log p(x|z)] - D_{\text{KL}}[q(z|x) \| p(z)].
\]
Here, \( q(z|x) \) is the approximate posterior distribution of the latent variables given the data, and \( D_{\text{KL}} \) represents the Kullback-Leibler divergence between the approximate posterior and the prior distribution \( p(z) \). The first term encourages reconstruction of the data, while the second term regularizes the latent space to follow a specified prior.

VAEs enable compact, probabilistic representations that bridge high-dimensional sensory inputs and task-specific outputs in robotic manipulation. Their ability to encode data into a structured latent space facilitates smooth trajectory generation, and adaptable grasp planning. In trajectory generation, VAEs allow robots to explore diverse paths by sampling from the latent space, maintaining continuity and adaptability in motion planning~\cite{fu2024mobile}. For grasp planning, VAEs generate diverse, physically plausible grasps by learning compact latent spaces, enabling adaptability to object variability and uncertainty~\cite{mousavian20196,Sundermeyer2021ContactGraspNet}.

\subsection{Diffusion Model}
The landscape of robotic manipulation has been significantly enriched by the advent of advanced generative models, particularly diffusion models. Among these, Denoising Diffusion Probabilistic Models (DDPM)~\cite{ho2020denoising} and Denoising Diffusion Implicit Models (DDIM)~\cite{song2020denoising} have emerged as powerful frameworks that offer unique advantages for enhancing robotic capabilities in complex manipulation tasks. These models that working by simulating a forward process where data is gradually corrupted with noise, followed by a reverse process that attempts to denoise the data and recover the original input. A neural network typically parameterizes the reverse process, and training the model involves learning to reverse the noise addition at each step.

The forward and reverse processes can be described as
\[
\text{Forward: } q(x_t | x_{t-1}) = \mathcal{N}(x_t; \sqrt{1-\beta_t} x_{t-1}, \beta_t \mathbb{I}),
\]
\[
\text{Reverse: } p_{\theta}(x_{t-1} | x_t) = \mathcal{N}(x_{t-1}; \mu_\theta(x_t, t), \sigma_\theta(x_t, t)).
\]
In the forward process, noise is gradually added to the data \( x_{t-1} \) over multiple timesteps, parameterized by \( \beta_t \). The reverse process aims to recover the data by learning the reverse conditional distribution \( p_{\theta}(x_{t-1} | x_t) \), where \( \mu_\theta(x_t, t) \) and \( \sigma_\theta(x_t, t) \) are learned by the model.

DDPM operates by gradually adding noise to an input data distribution through a series of time steps, transforming it into a simple noise distribution. The model is then trained to reverse this process, gradually denoising the data to reconstruct the original input. This mechanism allows DDPM to generate diverse and realistic outputs that capture the nuances of real-world interactions, which is crucial for robotic manipulation tasks that require adaptability and precision.

On the other hand, DDIM introduces an implicit approach to the diffusion process, allowing for more efficient sampling without compromising the quality of the generated outputs. By providing a deterministic mapping from the noise space back to the data space, DDIM enables faster inference times while maintaining the high fidelity of the generated trajectories. This is particularly beneficial in robotic applications where real-time implementation is essential, such as in dynamic environments where robots must adapt to changing conditions and execute tasks with minimal latency.

\subsection{Probabilistic Flow}
Probabilistic flows~\cite{rezende2015variational} are generative models that learn invertible transformations between a simple prior distribution (e.g., Gaussian) and the complex data distribution (e.g., Images). These models rely on an invertible mapping function, which allows for both efficient likelihood estimation and exact inference. The generative process can be described as applying a sequence of invertible transformations to a simple latent variable.
The generative process is defined by
\[\text{Forward: }z = f_{k+1}(h_k) \circ f_k(h_{k-1}) \circ \cdots \circ f_2(h_1) \circ f_1(x),\]
\[\text{Reverse: }x = f_1^{-1}(h_1) \circ f_2^{-1}(h_2) \circ \cdots \circ f_k^{-1}(h_k) \circ f_{k+1}^{-1}(z),\]
where \( f \) is an invertible function, and \( p(z) \) is the prior distribution over the latent variable. The probabilistic flow model learns the invertible function so that it could gradually transforms the complex data distribution to the simple prior distribution, and uses the reverse function to generate the data.

% Probabilistic flows are firstly used for image generation~\cite{kingma2018glow}.
% Unlike models such as GANs and VAEs, flow-based models explicitly learn the data distribution. This explicit learning allows for the direct computation of exact log-likelihoods, facilitating straightforward model training and evaluation. These characteristics enable flow-based models to be widely used in robotic tasks such as anomaly detection~\cite{mantegazza2022outlier,brockmann2023voraus}, navigation~\cite{wellhausen2020safe}, and manipulation~\cite{xu2023unidexgrasp}.
Unlike models such as GANs and VAEs, flow-based models explicitly learn the data distribution. This explicit learning enables the direct computation of exact log-likelihoods, simplifying model training and evaluation. Thanks to these characteristics, flow-based models have found broad application in robotics, including anomaly detection~\cite{mantegazza2022outlier,brockmann2023voraus}, navigation~\cite{wellhausen2020safe}, and manipulation~\cite{xu2023unidexgrasp}.

\subsection{Auto-regressive Model}
Auto-regressive Models become super popular after the large language model. ChatGPT~\cite{ouyang2022training} shows its strong zero-shot generalization ability in the natural language processing area. Auto-regressive models generate the tokens step by step, conditioning each step on the previous ones. Auto-regressive models factorize the likelihood of the data as a product of conditional distributions, where each data point is generated based on the previous ones. The likelihood factorization is given by
\[
p(x_1, x_2, \dots, x_n) = \prod_{i=1}^n p(x_i | x_1, x_2, \dots, x_{i-1}).
\]
Here, the probability \( p(x_i | x_1, \dots, x_{i-1}) \) is modeled by an auto-regressive model, which learns to capture the dependencies between the data points in the sequence.

Auto-regressive models have shown their excellent generation ability in the natural language processing domain~\cite{radford2018improving, radford2019language, brown2020language, achiam2023gpt, ouyang2022training} and visual generation domain~\cite{chen2020generative, ramesh2021zero, sun2024autoregressive, tian2024visual}, especially with the larger training data and larger model size~\cite{kaplan2020scaling}. In the robotics domain, auto-regressive models could serve as the middle module to generate languages or images for task decomposition. In addition, vision-language-action models~\cite{brohan2023rt, kim2024openvla} extends the multi-modality large language model to involve the action generation in the whole auto-regressive generation process.

\begin{figure*}
  \includegraphics[width=\textwidth]{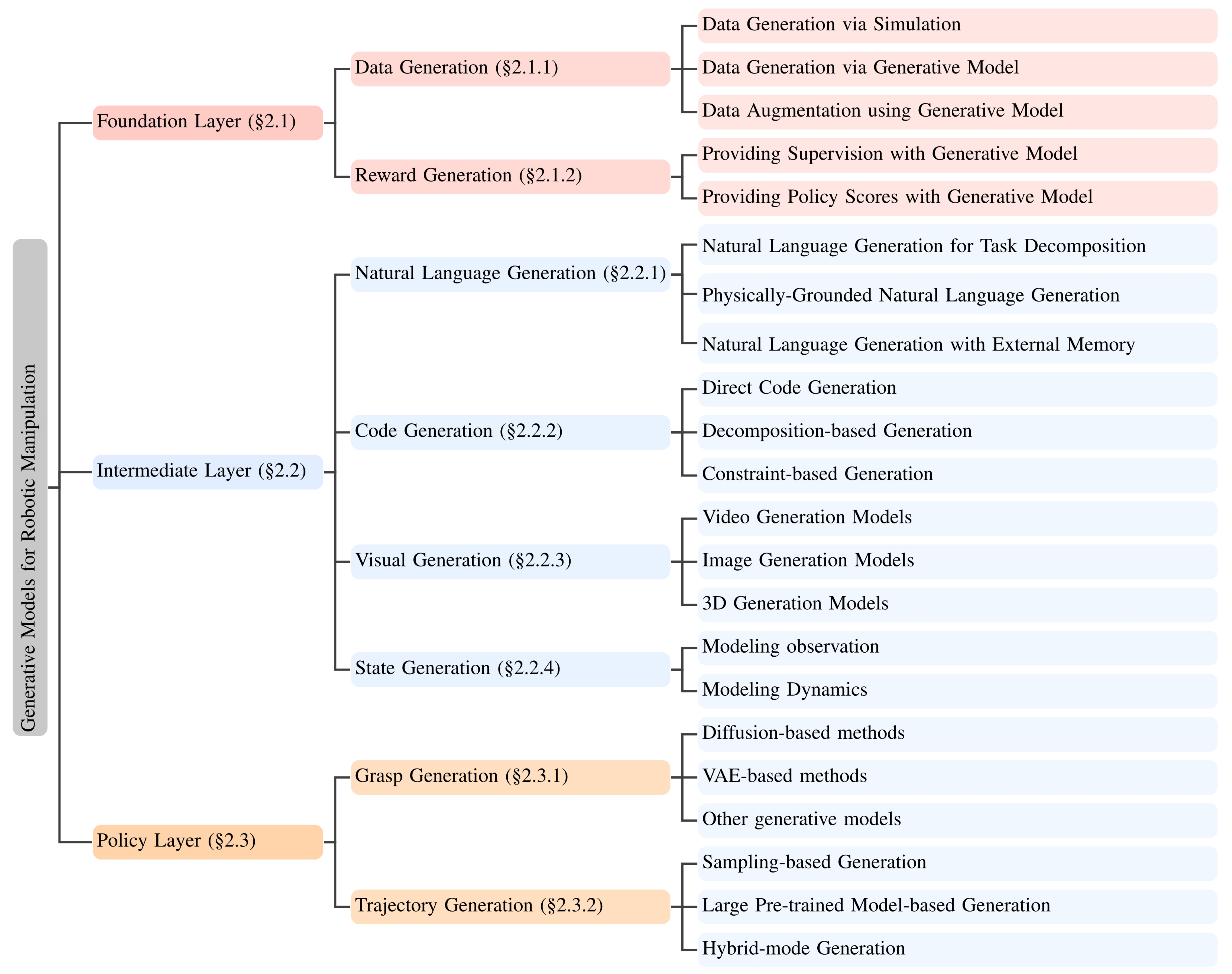}
  \caption{Overview of generative models for robotic manipulation. The taxonomy shows three main categories, each with detailed sub-categories describing specific approaches and methodologies in robotic manipulation.}
  \label{figure:taxonomy}
\end{figure*}

% \input{sections/tax_test}

% \section{Generative Model-based Methods: A Survey}
% \label{section:generation}
% To systematically understand the role of generative models in robotic manipulation, we categorize their applications into three hierarchical layers: \textbf{Foundation Layer}, \textbf{Intermediate Layer}, and \textbf{Policy Layer}. This structure reflects the progressive flow from fundamental data synthesis to high-level decision-making to low-level control. The foundation layer focuses on generating essential resources, such as synthetic data to augment limited datasets and reward signals to guide reinforcement learning, forming the backbone for model training and evaluation. Building on this, the intermediate layer encompasses tasks like language, code, visual, and state generation, which enable robots to interpret instructions, process sensory data, and reason about their environment, bridging perception and action. Finally, the policy layer directly addresses the core robotic manipulation problem, including grasp generation and trajectory planning, translating insights from the lower layers into actionable control strategies. This layered framework highlights the interdependence of these components, ensuring a comprehensive and scalable approach to robotic learning and control.

\section{Foundation Layer} \label{section:foundation_layer}
In robotic manipulation systems, the Foundation Layer serves as the underlying infrastructure that enables effective learning and decision-making. This layer is responsible for generating the essential building blocks required for training and evaluation. 
\subsection{Data Generation}
\label{subsection:data_generation}
\ 
% \newline
%TODO:add one paragraph
% \indent \hl{ add  one paragraph,Why, What, How. Ref code generation}
%taxonomy for data gen.
% \begin{figure}
%   \centering
%   \includegraphics[width=0.48\textwidth]{figures/taxonomy_dataG.pdf}
%   \caption{Taxonomy of data generation in robotic manipulation task, organized around generative AI.}
%   \label{figure:taxonomy_data}
% \end{figure}
\begin{table*}[h]
\centering
\caption{Taxonomy of data generation in robotic manipulation task, organized around generative AI.}
\begin{tabular}{@{}lll@{}}
\toprule
\multicolumn{1}{c}{Taxonomy}                              & \multicolumn{1}{c}{Category} & \multicolumn{1}{c}{Method}                                                                                                                     \\ \midrule
Data Generation via Simulation                            & Traditional Simulator        & e.g. PyBullet, MuJoCo, CoppeliaSim, Unity                                                                                                                                                                                                                 \\ \midrule
\multirow{2}{*}{Data Generation via Generative Model}     & Scenario Generation             & e.g. GRUtopia, Gen2Sim, RoboGen, HOLODECK, Sora, UniSim                                                                                                          \\ \cmidrule(l){2-3} 
                                                          & Demonstration Generation     & \begin{tabular}[c]{@{}l@{}}e.g. MimicGen, CyberDemo, RoboGen, DMD, DexMimicGen, \\ I-Gen, Skill Acquisition, GRRsidents, GRUtopia \end{tabular}   
                                                           % \\ \cmidrule(l){2-3} 
                                                          % & World Gereration              & e.g. Sora, UniSim        
                                                          \\ \midrule
\multirow{3}{*}{Data Augmentation using Generative Model} & LLM/VLM-based                & e.g. ROSIE, RoVi-Aug, DIAL, RACER                                                                                                              \\ \cmidrule(l){2-3} 
                                                          & Diffusion-based              & \begin{tabular}[c]{@{}l@{}}e.g. GenAug, ROSIE, RoVi-Aug, CACATI, Roboagent,  \\ ISAGrasp, DALL-E-Bot\end{tabular}                              \\ \cmidrule(l){2-3} 
                                                          & Point-based                  & e.g. ExAug                                                                                                                                     \\ \bottomrule
\end{tabular}
\label{table:taxonomy_data}
\end{table*}
Data generation is fundamental in advancing robotic manipulation and addressing data scarcity. In this subsection, we provide an overview of existing efforts categorized into three aspects. Data Generation via simulation, Data Generation via a Generative Model, and Data Augmentation using a Generative Model. As shown in Table~\ref{table:taxonomy_data}, we present a granular and layered classification of data generation in robotic manipulation tasks. Together, these techniques form a comprehensive strategy for equipping robotic systems to handle the complexity of real-world environments effectively.

\textbf{Data Generation via Simulation.}
Simulation-based data generation plays a vital role in expanding the availability and diversity of data for manipulation. Modern simulation platforms, powered by high-fidelity physics engines such as PyBullet~\cite{coumans2016pybullet}, MuJoCo~\cite{6386109}, CoppeliaSim~\cite{rohmerversatile}, NVIDIA Omniverse and Unity, enable accurate modeling of physical interactions, including friction, contact, and collisions. These environments allow researchers to design low-cost, controllable training setups that support scalable data generation across various scenarios. 

Despite their advantages, simulation-based methods face Sim-to-Real~\cite{zhang2023building}, where there is a gap between simulated and real world environments. Researchers have developed techniques such as domain randomization to address this~\cite{ramos2019bayessim, lyu2024cross}, introducing variability in visual and physical properties (e.g., lighting, textures) to better align synthetic data distributions with real-world conditions. Dynamic simulation environments also allow fine-grained control over scene parameters such as object shapes, materials, and viewpoints, enabling researchers to curate robust datasets. These methods demonstrate significant success in improving real-world performance while reducing dependency on large datasets.

\textbf{Data Generation via Generative Model.}
Unlike simulations, generative models can produce realistic data without relying on physics engines. Some researchers have begun to utilize foundation models (such as VLM, and diffusion model) to generate data to address data scarcity. In this subsection, we will review scenario generation and presentation generation in manipulation tasks.

\textit{Scenario Generation:} Scenario generation produces enriched data in the simulation for specific tasks. Recent investigation~\cite{li2024foundation} summarizes two common categories for creating simulation scenes for manipulation tasks: (i) Real-to-Sim method, which uses generative models to convert real-world scenes into simulations. (ii) Task-specific method, which involves custom assets tailored to specific tasks. In the former, the 3D models of objects can be constructed from point clouds or images using the generative method. GRUtopia~\cite{wang2024grutopia} further advances scenario generation by creating 100k interactive, finely annotated scenes within 3D society for various robots. While these methods can generate scenes directly based on predefined conditions, challenges remain in accurately capturing the material and physical properties of objects. In the latter, tailored assets offer greater flexibility, but these assets must be properly imported and configured. 
% For example, early efforts GraspNet~\cite{mousavian20196} used variational autoencoders to generate grasp poses. 
In most recent work, the diffusion model is used to develop a multi-gripper grasping scenario generation framework~\cite{freiberg2024diffusion}, which integrates 8 gripper types and over 1k objects to generate cluttered scenes. Gen2Sim~\cite{katara2024gen2sim} uses LLM and vision generative models to automatically generate 3D assets and task descriptions, providing a diverse asset repository for manipulation. Similarly, RoboGen~\cite{wang2023robogen} leverages LLM to first generate asset sizes and configurations, and then create the corresponding assets via text-to-image-to-3D technology. Recently, HOLODECK~\cite{yang2024holodeck} system is proposed to generate diverse 3D scenes that match user-supplied prompts. Moreover, the world simulator is showing significant potential, such as Sora~\cite{brooks2024video} and UniSim~\cite{yang2023learning}, to learn Internet-scale data to simulate diverse environments, further advancing simulation capabilities.

\textit{Demonstration Generation:} Current demonstration generation methods expect to enhance manipulation using minimal human demonstrations. For example, MimicGen~\cite{mandlekar2023mimicgen} introduces a data generation system that automatically creates new demonstrations for novel scenes, generating over 50k demonstrations across 18 tasks from just 200 human demonstrations. CyberDemo~\cite{wang2024cyberdemo} simulates human demonstrations to perform real-world tasks, accounting for both visual and physical variations. On the other hand, DMD~\cite{zhang2024diffusion} leverages state-of-the-art diffusion models to synthesize enhanced expert demonstrations. RoboGen~\cite{wang2023robogen} proposes a full pipeline for generating endless skill demonstrations. DexMimicGen~\cite{jiang2024dexmimicgen} proposes a large-scale generation system that creates 21k demonstrations from just 60 source human demonstrations, exploring the effects of grasping performance across several generation models. I-Gen~\cite{hoque2023interventional} generates corrective intervention demonstration data from limited human interventions. And some work~\cite{ha2023scaling} leverage LLMs for language-guided skill acquisition, generating diverse and rich manipulation demonstrations. Meanwhile, GRResidents~\cite{wang2024grutopia} employs an LLM-driven system for task generation and assignment, producing 300 episodes for loco-manipulation benchmarks.

However, these methods still encounter limitations, such as insufficient diversity in assets and robotic forms. Additionally, data generated by these models may exhibit some lack of controllability and may fall short in physical consistency, particularly in complex scenarios involving physical interactions. 
% To generate data that more accurately reflects real environments, it is often necessary to integrate generative models with physical models.

\textbf{Data Augmentation using Generative Model.}
A prevalent usage for data augmentation is leveraging generative models to enhance data diversity in manipulation. Traditional studies applied augmentation techniques in low-level visual spaces, such as color jittering, blurring, and cropping. As generative models demonstrate powerful content generation capabilities, e.g., text-to-image diffusion models and vision-language models, these advanced models are increasingly being utilized for broader data augmentation.

In detail, diffusion models are commonly used for modifying or editing textures and shapes to perform augmentation procedures. For example, GenAug~\cite{chen2023genaug} uses a depth-guided diffusion model to change backgrounds, object textures, or categories, and add distractors to perform scene augmentation. ROSIE~\cite{yu2023scaling} utilizes a T2I model to create a variety of unseen objects, backgrounds, and distractors based on an existing dataset. RoVi-Aug~\cite{chen2024rovi} uses state-of-the-art diffusion models to expand data with different viewpoints. ISAGrasp~\cite{chen2022learning} proposes implicit shape-augmented grasping via a correspondence-aware generative model to extrapolate human demonstrations for dexterous manipulation. DALL-E-Bot~\cite{kapelyukh2023dall} can rearrange objects within a scene, generating images that represent natural, human-like object placements. Video painting is used to mask human hands~\cite{lepertshadow} and replace them with grippers to augment the data~\cite{bahl2022human}. Mirage~\cite{chen2024mirage} introduces a cross-painting method for unseen robots, which successfully achieves zero-shot transfer between different robotic arms and grippers. CACTI~\cite{mandi2022cacti} leverages diffusion models and in-painting for controllable generation of scenes, enabling automated data augmentation. Similarly, RoboAgent~\cite{bharadhwaj2024roboagent} employs in-painting techniques to propose semantic augmentation, rapidly enriching existing datasets and equipping robots with multiple manipulation skills. And ExAug~\cite{hirose2023exaug} utilizes point clouds to extract 3D information to create more complex and structured augmentations. Note that an important challenge of these approaches is to ensure that the augmented data is physically accurate while guaranteeing semantic and visual diversity. For example, in-painting techniques that edit the objects within the grasper may result in physically unrealistic grasps, leading to poor performance on downstream tasks.

Moreover, LLM and VLM can also contribute to data augmentation in grasping tasks. ROSIE~\cite{yu2023scaling} first locates the augmented regions using an open vocabulary segmentation model, which is then edited. Similarly, RoVi-Aug~\cite{chen2024rovi} also uses VLM to perform segmentation procedures before augmenting the data. RACER~\cite{dai2024racer} proposed an automatic data enhancement pipeline for failure recovery and extended the expert demonstration with LLM. DIAL~\cite{xiao2022robotic} acts as an automated annotator for robotic datasets, effectively importing semantic knowledge from CLIP into existing data.

\subsection{Reward Generation}
\label{subsection:reward_generation}
% \ 
% \newline
% \indent
% VLM-based reward generation
%% Language-driven Grasp Detection
%% Distilling Internet-Scale Vision-Language Models into Embodied Agents
%% Policy Adaptation from Foundation Model Feedback
%% Policy Improvement using Language Feedback Models
Reward generation refers to the process of learning a reward function that guides policy optimization to achieve the highest task success rate. Limitations before the usage of generative models primarily include the challenge of sparse rewards~\cite{yechengeureka_jason_ma2023}, where agents rarely reach the desired goal early in training, resulting in insufficient feedback to effectively optimize policies. The structure of this section focuses on two key applications of generative models in reward generation: providing supervision signals and offering policy scores. Providing supervision signals using generative models involves leveraging large-scale pre-trained models (e.g., Vision-Language Models, VLMs) to generate structured information, such as success indicators~\cite{yuqing_du2023} or goal descriptions, which serve as detailed feedback for policy learning. Providing policy scores using generative models entails using generative models to evaluate policy performance online, such as measuring the distance between predicted actions and goals or generating constraints to optimize policy selection, thereby improving task execution efficiency and stability.

% Rewrite it following policy generation

\textbf{Providing Supervision with Generative Model.}
% provide supervision
The advancement of large-scale pre-trained foundation models has led to significant processes in the generalization ability of robotic manipulation skills. While it is challenging to generalize policy across complex tasks, the large-scale pre-trained foundation models achieve high accuracy in a zero-shot learning manner to unseen environments. Some work adopts large-scale pre-trained foundation models to supervise adapting the manipulation policy.

% Policy Adaptation from Foundation Model Feedback
Ge~\etal~\cite{ge_policy_2023} proposed PAFF framework, which collects domain-specific data using pre-trained foundation models to provide feedback to relabel the domain-specific demonstrations.
They cast the relabelling demonstrations as a visual-to-language retrieval tasks, and compare off-the-shelf pre-trained foundation models, including MoCo~\cite{kaiming_he2020}, DenseCL~\cite{xinlong_wang2021}, MAE~\cite{oren_kraus2023}, GLIP~\cite{liunian_harold_li2022}, CLIP~\cite{radford2021learning} and MDETR~\cite{aishwarya_kamath2021}, in relabelling demonstrations.
By harnessing the generalization capabilities inherent in large-scale vision-language foundation models, specifically in the context of recognizing visual content, they relabeled the recorded demonstrations for fine-tuning the policy in an automatic way and updated the policy in unseen environments with a test-time training strategy.
% Language-driven Grasp Detection
% provide supervision
Building upon the idea of leveraging foundation models for robotics, Vuong~\etal~\cite{vuong2024language} introduce GraspAnything++ dataset, which supports supervision at multiple levels of granularity, including scene-level, object-level, and part-level annotations.
They use Stable Diffusion~\cite{Rombach2022HighResolution} to proceed text-to-image generation for scene generation, adopt OFA~\cite{wang2022ofa} and SAM~\cite{kirillov2023segment} to provide object-level masking, and leverage VLPart~\cite{peize_sun2023} to provide part-level masking, and generate grasp poses with RAGT-3/3~\cite{cao2023nbmod} pre-trained models.
Blank~\etal~\cite{blank2024scaling} propose a similar but object-centric automated labeling pipeline based on VLMs in a concurrent time.
The downside of emsembling multiple foundation models is from the significant computational cost.

% focuses on language-driven grasp detection. Along with VLM foundation models, they create GraspAnything++ dataset supporting scene-level, object-level, and part-level language instructions for generating data of both images and grasp ground-truths. They propose a new diffusion model-based method for conditional generation with a new training strategy that explicitly contributes to the denoising process to detect the grasp poses. The proposed approaches are validated on their GraspAnything++ dataset, which is with language instructions, and also image-only grasp detection datasets, like Jacquard, Cornell, VMRD, and OCID-grasp.

% Distilling Internet-Scale Vision-Language Models into Embodied Agents
% provide supervision
During the early stage of training, rewards are often sparse as agents seldom accomplish the intended objective. Sumers~\etal~\cite{sumers_distilling_2023} treat generative vision-language models as relabelling functions and adopt the Hindsight Experience Replay (HER)~\cite{andrychowicz2017hindsight} framework to densify rewards. They use a generative VLM (Flamingo~\cite{baptiste_alayrac2022}) to generate language instructions for failed trajectories and supervise the training of language-conditioned agents. It distillates the VLM’s domain-general language grounding into domain-specific embodied agents.
Similarly under the HER framework with VLM as a relabelling function, DIAL employs a CLIP model fine-tuned on robotic demonstrations and autonomous data that has been hindsight-labeled by human annotators. This fine-tuned model is then used to further relabel a larger set of offline robotic trajectories, enabling the training of a language-conditioned policy through behavior cloning.

Besides imitation learning, Xie~\etal~\cite{tianbao_xie2023} adopt VLMs to generate reward code to train reinforcement-learning policies. Given the observation and action, the VLMs generate Pythonic representations for reward shaping. Background knowledge is injected with pre-defined helper functions. In order to improve the grounding of VLMs in specific tasks, They inject few-shot examples and human feedback into the prompts when training RL policies, and Ma~\etal~\cite{yechengeureka_jason_ma2023} further enhance the reward design process by leveraging an iterative evolutionary search mechanism, where reward candidates are sampled, evaluated and refined based on textual feedback summarizing training performance. This approach allows EUREKA to autonomously generate interpretable and task-specific dense reward functions that outperform human-designed rewards in most cases.

Although VLMs are effective in providing meaningful signals, their inaccuracies can also lead to misleading outcomes. To address this reward misalignment issue, Fu~\etal~\cite{yuwei_fu2024} thoroughly analyzed the problem and proposed a two-fold solution. First, they introduce lightweight projection layers combined with contrastive learning to align the rewards more effectively. Second, they incorporate a Relay RL mechanism, which employs an auxiliary SAC agent to assist exploration when the primary VLM agent gets stuck in local minima. This collaborative approach allows the agent to escape suboptimal regions and collect diverse samples, which in turn enhances reward alignment and policy learning. Experimental results show that their method significantly improves performance across various sparse reward tasks, demonstrating both the effectiveness of reward alignment and the critical role of Relay RL in addressing exploration challenges.

\textbf{Providing Policy Scores with Generative Model.}
The ability to score a policy in deploying in environments is the core of sample-based planning. A typical way is to measure the distance between the policy prediction and the goal in metric space~\cite{minttu_alakuijala2022,yecheng_jason_ma2023,yan_hu2023,suraj_nair_00032022,mohit_shridhar2021}.
The downside is the task-related hand-crafted distance function or objective function for constructing the metric space. It is very challenging for this scoring approach to generalize beyond the pre-defined task space.

Generative VLMs also act as reward functions online for policy selection in deployment.
% VLMPC: Vision-Language Model Predictive Control for Robotic Manipulation
% policy scoring
Zhao~\etal~\cite{zhao2024vlmpc} design reward functions with the assistance of VLMs for Model predictive control.
Besides the typical pixel-distance cost between future predictions and goal images~\cite{finn2017deep}, they add constraints generated using VLMs to avoid collision with interference objects.
The reward function is a combination of these two costs with an adaptive weight generated by VLMs. Even though it improves the efficiency, it still requires human-engineered design between the predicted images and goal images. Yu~\etal~\cite{yu_2023_arxiv} decompose the reward function generation into two stages: motion description, interprets the task description into a structured natural language description of the desired robot motion; and reward coding, generates code based on pre-defined functions and low-level motion description in a few-shot learning way.
Without explicitly estimating future images, the VLMs directly estimate the reward instead. Its integration into the MPC framework improves the stability of the control system.

% Learning Reward for Robot Skills Using Large Language Models via Self-Alignment
% policy scoring
Directly adopting generative VLMs as reward functions is challenging to ground the domain-general model to domain-specific tasks. It requires to be further grounded with specific task-related environment information~\cite{yuwei_zeng2024, sumers_distilling_2023}.
Zeng~\etal~\cite{yuwei_zeng2024} extend the bi-level optimization structure in inverse reinforcement learning~\cite{ke2021imitation}. The inner loop of the bi-level optimization contains a learned reward function scoring the sampled strategies and conducting the execution, which the outer loop updates this reward function by aligning the ranking between generative VLMs and the output of the reward function. Besides grounding to domain-specific tasks, it is token efficient than adopting VLMs as reward functions.
% Policy Improvement using Language Feedback Models
Zhong~\etal~\cite{zhong_policy_2024} propose a pipeline to adopt LLM to facilitate policy improvement. They use LLM to generate feedback sets to train a feedback model. By identifying desirable actions with the feedback model online, they update the policy.

\section{Intermediate Layer}\label{section:intermediate_layer}
In robotic manipulation paradigms, the Intermediate Layer serves as a crucial component that connects high-level task planning with low-level policy execution. Its primary role is to generate structured, interpretable representations that bridge the abstract commands from task planners and the executable actions required by robotic systems. Depending on the task and context, the Intermediate Layer involves various generation mechanisms, including natural language and code generation for task decomposition, visual and state generation for producing scene-level representations and future predictions. These intermediate outputs simplify the complexity of robotic decision-making by breaking down tasks into manageable sub-goals, ensuring that the robot can operate effectively in dynamic and uncertain environments. By modularizing these processes, the Intermediate Layer enhances the adaptability and scalability of robotic systems across diverse applications.
\subsection{Natural Language Generation}
% \newline
% \indent
\label{subsection:language_generation}
Natural language generation has emerged as a powerful tool for creating executable tasks for robots in a linguistic format. While robots often struggle with complex, long-range tasks, decomposing these tasks into sequences of subtasks before execution has proven to be an effective strategy. Large Language Models (LLMs) have demonstrated remarkable task planning capabilities~\cite{huang2022language}, making them ideal for this purpose. As a result, numerous studies~\cite{dalal2024plan, zhao2024large, ha2023scaling, liu2023reflect, lin2023gesture, wang2024large} have leveraged LLMs to break down long-horizon robotic tasks into manageable sequences of subtasks. This approach not only simplifies the execution process for robots but also enhances their ability to handle more sophisticated and varied tasks.

\begin{figure}
    \centering
    \includegraphics[width=\linewidth]{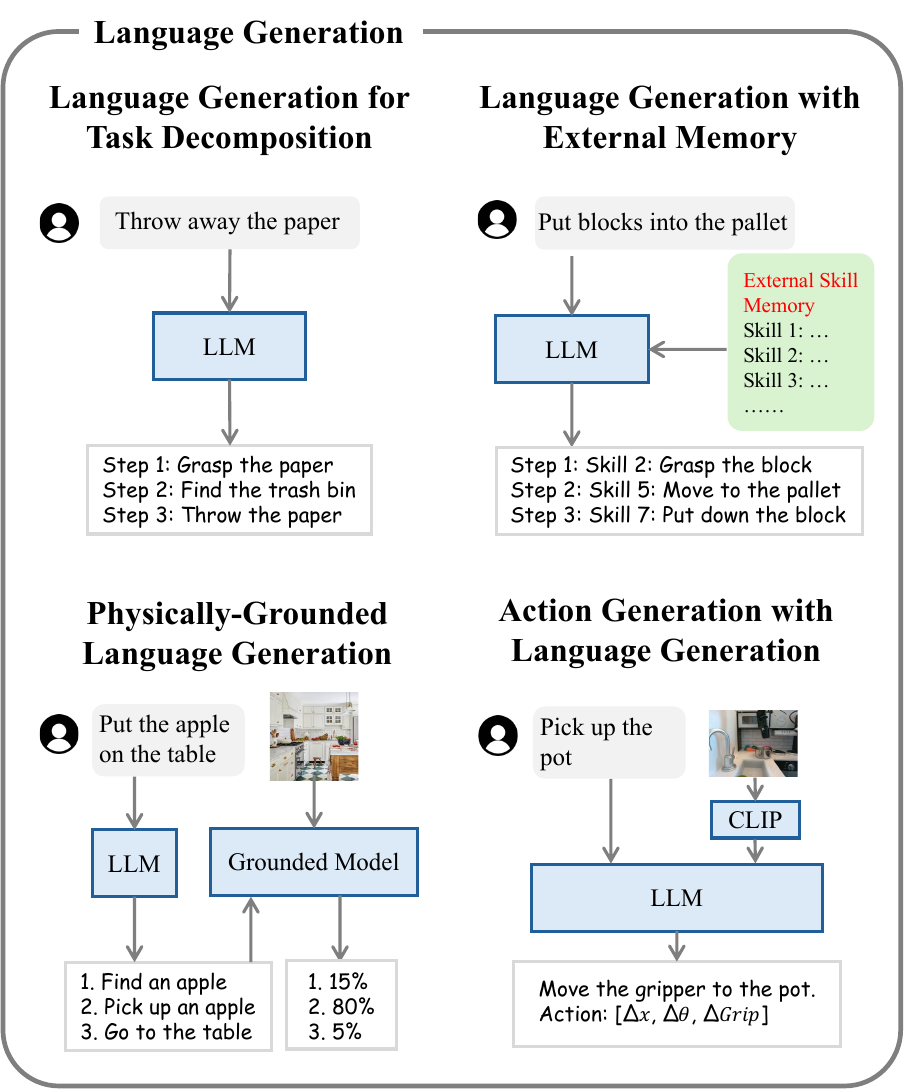}
    \caption{Overview of natural language generative methods for manipulation. Natural language generation could be used for task decomposition, and external memory could be introduced to save and retrieve pre-defined skills. Physically-grounded language generation considers the available operations in the current space. Vision-language-action models generate the action along with the language.}
    \label{fig:language_generative_models}
\end{figure}

This section explores the advancements brought by generative models and their applications in robotic task planning. Specifically, we discuss Natural Language Generation for task decomposition, which involves breaking down complex tasks into manageable sub-goals using language as an intermediary; physically-grounded language generation, where language outputs are tied to physical states or interactions in the environment, ensuring relevance to robotics; and Language Generation with External Memory, which leverages memory-augmented architectures to enable reasoning over extended contexts and improve task planning in long-horizon scenarios. Figure~\ref{fig:language_generative_models} shows the structure of this section.

\textbf{Natural language Generation for Task Decomposition.} The task planning capabilities of Large Language Models (LLMs) were first comprehensively evaluated in \cite{huang2022language}. This study assessed the executability and correctness of GPT-2, GPT-3, and Codex, concluding that LLMs of sufficient size can generate mid-level tasks without fine-tuning. LLMs have advanced rapidly in recent years, particularly following the release of ChatGPT. A more recent evaluation of language plan generation abilities in popular LLMs is presented in \cite{li2024embodied}, encompassing models such as Claude-3, Claude-3.5, Cohere Command R, Gemini 1.0, Gemini 1.5, GPT-3.5, GPT-4, GPT-4o, Llama 3, Mistral, and o1. Table~\ref{tab:eval_subgoal_decomposition} demonstrates that o1-preview achieves the highest task success rate on both VirtualHome~\cite{puig2018virtualhome} and BEHAVIOR~\cite{srivastava2022behavior} datasets, as well as the best execution success rate on BEHAVIOR. OpenAI's o1 model's capacity to generate extended chain-of-thought reasoning prior to producing a final answer endows it with superior reasoning capabilities compared to GPT-4o. This underscores the critical importance of robust reasoning abilities in effective task decomposition.

A large-scale language plan generation dataset is introduced in EmbodiedGPT~\cite{mu2024embodiedgpt} based on Ego4D dataset~\cite{grauman2022ego4d}. EmbodiedGPT also proposes a general framework that integrates both high-level language plan generation and low-level action generation. When multi-modality information are encountered, Matcha~\cite{zhao2023chat} agent is proposed to guide epistemic actions and to analyze the resulting multi-modal sensations (including vision, sound, haptics, and proprioception). Additionally, it facilitates the planning of entire task executions based on information acquired interactively.

Although LLMs demonstrate robust language planning capabilities for manipulation tasks, they are too large to be deployed on a capacity-limited and off-the-shelf devices. To address this challenge, DEDER~\cite{choiembodied} proposes an innovative solution. This approach constructs a task decomposition dataset with an embodied knowledge graph using an LLM, and subsequently trains a compact language model to learn policies from this graph. Through this process, the reasoning abilities of the LLM are effectively distilled into a smaller language model, enabling real-time inference on edge devices. This method bridges the gap between the powerful capabilities of LLMs and the practical constraints of resource-limited hardware.

% REFLECT~\cite{liu2023reflect}  hierarchically summaries the observation sequences and proposes a failure reasoning module to fix the failures.

\textbf{Physically-Grounded Language Generation.}
LLMs often struggle to consider real-world conditions when decomposing tasks, potentially producing sub-plans that are not executable by embodied agents. Several approaches have been developed to address this issue.
SayCan~\cite{Ahn2022DoAsICan} employs temporal-difference-based (TD) reinforcement learning to train a value function model, estimating the achievability of each plan. Grounded Decoding~\cite{huang2024grounded} trains a grounded model to generate feasible actions, with final plans determined by both the LLM and the grounded model. Inner Monologue~\cite{huang2022inner} incorporates various types of grounded closed-loop feedback into the robot planning pipeline, including success detection, passive scene description, and active scene description, demonstrating the significance of physically grounded feedback for task success.
PhyGrasp~\cite{guo2024phygrasp} utilizes an LLM to generate part and physical information, integrating them to produce task-specific grasping pair matches. Ren et al.~\cite{ren2023leveraging} gather diverse natural language descriptions for different tools and apply language-conditioned meta-learning to develop effective policies.
These approaches aim to bridge the gap between language models and physical reality, enhancing the applicability of LLM-generated plans in real-world scenarios.

\textbf{Language Generation with External Memory.}
Several studies have explored the integration of external memory with Large Language Models (LLMs) to enhance task performance. SayPlan~\cite{rana2023sayplan} leverages the hierarchical structure of 3D scene graphs (3DSG) to enable LLMs to perform semantic searches for task-relevant subgraphs from a condensed version of the full graph. RoboMP$^2$~\cite{lv2024robomp} implements Retrieval Augmented Generation (RAG)~\cite{lewis2020retrieval, liu2021makes} to allow Multimodal Large Language Models (MLLMs) to select semantically similar plans from a codebase. Text2Motion~\cite{lin2023text2motion} retrieves robot skills by conducting geometric feasibility planning~\cite{agia2023stap} throughout its search process. These approaches demonstrate the potential of combining LLMs with external knowledge sources to improve language generation and task execution in various domains.

% \textbf{Action Generation with Natural Language Generation.}
% Vision-Language-Action (VLA) models regard actions as a set of language tokens, LLMs to generate actions and language simultaneously. RT-2~\cite{brohan2023rt} introduced the VLA concept and developed two VLA models based on PaLI-X~\cite{chen2023pali} and PaLM-E~\cite{driess2023palm}. VLA models demonstrate enhanced generalization performance, leveraging the extensive pre-training knowledge of LLMs.
% As RT-2 is not open-source, OpenVLA~\cite{kim2024openvla} was proposed as a publicly accessible alternative. OpenVLA is built upon the Prismatic multi-modality LLM~\cite{karamcheti2024prismatic} and utilizes DINOv2~\cite{oquab2023dinov2} and SigLIP~\cite{zhai2023sigmoid} as visual encoders. It directly outputs action tokens using the LLM. Pre-trained on the large-scale Open-X-Embodiment dataset~\cite{o2023open}, OpenVLA exhibits superior performance compared to RT-2.

\begin{table}[t]
\setlength{\tabcolsep}{12pt}
\centering
\caption{Subgoal Decomposition Performance of common LLMs. V: VirtualHome, B: BEHAVIOR. SR: Successful Rate.}
\begin{tabular}{lcccc}
\toprule
\multirow{2}{*}{Model} & \multicolumn{2}{c}{Task SR} & \multicolumn{2}{c}{Execution SR} \\ \cmidrule(lr){2-3} \cmidrule(lr){4-5}
& V & B & V & B \\
\midrule
Claude-3 Haiku & 78.4 & 30.0 & 82.8 & 35.0 \\
Claude-3 Sonnet & 83.1 & 39.0 & 86.4 & 43.0 \\
Claude-3 Opus & 86.7 & 41.0 & 89.9 & 47.0 \\
Claude-3.5 Sonnet & 89.1 & 39.0 & 92.0 & 44.0 \\
Cohere Command R & 71.3 & 15.0 & 78.1 & 25.0 \\
Cohere Command R+ & 77.8 & 25.0 & 83.7 & 37.0 \\
Gemini 1.0 Pro & 70.4 & 24.0 & 84.6 & 33.0 \\
Gemini 1.5 Flash & 89.1 & 34.0 & 94.1 & 42.0 \\
Gemini 1.5 Pro & 87.0 & 31.0 & 91.1 & 37.0 \\
GPT-3.5-turbo & 69.2 & 24.0 & 81.4 & 36.0 \\
GPT-4-turbo & 85.5 & 38.0 & 94.1 & 47.0 \\
GPT-4o & 87.6 & 49.0 & 91.1 & 55.0 \\
Llama 3 8B Instruct & 48.8 & 22.0 & 58.0 & 29.0 \\
Llama 3 70B Instruct & 78.4 & 21.0 & 87.3 & 30.0 \\
Mistral Large & 84.3 & 31.0 & 92.0 & 38.0 \\
Mixtral 8x22B MoE & 80.5 & 28.0 & 90.2 & 33.0 \\
o1-mini & 79.3 & 31.0 & 84.6 & 39.0 \\
o1-preview & 89.4 & 57.0 & 93.2 & 62.0 \\
\bottomrule
\end{tabular}
\label{tab:eval_subgoal_decomposition}
\end{table}
\subsection{Code Generation}
\label{subsection:code_generation}
\begin{figure*}
  \includegraphics[width=\textwidth]{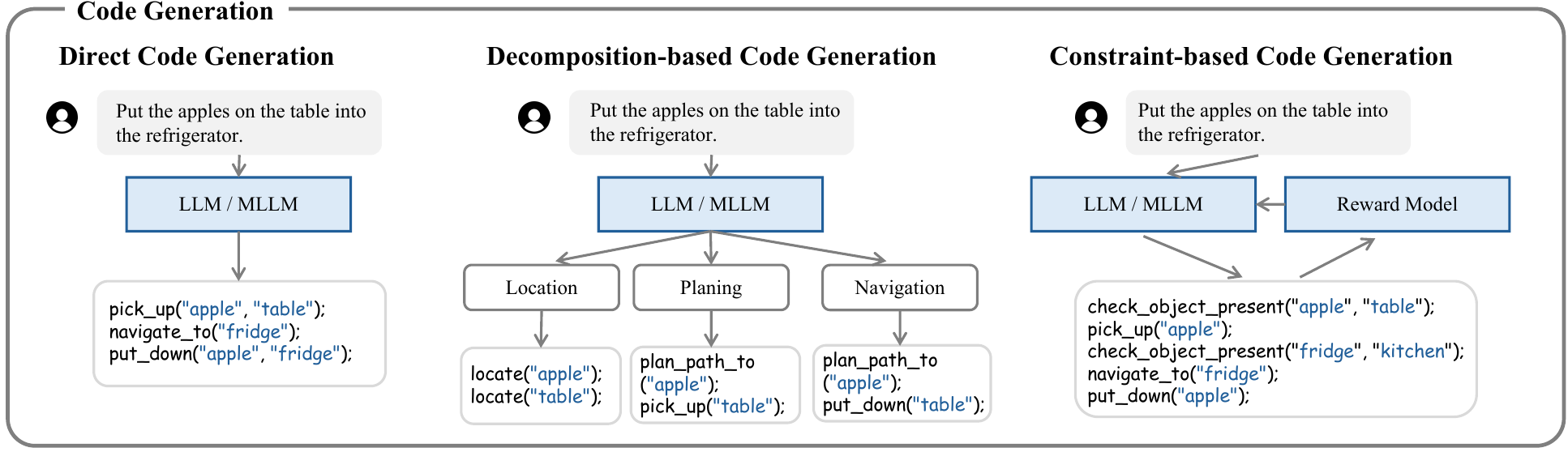}
  \caption{Overview of code generation. Research in robotic code generation can be broadly categorized into three approaches: direct code generation, decomposition-based code generation, and constraint-based code generation.}
  \label{figure:code_gen}
\end{figure*}

% \newline
\indent
Building upon advancements in language generation, code generation aims to translate high-level natural language instructions into executable robot control programs. By leveraging LLMs or MLLMs, these approaches strive to make robot programming more accessible, allowing non-experts to specify tasks at a conceptual level while automated systems handle the complexities of translating those instructions into low-level commands.
As shown in Figure \ref{figure:code_gen}, current research in robotic code generation can be broadly categorized into three approaches: direct code generation, decomposition-based code generation, and constraint-based code generation.

\textbf{Direct Code Generation.}
Direct code generation methods leverage LLMs to produce executable code directly from natural language instructions without intermediate decomposition. 
Code as Policies \cite{liang2023code} first demonstrated that LLMs can generate robot action sequences from high-level commands using predefined APIs, showing promising results in both manipulation and navigation tasks. Building on this, STEVE \cite{zhao2023see} employs a three-layer architecture—vision perception, language instruction, and code action—to translate multimodal inputs into executable code. RoboScript \cite{chen2024roboscript} further integrates perception (e.g., object detection, grasp planning) and motion planning tools into a unified framework, supporting deployment in simulation and on physical robots. RobotGPT \cite{jin2024robotgpt} introduces decision and verification mechanisms to improve reliability and safety, while Instruct2Act \cite{huang2023instruct} demonstrates adaptability to various instruction modalities by generating Python programs that encompass perception, planning, and control loops. Zhi et al. \cite{zhi2024closed} propose a closed-loop approach that continuously refines code execution based on feedback, enhancing resilience to environmental changes. Although direct methods are appealing for their simplicity and broad applicability, they may struggle when facing highly complex, long-horizon tasks that require intricate reasoning, precise motion control, or error recovery capabilities.

\textbf{Decomposition-based Code Generation.}
To improve the reliability and interpretability of code generation, decomposition-based approaches first break down complex tasks into simpler subtasks before generating code. ProgPrompt \cite{singh2023progprompt} proposes a two-stage framework with a Task Decomposition Agent (TDA) and a Script Generation Agent (SGA). The TDA analyzes the high-level task and decomposes it into a sequence of primitive actions, while the SGA generates specific code for each primitive action using predefined templates and APIs. 
RoboCodeX \cite{mu2024robocodex} takes a similar idea further, using a tree-structured framework to decompose tasks into object-centric manipulation units that incorporate physical preferences (e.g., grasp points, motion constraints).
% RoboCodeX \cite{mu2024robocodex} introduces a more sophisticated tree-structured multimodal code generation framework. It decomposes high-level instructions into multiple object-centric manipulation units, each incorporating physical preferences such as grasp positions, approach directions, and motion constraints. The framework also includes specialized multimodal reasoning datasets and iterative self-updating mechanisms to enhance code generation quality.
GenSim \cite{wang2024gensim} leverages LLMs to automatically generate rich simulated environments and expert demonstrations, aiding multi-task policy training and scaling task complexity. 
% GenSim \cite{wang2024gensim} presents an innovative approach that uses LLMs to automatically generate rich simulation environments and expert demonstrations. It operates in two modes: goal-directed generation, where LLM proposes a task curriculum to solve a target task, and exploratory generation, where LLM bootstraps from previous tasks to iteratively propose novel tasks. The framework effectively scales up simulation tasks and enables better task-level generalization when used for multitask policy training.
Octopus \cite{yang2024octopus} and Alchemist \cite{karli2024alchemist} integrate real-time feedback and interactive dialogues, enabling users to refine generated code through natural language instructions and visualization tools.
% Octopus \cite{yang2024octopus} presents an embodied vision-language programmer that breaks down tasks based on environmental feedback. It employs a novel feedback-driven learning scheme that enables the system to learn from execution experiences and improve its code generation capabilities. Similarly, Alchemist \cite{karli2024alchemist} proposes an end-to-end system for natural language-based robot program authoring. It supports iterative development through a chat-based interface, allowing users to refine and debug generated code through natural language interactions. The system also provides visualization tools and a modular architecture that facilitates code reuse and adaptation. These decomposition-based methods demonstrate superior reliability and interpretability compared to direct generation approaches, as they provide better control over the code generation process and make it easier to identify and fix errors. 
Such methods often show higher success rates and improved performance on complex, multi-step tasks. However, the additional complexity in framework design, data requirements, and runtime overhead can increase development costs.

\textbf{Constraint-based Code Generation.}
In real-world robotics, physical and spatial constraints cannot be ignored. Constraint-based approaches explicitly incorporate these factors into the code generation process to ensure the generated code respects real-world limitations and requirements. 
ReKep \cite{huang2024rekep} introduces Relational Keypoint Constraints that specify spatial and temporal relationships between keypoints in the environment. These constraints guide the code generation process by ensuring that generated robot motions maintain desired spatial relationships and follow natural motion patterns. VoxPoser \cite{huang2023voxposer} proposes composable 3D value maps as constraints for robot manipulation. These maps encode spatial preferences and constraints, which are used to optimize robot trajectories and generate corresponding control code. Language to Rewards \cite{yu_2023_arxiv} presents a novel approach that translates natural language instructions into reward functions, which then serve as constraints for generating robot control code through reinforcement learning. When Prolog Meets Generative Models \cite{saccon2024prolog} introduces a hybrid approach that combines the logical reasoning capabilities of Prolog with the generative power of LLMs. The system uses Prolog to maintain a knowledge base of physical constraints and rules, which guides the LLM in generating code that respects these constraints. The framework also includes mechanisms for temporal parallel plan generation and automated translation of plans into executable code. By incorporating explicit constraints, these methods can better handle complex spatial relationships and physical interactions, producing more reliable and physically feasible robot behaviors. However, they may require more sophisticated constraint specification and optimization techniques compared to other approaches.

Direct generation methods offer simplicity and end-to-end solutions but may struggle with complex tasks. Decomposition-based methods provide better reliability and interpretability but require more sophisticated architectures. Constraint-based methods excel at handling physical constraints but may need more complex optimization techniques. Future research might focus on combining the strengths of these different approaches while addressing their respective limitations.

% \begin{table*}[ht]
% \setlength{\tabcolsep}{12pt}
% \centering
% \caption{Comparative Overview of Code Generation Approaches}
% \label{tab:code_generation_compare}
% \begin{tabular}{l|c|c|c}
% & Direct & Decomposition & Constraint \\ 
% Complex Task & Moderate & High & High \\
% Interpretability & Low & High & Moderate \\
% Physical Feasibility & Limited & Moderate & High \\
% Engineering Overhead & Low & High & High \\
% User Interaction & Low & Moderate-High (dialog) & Moderate (constraint design) \\
% \end{tabular}
% \end{table*}
\begin{figure*}
  \includegraphics[width=\textwidth]{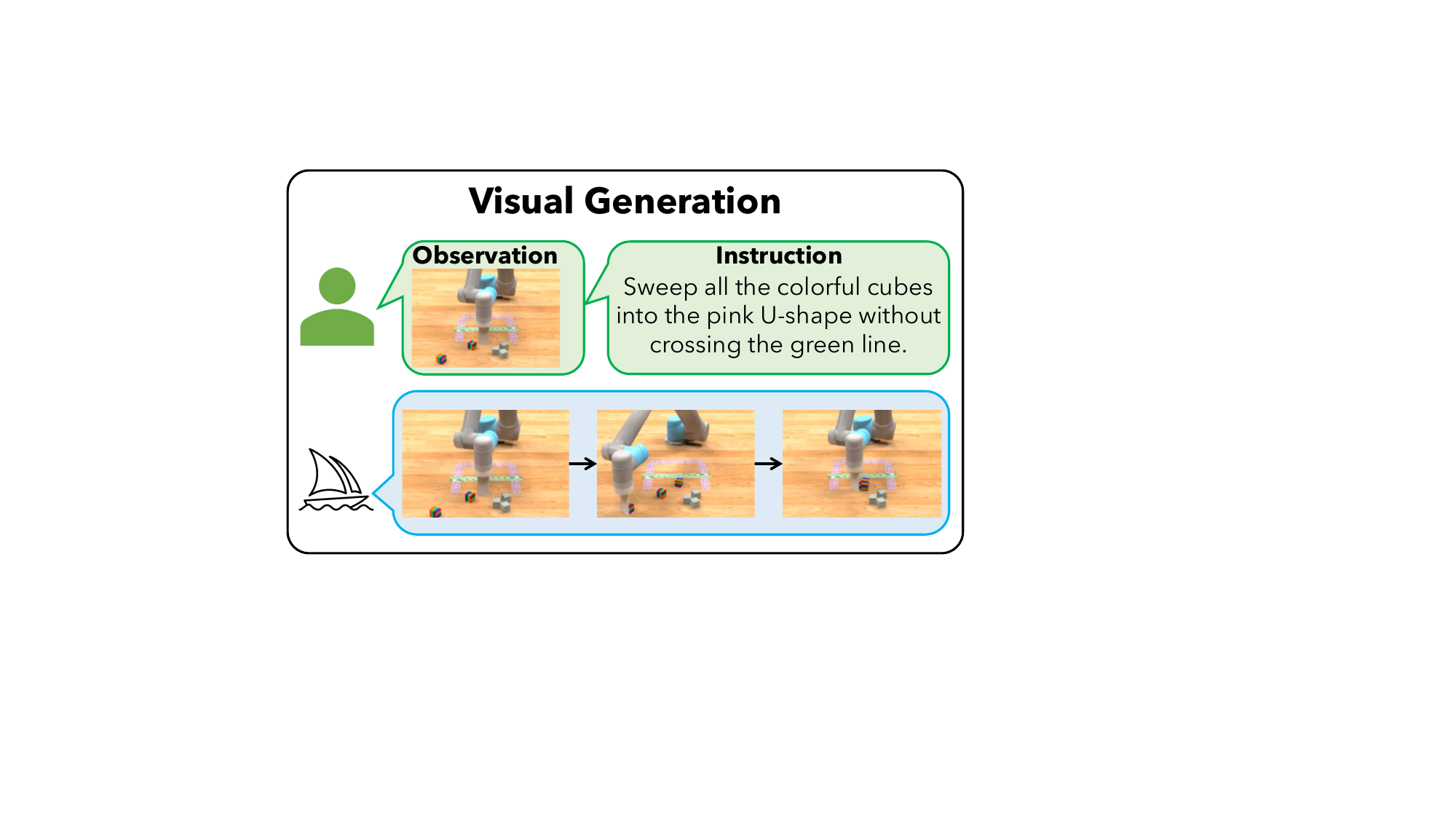}
  \caption{
      Typical pipelines of visual generation models utilized in robot manipulation. 
      Note that the figures of RGB and RGB-D observations are from VIMA-Bench~\cite{jiang2023vima} and GNFactor~\cite{ze2023gn-factor}, respectively.
  }
  \label{figure:visual_gen}
\end{figure*}

\subsection{Visual Generation}
\label{subsection:visual_generation}
% \newline
\indent
% 具体的模型写一下
% 好几类生成的那种unified model，可以放在前面单独列出来。（introduction也是）
%%%%%%%%%%%%%%%%%%%%%%%%%%%%%%%%%%%%%%%%%%%%%%%%%%%%%%%%%%%%%%%%
%                        Final version                         %
%%%%%%%%%%%%%%%%%%%%%%%%%%%%%%%%%%%%%%%%%%%%%%%%%%%%%%%%%%%%%%%%
Due to the evolution of the generation paradigm and larger-scale pre-training, the quality of images, videos, and point clouds generated by visual generation systems~\cite{Rombach2022HighResolution,karras2020StyleGAN,wang2023modelscopeT2V,ho2022video_diffusion_models} has improved substantially.
These advancements in visual generative models are transforming robotic manipulation, enabling robots to better interpret and interact with their environments through synthesized visual cues.
By leveraging highly realistic visual outputs, robots can simulate and predict complex manipulation tasks in virtual settings before attempting them in the real world, as illustrated in Figure ~\ref{figure:visual_gen}. 
This not only enhances the robot's ability to learn and generalize across diverse tasks, but also complements recent progress in language and code generation, enabling robots to interpret and execute instructions across multiple modalities. 
As a result, these advancements pave the way for safer and more adaptable robot manipulation, contributing to a deeper understanding of multi-object interactions, physical constraints, and task-specific movements.

% \subsubsection{Video Generation Models}
\textbf{Video Generation Models.}
Video generation models focus on generating videos conditioned on text or images.
They holds significant potential for robot manipulation because they enable robots to visualize and plan complex sequences of actions. 
Additionally, video generation models can incorporate temporal dynamics, offering a better understanding of how actions evolve over time, which is crucial for tasks involving movement, interaction with objects, or long-horizon planning. 
% This predictive capability not only enhances a robot’s decision-making process but also improves its ability to adapt to dynamic environments and handle a wide range of manipulation challenges with greater efficiency and safety.

\citet{du2024UniPI} introduces an innovative approach UniPi to creating generalized agents by harnessing text-to-video diffusion models~\cite{ho2022video_diffusion_models}.
UniPi redefines sequential decision-making in robot manipulation as text-conditioned video generation. 
By leveraging the planning capabilities inherent in video diffusion models, UniPi generates sequences of future frames that visually depict intended actions throughout the manipulation process.
Control actions are subsequently derived directly from these generated frames, effectively guiding the agent in executing the task.

Following this line of research, SLOWFAST-VGEN~\cite{hong2024SlowFast-VGen} incorporates episodic memory to enhance action-conditioned long video generation.
VLP~\cite{cen2024left_right} extends this single-modal planning to multi-modal planning by introducing the collaboration of vision planning by Stable Video Diffusion~\cite{blattmann2023stablevideodiffusion} and language planning by ChatGPT.
Similarly, HiP~\cite{ajay2024HiP} utilizes multiple foundation models for hierarchical planning, where video diffusion models form the core of visual planning given a language subgoal and the initial observation.

However, previous methods, such as UniPi, face limitations due to the open-loop nature of planning. 
As pointed out by \citet{bu2024CLOVER}, planned actions can deviate from actual trajectories if not verified, which can be especially problematic in long-horizon tasks and dynamic environments. 
To address this, CLOVER~\cite{bu2024CLOVER} proposes a feedback mechanism to improve the adaptability and accuracy of robotic control.

These approaches typically rely on separate models for video generation and action planning, which can result in error accumulation due to the interplay between the models. 
GR-1~\cite{wu2023unleashing} addresses this limitation by introducing a unified GPT-style transformer designed for multi-task visual robot manipulation. 
By utilizing large-scale video generative pre-training, GR-1 successfully transfers generative capabilities to robot manipulation tasks. 
The follow-up work, GR-2~\cite{cheang2024gr-2}, builds on GR-1, significantly scaling up both the pre-training dataset and the range of tasks the model can handle, including complex scenarios like bin-picking with over 100 objects.

% \subsubsection{Image Generation Models}
\textbf{Image Generation Models.}
In robot manipulation tasks, image generation models provide an alternative to video generation models. 
\citet{black2023SuSIE} proposes SuSIE, which leverages an image diffusion model to reframe the planning problem as an image editing task~\cite{brooks2023instructpix2pix}. 
Using language commands, SuSIE’s image editing model generates intermediate subgoals that guide a low-level policy in executing the task. 
This approach focuses on generating discrete, action-relevant images, which can simplify task planning by providing clear, actionable visual cues.

In a similar vein, CoTDiffusion~\cite{ni2024CoTDiffusion} enhances image generation by incorporating multi-modal prompts through a semantic alignment module, enabling the generation of coherent subgoal images conditioned on various input signals. 
This follows a chain-of-thought approach, improving instruction-following in long-horizon manipulation tasks, where reasoning over extended action sequences is critical.

Surfer~\cite{ren2024surferprogressivereasoningworld} draws inspiration from I-JEPA~\cite{assran2023I-JEPA}, generating the next action frame by conditioning on the encoding of the current action frame and predictions of subsequent actions. 
This progressive reasoning model supports coherent decision-making over multiple steps, enhancing long-term task execution.

% \subsubsection{3D Generation Models}
\textbf{3D Generation Models.}
In robot manipulation tasks, while traditional image and video generation models provide insights into 2D visual representations, 3D generation models are essential for capturing the spatial and semantic complexity of real-world environments. 
One approach, TAX-Pose~\cite{pan2023tax-pose}, processes two segmented point clouds to predict a new point cloud, combining target points and residual corrections to generate accurate 3D structures. 
This enables more precise planning in environments where spatial relationships are critical.

IMAGINATION POLICY~\cite{huang2024imagination-policy} expands the use of generative models by predicting the movement of points in a 3D space. It employs an iterative velocity model to simulate the evolution of the scene and envision potential future states. 
These predicted states are then converted into actionable commands using rigid-body motion estimation, facilitating more dynamic and continuous task execution.

Several approaches shift focus to 3D scene radiance, providing richer, more detailed representations of the environment. 
\citet{dasgupta_uncertainty-aware_2024} proposes an ensemble of partially constructed Neural Radiance Fields (NeRF) models to assess model uncertainty, enabling the selection of optimal actions—whether visual or re-orientational—by balancing informativeness and practicality. 
Another model, GNFactor~\cite{ze2023gn-factor}, enhances NeRF’s generalizability by incorporating a reconstruction loss alongside behavior cloning. 
This method employs a Perceiver Transformer~\cite{shridhar2022peract} for decision-making, which has shown strong performance in both simulated and real-world robotic manipulation tasks.

Despite their advances, these methods often fail to account for the dynamic, interactive nature of scenes, particularly in tasks where objects interact over time. 
% To address this limitation, ManiGaussian~\cite{lu2025manigaussian} introduces a dynamic Gaussian Splatting framework that captures spatiotemporal dynamics within the scene. 
% By parameterizing these dynamics through a Gaussian world model, ManiGaussian improves the model's ability to predict object interactions, resulting in more adaptive and robust manipulation strategies in complex, real-world environments.
To address this limitation, ManiGaussian~\cite{lu2025manigaussian} introduces a dynamic Gaussian Splatting framework to model the spatiotemporal dynamics of the scene.
By using a Gaussian world model, the system can parameterize the distribution of these dynamic interactions, enabling the robot to adapt and plan more effectively in complex and interactive environments.

\subsection{State Generation}
\label{subsection:state_generation}

State Generation is the process of creating meaningful and compact representations of an environment or task dynamics, enabling robots to interpret and interact with their surroundings effectively. 
In contrast to generating visual predictions in input space, it is efficient and compact to generate predictions in latent space~\cite{lecun2022path}.
Compared to predictions in image space as discussed in Section~\ref{subsection:visual_generation}, latent states possess a footprint with limited memory, thereby facilitating the capability to simulate thousands of trajectories concurrently.
Before generative models, state representation often depended on explicit, high-dimensional sensory data, which was not only computationally expensive but also struggled to generalize across complex or dynamic scenarios.

This section explores two key aspects: how generative models can enhance observation modeling and dynamics modeling. Modeling observation leverages generative models to transform raw sensory inputs into structured latent representations, capturing essential features while reducing noise and redundancy, which streamlines downstream tasks like planning and control. Modeling dynamics, on the other hand, focuses on predicting state transitions and understanding action dependencies, allowing robots to anticipate future states and plan trajectories with greater accuracy and adaptability in evolving environments.

\begin{figure}
    \centering
    \includegraphics[width=\linewidth]{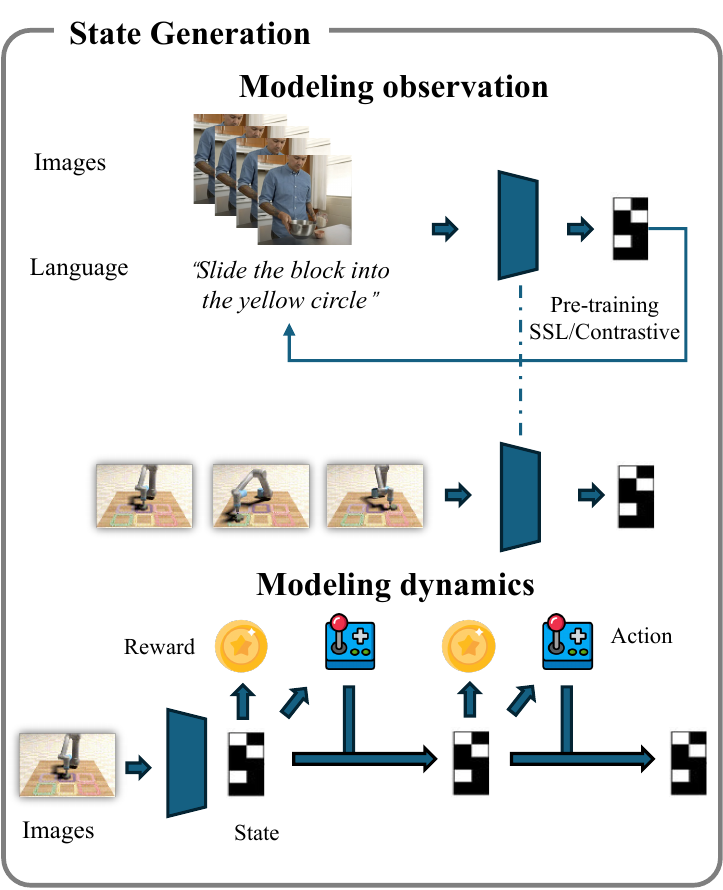}
    \caption{Overview of state generative models. The latent state can represent the environment's observation (top) or encode the dynamic information of the task (down).}
    \label{fig:state_generation}
\end{figure}

\textbf{Modeling observation.}
To effectively represent the environment for downstream tasks, modeling observation focuses on learning compact and meaningful visual or multi-modal representations that capture relevant features from raw sensory inputs.
% R3M, VIP, LIV, EC2
Nair~\etal~\cite{suraj_nair_00032022} study how visual representations pre-trained on ego-centric human video data for downstream robotic manipulation tasks. Three solutions, time-contrastive learning, video-language alignment, and L1 penalty, are adopted to achieve this goal. The approach is evaluated on three simulation benchmarks and the real world setting.

Ma~\etal~\cite{ma_vip_2023} propose an objective function for pre-training representation on human videos, and the distance in embedding space can be used for reward scoring. They pre-train an encoder on the Ego4D dataset and test in the simulation environment FrankaKitchen by comparing with CLIP and R3M.
In the following work LIV~\etal~\cite{yecheng_jason_ma2023}, they focus on language-image representation pretraining and propose an objective function for learning the multi-modal pretraining representation based on the EpicKitchen dataset. The authors also claim that this objective supports finetuning on domain-specific dataset and the learned representation can be used along with L2 norm distance function for calculating intermediate reward of a policy.

Different from learning vision-language representation with contrastive learning, Mu~\etal~\cite{mu2023ec2} get inspired from neural language processing and adopt emergent communication scheme to learn multi-modality representation. They pre-train a speaker and listener via an emergent communication game to generate emergent language, which could be considered as a domain-specific latent code. In the emergent communication game, a language model is also trained to predict masked trajectory, so that it can be used for policy generation following videos or language instructions.

\textbf{Modeling Dynamics.}
Modeling dynamics emphasizes capturing the dependencies between sequential states and actions, enabling the prediction of future states and transitions in a task, which is crucial for long-horizon planning and decision-making.
Hafner~\etal~\cite{hafnerdream} propose Dreamer, which represents the state with latent code and generate future state in long-horizon latent imagination. The generated state is integrated in an actor critic RL framework which optimizes a parametric policy by propagating gradients through learned latent dynamics. The Dreamer pipeline has shown success in Artari~\cite{hafnermastering}, Minecraft games and Visual Control tasks~\cite{hafner2023mastering}.
Bauer~\etal~\cite{bauer_doughnet_nodate} extend the concept of latent space dynamics modeling to deformable object manipulation by encoding visual input into latent space and estimating dynamics with auto-regressive mechanisms. Their work goes beyond Dreamer by jointly modeling both geometrical and topological changes, enabling predictions of splitting and merging events in elastoplastic objects. This integration of topological reasoning complements Dreamer's focus on long-horizon latent imagination, demonstrating the versatility of latent space modeling for tasks that require complex structural transformations.

Learning dynamics from scratch without any preliminary knowledge demands substantial interactions with the environment. 
Build upon the success of large-scale pre-training approaches and contrastive learning, researchers adopt self-supervised learning approaches to pre-train the model as a video prediction model~\cite{suraj_nair_00032022,yecheng_jason_ma2023,ma_vip_2023,seo2022reinforcement}. The states, generated by pre-training models, encapsulate the visual and language information and be used as reward or integrated in policy network in downstream tasks.
These video prediction models capture the transition dependencies between sequential states, yet cannot depict the causality that connects these sequential states due to their lack of action modeling.
To solve this problem, Zhang~\etal~\cite{zhang_prelar_nodate} propose an action-conditional pre-training scheme with learnable action representations. 
They design an inverse dynamics encoder to predict action latent states from successive images and align this action latent space with real action representations. This approach bridges the gap between pre-training and fine-tuning, enabling the transfer of learned representations to downstream tasks effectively.

\section{Policy Layer}\label{section:policy_layer}
In robotic manipulation paradigms, the policy layer modules aim to generate the executable actions for the robotic hardware systems given the high-level task commands and observations including visual and proprioceptive measurements. In general, the resultant action space would be either target end-effector pose or a trajectory of the end-effector.
\subsection{Grasp Generation}
\label{subsection:grasp_generaiton}
% \\
% \indent
As the first instance of policy generation in robotic manipulation, the grasp generation class focuses on finding the target end-effector pose, often represented as a 4-D or 6-D coordinate in the robot's space, given inputs including RGB images, depth images or point clouds, meshes, and more recently, implicit shape representations such as Neural Radiance Fields (NeRF)~\cite{wang2024nerf}. These inputs provide rich spatial and visual information about the object to be grasped, allowing models to reason about the geometry, appearance, and affordance of the object.

% These approaches typically rely on various types of sensor inputs to infer optimal grasp strategies. Common input modalities include RGB or depth images, point clouds, meshes, and more recently, implicit shape representations like Neural Radiance Fields (NeRF)~\cite{wang2024nerf}. These inputs provide rich spatial and visual information about the object to be grasped, allowing models to reason about the geometry, appearance, and affordances of the object.

% The output of grasp synthesis models is often a grasp pose, represented as a 4-D or 6-D transformation of the end-effector in the robot's coordinate space. 
In some cases, intermediate representations, such as grasp rectangles in the image space, are used to simplify the problem formulation. Some approaches predict grasp quality as output when the input includes both the observation and candidate poses, enabling the model to assess the likelihood of success for each candidate grasp.

\begin{table*}[ht!]
\centering
\caption{Generative Models for Grasp Generation}
\label{table:grasp_generation}
% \resizebox{0.5\textwidth}{!}{
\begin{tabular}{l|ccc|ccc|cc|ccc|ccc|ccc}
\toprule
\multirow{3}{*}{\textbf{}} & \multicolumn{3}{c|}{\shortstack{\textbf{Generative} \\ \textbf{Framework}}} & 
\multicolumn{3}{c|}{\textbf{Input}} & \multicolumn{2}{c|}{\textbf{Observation}} & \multicolumn{3}{c|}{\textbf{Data}} & 
\multicolumn{3}{c|}{\textbf{Hand Type}} & \multicolumn{3}{c}{\textbf{Others}} \\
\cmidrule{2-18}
 & \rotatebox{90}{Diff} & \rotatebox{90}{VAE} & \rotatebox{90}{Other} & \rotatebox{90}{2D} & \rotatebox{90}{3D} & \rotatebox{90}{Multi} & \rotatebox{90}{Full} & \rotatebox{90}{Partial} & \rotatebox{90}{Real} & \rotatebox{90}{Sim} & \rotatebox{90}{Hybrid} & \rotatebox{90}{Parallel-jaw} & \rotatebox{90}{Multi-fingered} & \rotatebox{90}{Human hand} & \rotatebox{90}{\shortstack{Real-world \\ Experiment}} & \rotatebox{90}{Language-guided} & \rotatebox{90}{Open-source} \\
\midrule
SE(3)-DiF~\cite{urain2023se} & \checkmark &  &  &  & \checkmark &  & \checkmark & \checkmark &  & \checkmark &  & \checkmark &  &  & \checkmark &  & \checkmark \\
Singh et al.~\cite{singh2024constrained} & \checkmark &  &  &  & \checkmark &  & \checkmark &  &  & \checkmark &  & \checkmark &  &  &  &  & \checkmark \\
Diffusion-EDFs~\cite{ryu2024diffusion} & \checkmark &  &  &  & \checkmark &  & \checkmark &  &  & \checkmark &  & \checkmark &  &  & \checkmark &  & \checkmark \\
Vuong et al.~\cite{vuong2024language} & \checkmark &  &  & \checkmark &  &  &  & \checkmark &  & \checkmark &  & \checkmark &  &  & \checkmark & \checkmark & \checkmark \\
Guo et al.~\cite{guo2024precise} & \checkmark &  &  & \checkmark &  &  &  & \checkmark &  &  & \checkmark & \checkmark &  &  & \checkmark &  &  \\
DexDiffuser~\cite{weng2024dexdiffuser} & \checkmark &  &  &  & \checkmark &  &  & \checkmark &  & \checkmark &  &  & \checkmark &  & \checkmark &  & \checkmark \\
DexGraspNet 2.0~\cite{zhang2024dexgraspnet} & \checkmark &  &  &  & \checkmark &  &  & \checkmark &  & \checkmark &  &  & \checkmark &  & \checkmark &  &  \\
Freiberg et al.~\cite{freiberg2024diffusion} & \checkmark &  &  &  & \checkmark &  &  & \checkmark &  & \checkmark &  &  & \checkmark &  & \checkmark &  &  \\
S2HGrasp~\cite{wang2024single} & \checkmark &  &  &  & \checkmark &  &  & \checkmark &  & \checkmark &  &  &  & \checkmark &  &  & \checkmark \\
G-HOP~\cite{ye2024g} & \checkmark &  &  &  & \checkmark &  & \checkmark &  &  & \checkmark &  &  &  & \checkmark &  &  &  \\
GraspNet~\cite{mousavian20196} &  & \checkmark &  &  & \checkmark &  &  & \checkmark &  & \checkmark &  & \checkmark &  &  & \checkmark &  & \checkmark \\
Contact-GraspNet~\cite{Sundermeyer2021ContactGraspNet} &  & \checkmark &  &  & \checkmark &  &  & \checkmark &  & \checkmark &  & \checkmark &  &  & \checkmark &  & \checkmark \\
Wu et al.~\cite{wu2022saga} &  & \checkmark &  &  & \checkmark &  & \checkmark &  &  & \checkmark &  &  &  & \checkmark &  &  & \checkmark \\
Jiang et al.~\cite{jiang2021hand} &  & \checkmark &  &  & \checkmark &  & \checkmark &  &  & \checkmark &  &  &  & \checkmark &  &  & \checkmark \\
GrainGrasp~\cite{zhao2024graingrasp} &  & \checkmark &  &  & \checkmark &  & \checkmark &  &  & \checkmark &  &  &  & \checkmark &  &  & \checkmark \\
Wu et al.~\cite{wu2022learning} &  & \checkmark &  &  & \checkmark &  &  & \checkmark &  & \checkmark &  &  & \checkmark &  &  &  & \checkmark \\
UniDexGrasp~\cite{xu2023unidexgrasp} &  &  & \checkmark &  & \checkmark &  &  & \checkmark &  & \checkmark &  &  & \checkmark &  &  &  & \checkmark \\
GraspAda~\cite{chen2023graspada} &  &  & \checkmark &  &  & \checkmark &  & \checkmark & \checkmark &  &  & \checkmark &  &  & \checkmark &  &  \\
\bottomrule
\end{tabular}
% }
\end{table*}

Learning-based grasp synthesis has been widely studied in the robotic community~\cite{bohg2013data,du2021vision,newbury2023deep}. Various learning models have been proposed to solve the grasp synthesis problem. Deterministic models, for instance, map the input observations directly to a grasp pose or grasp quality score. These models aim to establish a mapping in which a specific observation corresponds to a determined output pose~\cite{Pinto2016Supersizing,qin2020s4g}. While proven effective in many cases, deterministic models may struggle with multi-modal grasp pose distributions, where multiple distinct grasp solutions exist for a single object—essentially mapping a non-injective, non-surjective function.

To address this limitation, generative models have gained popularity in recent years. Unlike deterministic approaches, generative models aim to learn a distribution over grasp poses. By mapping from a complex, multi-modal grasp distribution to simpler, more tractable distributions, generative models can better capture the inherent uncertainty and variety in grasping tasks. This is particularly beneficial for objects with complex shapes, where multiple valid grasp configurations may exist. The ability of generative models to represent such multi-modal distributions makes them well-suited for robotic grasping, as they naturally accommodate the variability and ambiguity present in real-world objects.

In this section, we will focus on generative models in robotic grasping policy synthesis. Specifically, we will discuss recent works that leverage different generative approaches, following the order of variational autoencoders (VAEs), diffusion models, and other generative models.

\textbf{VAE-based methods.}
Variational Autoencoders (VAEs) have emerged as a powerful framework for addressing the inherent complexities and multimodal nature of robotic grasp generation. By learning compact latent representations and sampling from learned distributions, VAEs facilitate the generation of diverse and physically plausible grasps, making them particularly well-suited for tasks that require adaptability to object variability and uncertainty. One notable application of VAEs in grasp synthesis is proposed by Mousavian et al.~\cite{mousavian20196}, which frames grasp generation as a sampling problem. The model employs a VAE to generate a diverse set of stable grasp poses directly from partial 3D point clouds. The latent space captures the multimodal distribution of grasps, allowing for efficient exploration of feasible solutions. Sundermeyer et al.~\cite{Sundermeyer2021ContactGraspNet} introduce a simplified yet effective representation for 6-DOF grasp generation in cluttered environments. Unlike traditional VAEs, the method projects grasp poses into a reduced 4-DoF space by rooting grasp configurations in the object’s observed point cloud. 

Beyond pure robotic grasping, Wu et al.~\cite{wu2022saga} extends the utility of VAEs to human grasp generation, focusing on whole-body interactions. The model employs a Conditional VAE (CVAE) to generate static grasping poses and detailed contact maps jointly. This integration enables the synthesis of realistic and diverse whole-body grasping motions, capturing both dexterous finger movements and full-body dynamics. Jiang et al.~\cite{jiang2021hand} further explores the role of contact consistency in human grasp generation through a novel CVAE architecture. The model ensures mutual consistency between hand-object contact points by introducing tailored loss functions that penalize misalignment between predicted hand configurations and object contact maps. 

In the context of dexterous grasping, Zhao et al.~\cite{zhao2024graingrasp} proposed CVAE framework predicts individual contact maps for each finger, leveraging these detailed representations to optimize grasp configurations. Wu et al.~\cite{wu2022learning} integrates VAE-based sampling with bilevel optimization to address the dual challenges of diversity and physical feasibility. Initial grasp configurations are sampled from the VAE, capturing the multimodal grasp distribution. These samples then seed a bilevel optimization framework that enforces constraints such as collision avoidance, wrench closure, and friction stability. Overall, the integration of VAEs into grasp synthesis underscores their ability to effectively capture the multimodal and uncertain nature of robotic and human grasping tasks by embedding task-specific constraints and leveraging latent-variable modeling.

\textbf{Diffusion-based methods.}
Urain et al.~\cite{urain2023se} utilize an energy-based diffusion model, taking object shape and noisy pose as inputs to output energy as the cost. For grasp pose generation, it employs auto differentiation to compute the score and inverse Langevin dynamics to denoise the 6D pose represented in $se(3)$ space. 
Singh et al.~\cite{singh2024constrained} extend SE(3)-DiF to enable region-constrained grasping by adopting a part-guided diffusion approach, allowing efficient and dense grasp generation on complex object shapes without conditionally labeled datasets.  
To enhance data efficiency and generalizability, Ryu et al.~\cite{ryu2024diffusion} use a bi-equivariant diffusion-based generative model on SE(3), enabling effective grasp pose generation from point cloud observations with minimal demonstrations. 
In addition, Vuong et al.~\cite{vuong2024language} present a language-driven approach for grasp detection, targeting grasp rectangles in image space. It introduces the open-vocabulary grasping dataset and a diffusion model with a contrastive training objective to improve denoising and grasp pose detection from language instructions. 
Guo et al.~\cite{guo2024precise} propose a two-stage diffusion framework that utilizes score-based diffusion networks to generate 4D grasp poses and their corresponding residuals from top-down RGB images, improving the precision of pick-and-place operations in robotic tasks.

The aforementioned works primarily apply diffusion models to two-finger grasping tasks, while recent studies have also extensively explored grasping with multi-finger end-effectors, including dexterous hands. Specifically, Weng et al.~\cite{weng2024dexdiffuser} use a conditional diffusion model based on Denoising Diffusion Probabilistic Models (DDPM)~\cite{ho2020denoising} to generate dexterous grasps on partial point clouds. They further leverage Mayer et al.~\cite{mayer2022ffhnet} for quality assessment, with refinement strategies to enhance success rates. Zhang et al.~\cite{zhang2024dexgraspnet} also employs DDPM to generate dexterous grasp poses in cluttered scenes, while the condition is based on local features extracted from the scene, and the target is a 12D vector representing the wrist pose, combining translation and flattened rotation matrices. 
To generalize the method to different hardware devices, Freiberg et al.~\cite{freiberg2024diffusion} propose a gripper-agnostic approach for grasping using an equivariant diffusion model, which encodes scenes with graspable objects and decodes grasp poses by integrating gripper geometry, demonstrating generalizability across multiple grippers from parallel-jaw to dexterous hands. 

Besides, recent works have also focused on human hand modeling, aiming to achieve more natural and lifelike grasping behaviors. 
Among them, Wang et al.~\cite{wang2024single} proposes a framework for generating human grasps from single-view scene point clouds. They use a Global Perception module to perceive the global shape of partially visible objects and a DiffuGrasp module based on a conditional diffusion model. This model targets hand parameters, progressively denoising them to generate stable grasps. To avoid penetration during training, the model employs a penetration loss that penalizes collisions between the hand and the object, ensuring natural and feasible grasp generation. 
Ye et al.~\cite{ye2024g} turn to develop a denoising diffusion-based generative model that captures the joint 3D distribution of hands and objects during interaction. Given a category-conditioned description, the model can synthesize both plausible object shapes and the relative configuration and articulation of the human hand.

\textbf{Other generative models.}
Xu et al.~\cite{xu2023unidexgrasp} utilize probabilistic models to generate diverse pre-grasp poses from point cloud data, effectively decoupling rotation from translation and articulation. The rotation space is represented by Implicit-PDF~\cite{murphy2021implicit}, a probabilistic model over SO(3), while the conditional distributions of translation and articulation are modeled using a generative flow model~\cite{kingma2018glow}. 
Rather than directly modeling the data distribution, Chen et al.~\cite{chen2023graspada} employ a conditional Generative Adversarial Network (cGAN)~\cite{mirza2014conditional} to adapt RGB-D data from new domains, aligning it with the training domain while preserving grasp feature consistency. By ensuring that image features conform to domain-specific grasp feature distributions, a pretrained grasp synthesis model can then be used to generate grasp poses, including angle, width, and quality score.

\subsection{Trajectory Generation}
\label{subsection:trajectory_generaiton}
% \newline
% \indent
Trajectory generation is another foundational aspect of robotic manipulation, directly impacting a robot’s ability to perform complex tasks accurately, efficiently, and safely. 

Different methods utilize various input formats to produce consistent and stable outputs. Besides using a single-step observation~\cite{kim2024openvla} of the environment—whose patterns resemble those of grasp pose generation—as input for next-action prediction, historical information, including sequential observations~\cite{yang2025octopus,black2024pi_0}, is widely employed to enhance action consistency. Moreover, multi-step action prediction is a common technique for generating stable and coherent trajectories by enabling the model to anticipate a sequence of future actions~\cite{zhao2024alohaunleashedsimplerecipe,yang2025octopus,black2024pi_0}.

From a methodological perspective, traditional trajectory planning methods encounter significant limitations, including high computational demands in high-dimensional configuration spaces, limited adaptability to dynamic environments, insufficient generalization to novel tasks, and challenges in managing long-horizon, multi-stage planning. Generative models present a promising paradigm shift by enabling efficient and direct trajectory generation, improved adaptability through data-driven learning, and enhanced generalization across diverse tasks and environments. These models also support the integration of multi-modal inputs, such as vision and language, while producing smooth and temporally consistent trajectories. Despite these advancements, generative models face critical challenges, including real-time inference constraints, reliance on diverse and high-quality training data, incorporation of physical and dynamic constraints, robustness to uncertainty in real-world applications, and the need for safety and interpretability. Addressing these limitations is pivotal for advancing generative approaches to meet the rigorous demands of robotic manipulation in real-world scenarios.

%TODO: Trajectory  Timeline
\begin{figure*}
  \centering
  \includegraphics[width=0.9\textwidth]{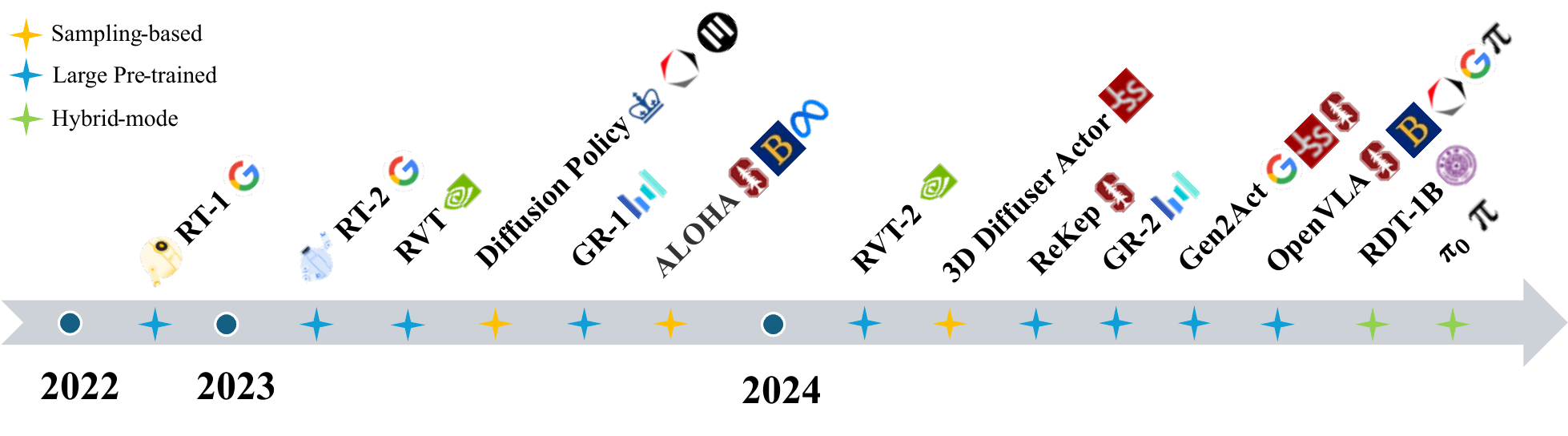}
  \caption{Timeline of representative manipulation trajectory generation works.}
  \label{figure:timeline_trajectory}
\end{figure*}

In this section, we categorize these models into three primary approaches based on their generation strategies: Sampling-based Generation, Large Pre-trained Model-based Generation, and Hybrid-mode Generation. Sampling-based Generation relies on probabilistic models, such as diffusion models or Gaussian Processes, to sample high-quality trajectories from a learned or predefined distribution, effectively handling high-dimensional spaces and complex constraints like obstacle avoidance. Large Pre-trained Model-based Generation leverages powerful pre-trained models, such as vision-language transformers or autoregressive architectures, to predict trajectories conditioned on high-level inputs like language commands or task descriptions, demonstrating strong generalization across diverse tasks and environments. Hybrid-mode Generation combines multiple generative techniques, integrating latent variable representations, pre-trained knowledge, and video dynamics to achieve efficiency and adaptability in complex, multi-stage tasks. Together, these approaches address critical challenges in trajectory generation, from multi-modal integration to long-horizon task planning, advancing the capabilities of robots in dynamic and unstructured environments.

\textbf{Sampling-based Generation}
refers to approaches that generate robotic manipulation trajectories or policies by sampling from a learned or predefined distribution of possible actions or states. These methods often rely on probabilistic models, such as Gaussian Processes, Monte Carlo methods, or diffusion models, to explore the space of feasible solutions and optimize for task-specific objectives. Sampling-based techniques are particularly useful for handling high-dimensional state and action spaces, as well as complex constraints like obstacle avoidance or dynamic environments. 

ALOHA~\cite{fu2024mobile} is designed for fine-grained bimanual manipulation tasks that require high precision and coordination, employing off-the-shelf hardware and custom 3D-printed components to enhance accessibility. Its key innovation, the Action Chunking with Transformers (ACT) algorithm, leverages a conditional variational autoencoder (CVAE) framework, allowing the robot to learn and execute precise tasks through imitation learning with just 10 minutes of demonstration data, achieving high success rates in tasks like battery insertion and cup opening. RoboAgent~\cite{bharadhwaj2024roboagent} introduces MT-ACT, which enhances a small dataset of 7,500 trajectories through semantic augmentations and a novel policy architecture, outperforming baselines in multi-task settings and confirming the effectiveness of its design choices. The Chunking Causal Transformer (CCT)~\cite{zhang2024autoregressiveactionsequencelearning} extends autoregressive models to robotics, efficiently predicting multiple future action tokens and improving accuracy. In contrast, the Autoregressive Policy (ARP) model based on CCT demonstrates superior performance across various tasks. 

The Diffusion Policy~\cite{chi2023diffusion} proposes a novel approach to learning visuomotor policies for robotic manipulation using a conditional diffusion model, framing the policy as a denoising diffusion process that effectively models multimodal action distributions and high-dimensional action sequences, enhancing stability during training and adapting well to tasks requiring temporal consistency. Key innovations include receding horizon control for continuous re-planning, visual conditioning for real-time observations, and a time-series diffusion transformer for managing high-frequency action changes, with empirical results showing a 46.9\% average performance improvement over state-of-the-art methods across 15 tasks. The 3D Diffuser Actor~\cite{3d_diffuser_actor} combines 3D scene representations and action diffusion for robot manipulation policies, utilizing tokenized 3D scene representations, language instructions, and noised robot pose trajectories as input, setting new state-of-the-art results on RLBench~\cite{james2019rlbench} and CALVIN~\cite{mees2022calvin} benchmarks, although it has limitations such as requiring camera calibration and slower performance compared to non-diffusion policies. The Equivariant Diffusion Policy~\cite{wang2024equivariantdiffusionpolicy} leverages equivariant neural models to improve sample efficiency and generalization, showing higher success rates in simulations and real-world experiments, although it has limitations in symmetry matching. EquiBot~\cite{yang2024equibotsim3equivariantdiffusionpolicy} combines SIM(3)-equivariant neural networks with diffusion models to enhance generalization and data efficiency, performing well in both simulation and real-world tasks. Crossway Diffusion~\cite{li2024crosswaydiffusionimprovingdiffusionbased} improves diffusion-based visuomotor policy by integrating an auxiliary self-supervised objective, demonstrating consistent improvements over baseline methods. The Consistency Policy~\cite{prasad2024consistency} addresses slow inference in diffusion models, achieving significant speedup while retaining performance, although it has drawbacks in multimodality. Lastly, ALOHA Unleashed~\cite{zhao2024alohaunleashedsimplerecipe} utilizes large-scale data collection with a Transformer-based architecture and diffusion policies to tackle challenging bimanual manipulation tasks, demonstrating superior performance and laying the groundwork for future multi-task learning models. 

Addressing the challenges of multi-stage bimanual manipulation tasks, which demand efficient arm coordination and face issues like multi-modal demonstrations and per-step/per-stage reliability, BiKC~\cite{yu2024bikckeyposeconditionedconsistencypolicy} is a hierarchical IL framework with a high-level keypose predictor and a low-level trajectory generator. The key pose predictor identifies key poses using heuristics and a merging method for bimanual coordination and is trained to predict target key poses. The trajectory generator is a consistency model to generate action sequences conditioned on observations and key poses. Experiments on simulated and real-world tasks (Transfer, Insertion, Screwdriver Packing, etc.) show BiKC outperforms baselines (ACT, DP) in success rate and operational efficiency, while also demonstrating multi-modal modeling and inference speed advantages, though it has limitations like sample smoothness and key pose representation that could be improved in future research.

\textbf{Large Pre-trained Model-based Generation}
refers to approaches that leverage large-scale pre-trained models, such as Transformers or GPT-style architectures, to generate robotic manipulation policies or trajectories. These models are typically trained on massive datasets, often incorporating multimodal inputs like vision, language, and sensor data, enabling them to learn rich representations and generalize across diverse tasks. By fine-tuning on specific robotic tasks, these models can generate high-quality, context-aware actions or plans, even in complex and dynamic environments.  Vision-Language-Action (VLA) models regard actions as a set of language tokens, LLMs to generate actions and language simultaneously. RT-2~\cite{brohan2023rt} introduced the VLA concept and developed two VLA models based on PaLI-X~\cite{chen2023pali} and PaLM-E~\cite{driess2023palm}. VLA models demonstrate enhanced generalization performance, leveraging the extensive pre-training knowledge of LLMs. 

RT-1~\cite{brohan2023rt1roboticstransformerrealworld} and RT-2 utilize Transformer-based architectures for robust performance in various tasks, while RVT~\cite{goyal2023rvtroboticviewtransformer} and RVT-2~\cite{goyal2024rvt2learningprecisemanipulation} focus on multi-view transformations to improve task execution efficiency and generalization in real-world applications. As RT-2 is not open-source, OpenVLA~\cite{kim2024openvla} was proposed as a publicly accessible alternative. OpenVLA is built upon the Prismatic multi-modality LLM~\cite{karamcheti2024prismatic} and utilizes DINOv2~\cite{oquab2023dinov2} and SigLIP~\cite{zhai2023sigmoid} as visual encoders. It directly outputs action tokens using the LLM. Pre-trained on the large-scale Open-X-Embodiment dataset~\cite{o2023open}, OpenVLA exhibits superior performance compared to RT-2. DeL-TaCo~\cite{yu2023usingdemonstrationslanguageinstructions} conditions robotic policies on task embeddings from visual demonstrations and language instructions, improving generalization and reducing the need for demonstrations. PERACT~\cite{shridhar2022peract} is a language-conditioned behavior-cloning agent for multi-task manipulation, outperforming traditional methods, while PolarNet~\cite{chen2023polarnet3dpointclouds} integrates point cloud representations with language instructions, showing robustness in real-world tasks. 

Im2Flow2Act\cite{xu2024flow} enables robots to learn manipulation skills from diverse data using object flow, while Dreamitate\cite{liang2024dreamitate} leverages video diffusion models for robust visuomotor policy learning. Gen2Act~\cite{bharadhwaj2024gen2act} combines zero-shot human video generation with limited robot demos, achieving high success rates in generalization. ReKep~\cite{huang2024rekep} employs relational keypoint constraints for task representation, while Surfer~\cite{ren2024surferprogressivereasoningworld} decouples manipulation into action and scene prediction. The introduction of GemBench~\cite{garcia2024generalizablevisionlanguageroboticmanipulation} provides a benchmark for evaluating generalization in vision-language robotic manipulation. GR-1~\cite{wu2023unleashing} and GR-2~\cite{cheang2024gr-2} explore large-scale video generative pre-training to enhance visual robot manipulation, achieving state-of-the-art results. 

\textbf{Hybrid-mode Generation}
refers to approaches that combine multiple generative techniques, such as latent variable models, pre-trained models, and video generation models, to enhance the robustness and versatility of robotic manipulation algorithms. By integrating the strengths of different methods, these approaches can address complex tasks that require both high-level planning and low-level control, while also handling multimodal inputs like vision, language, and sensor data. GenDP~\cite{wang2024gendp3dsemanticfields} framework addresses generalization limitations of diffusion-based policies by using 3D semantic fields generated from multi-view RGBD observations, significantly improving success rates on unseen instances and enabling category-level generalization. The Hierarchical Diffusion Policy (HDP)~\cite{ma2024hierarchical} factors the policy into a hierarchical structure for efficient visual manipulation, outperforming state-of-the-art methods and demonstrating the importance of kinematics awareness, while facing challenges in long-horizon tasks. A novel approach called Large Language model Reinforcement Learning Policy (LLaRP)~\cite{szot2024largelanguagemodelsgeneralizable} adapts large language models (LLMs) to function as generalizable policies for embodied visual tasks, leveraging pre-trained LLMs through reinforcement learning to process text instructions and visual observations, achieving a 42\% success rate on 1,000 new tasks and introducing the Language Rearrangement benchmark for evaluating language-conditioned rearrangement tasks. The Robotics Diffusion Transformer (RDT)~\cite{liu2024rdt1bdiffusionfoundationmodel} uses diffusion models to handle multi-modal action distributions with a scalable Transformer architecture, proposing a Physically Interpretable Unified Action Space to train on heterogeneous multi-robot data. Pre-trained on a large multi-robot dataset and fine-tuned on a self-collected bimanual dataset, RDT shows strong zero-shot generalization, effective instruction following, and few-shot learning capabilities in experiments on a real robot. Finally, $\pi_0$~\cite{black2024pi_0} presents a novel architecture for generalist robot control, leveraging a pre-trained vision-language model (VLM) as its backbone and integrating semantic understanding from vast datasets with a flow matching approach to generate continuous actions, achieving high-frequency control (up to 50 Hz). Trained on a diverse dataset across different robots and tasks, $\pi_0$ enables zero-shot task performance and fine-tuning for complex tasks like folding laundry and assembling objects, positioning it as a significant step towards versatile, general-purpose robot learning models comparable to large language models but focused on physically situated, multi-stage robotic tasks.

\begin{table*}[ht]
\centering
\caption{TOP 10 Robot Manipulation Works on RLBench(18 tasks, 100 demo/task).}
\begin{tabular}{clclllllc}
\toprule
Rank & Model & GenAI & Framework & Successful Rate & Training Time & Inference Speed &Input  Image Size & Year \\
\midrule
1 & APR+~\cite{zhang2024autoregressiveactionsequencelearning}                   &\checkmark & Autoregressive & 86.0 & - & - & - & 2024 \\
2 & 3D-LOTUS~\cite{garcia2024generalizablevisionlanguageroboticmanipulation}    &\checkmark & Other          & 81.5 & 0.28  hours & 9.5 fps & 256 $\times$ 256 & 2024 \\
3 & RVT-2~\cite{goyal2024rvt2learningprecisemanipulation}                       &\checkmark & Other & 81.4   & 0.83  hours & 20.6 fps & 128 $\times$ 128 & 2024 \\
4 & 3D Diffuser Actor~\cite{3d_diffuser_actor}                                  &\checkmark & Diffusion      & 81.3 & - & - & 256 $\times$ 256 & 2024 \\
5 & Act3D~\cite{gervet2023act3d}                                                &\checkmark & Other          & 65.0 & 5  hours & - & 256 $\times$ 256 & 2023 \\
6 & RVT~\cite{goyal2023rvtroboticviewtransformer}                               &\checkmark & Other          & 62.9 & 1  hours & 11.6 fps & 128 $\times$ 128 & 2023 \\
7 & PerAct~\cite{shridhar2022peract}$^*$                                        &\checkmark & Other          & 49.4 & 16  hours & 4.9 fps & 128 $\times$ 128 & 2022 \\
8 & PolarNet~\cite{chen2023polarnet3dpointclouds}                               &\checkmark & Other          & 46.4 & - & - & 128 $\times$ 128 & 2023 \\
9 & PerAct~\cite{shridhar2022peract}                                            &\checkmark & Other          & 42.7 & 16  hours & - & 128 $\times$ 128 & 2022 \\
10& C2FARM-BC~\cite{james2022coarse}$^*$                                        &$\times$   & -              & 20.1 & - & - & 128 $\times$ 128 & 2021 \\

\bottomrule
&* (Evaluated in RVT)
\end{tabular}
\label{tab:robot_manipulation_on_RLBench}
\end{table*}

Each generative approach provides distinct benefits based on the operational requirements of different robotic tasks. Most of the research has conducted extensive validation on the RLBench benchmark, as illustrated in the table~\ref{tab:robot_manipulation_on_RLBench}. Latent space models offer smooth trajectory generation, favoring tasks where continuity and consistency of movement are critical. Sampling-based models are optimal where trajectory accuracy and control are paramount, particularly in structured environments with predictable obstacles. Hybrid generation in achieving a balanced trade-off, enabling versatile solutions across various robotic manipulation tasks. Generative models for robotic trajectory generation represent a trade-off between precision, adaptability, and efficiency.

\section{Future Directions}
\label{section:future_direction}
As generative models continue to revolutionize robotic manipulation, several promising avenues emerge to address existing challenges and unlock new capabilities. First, grounding domain-general models to domain-specific constraints through hybrid explicit-implicit representation learning and task-centric adaptation, as exemplified by Robotic-CLIP~\cite{nguyen2024robotic} and ManiGaussian~\cite{lu2025manigaussian}, will enable robots to inherit visual generalization while encoding action-specific dynamics. Second, establishing unified benchmarks with standardized tasks, datasets, and evaluation metrics is essential to facilitate fair comparisons and reproducibility across studies. Finally, enhancing physical law awareness through physics-augmented training, physics-constrained generation, and reinforcement learning with physical feedback will bridge the sim-to-real gap and improve generalization to real-world environments. By addressing these challenges, the field can achieve robust, adaptive, and efficient robotic manipulation systems capable of operating in complex, unstructured settings.

\paragraph{Grounding domain-general models to domain-specific data}

One fundamental challenge in applying generative AI to robotic manipulation lies in grounding domain-general models to domain-specific constraints. While vision-language models (VLMs)~\cite{radford2021learning, kirillov2023segment}  demonstrate remarkable zero-shot generalization for object recognition and segmentation tasks, their representations lack crucial physical interaction features, like spatial relationships, force dynamics. This limitation stems from their training objectives being centered on static internet-source data rather than dynamic physical interactions.

Two critical issues compound this challenge. First, semantic-physical representation misalignment occurs when vision-language models (VLMs) optimized for web-scale visual understanding fail to encode robotic-relevant attributes. For instance, the image-text alignment in CLIP-based architectures~\cite{radford2021learning} struggles with dynamic action semantics in manipulation videos, necessitating temporal-aware model structures and domain-specific fine-tuning on action sequences to bridge this gap. Second, data-induced physical bias arises from discrepancies between internet-scale pretraining data and real-world physical laws. This manifests in catastrophic operational failures when VLMs misinterpret material properties: robots may mistakenly classify soft rubber as rigid metal due to visual similarities in surface reflectivity, leading to inappropriate grasping strategies (e.g., excessive grip forces). Empirical evidence from VLBiasBench~\cite{zhang2024vlbiasbench} reveals that VLMs achieve only 62\% accuracy in distinguishing "metal vs. plastic" materials, with a 78\% tendency to misclassify reflective surfaces as metals while ignoring mechanical characteristics like stiffness or compliance. Such biases originate from VLMs' overreliance on static web imagery lacking physical interaction dynamics and multimodal annotations aligned with material science principles.

To overcome these barriers, two emerging research directions show particular promise.  
On the one hand, hybrid explicit-implicit representation learning combines neural fields with foundation models to dynamically encode both geometric precision and task semantics. Recent extensions like ManiGaussian further demonstrate how Gaussian Splatting can regularize 3D Gaussian primitives for learning dynamics with visual foundation features~\cite{lu2025manigaussian}.  
On the other hand, task-centric adaptation with interaction-aware datasets has gained traction. Robotic-CLIP~\cite{nguyen2024robotic} pioneers this paradigm by fine-tuning CLIP on 7.4 million action frames through contrastive learning, preserving original semantic alignment while significantly improving performance in language-driven robotic tasks. It introduces an Adapter Network that efficiently maps action-specific dynamics into the CLIP embedding space without altering the original model weights, enabling parameter-efficient adaptation. Experimental results demonstrate that Robotic-CLIP outperforms other CLIP variants in tasks such as language-driven grasp detection and policy learning, achieving higher success rates and better alignment between textual instructions and visual frames. This approach enables robots to inherit visual generalization while encoding action-specific dynamics through parameter-efficient adaptation.

\paragraph{Fragmented benchmarks hinder progress and fair comparisons}
The lack of unified benchmarks in embodied manipulation algorithms poses a significant challenge, hindering fair performance comparisons and reproducibility across studies. Current issues include diverse task definitions~\cite{james2019rlbench,fu2024mobile,chi2023diffusion,3d_diffuser_actor}, inconsistent datasets~\cite{james2019rlbench, mees2022calvin,o2023open}, varying simulation platforms~\cite{6386109,coumans2016pybullet,makoviychuk2021isaac}, and non-standardized evaluation metrics. To address this, future efforts should focus on establishing standardized tasks, datasets, and simulation environments, while developing comprehensive evaluation frameworks that incorporate success rate, completion time, energy efficiency, and generalization ability. Additionally, promoting open-source code, reproducibility challenges, and cross-platform compatibility will enhance transparency and collaboration within the research community. By unifying benchmarks, the field can accelerate progress and enable more reliable advancements in embodied manipulation algorithms. 

\paragraph{Limited physical law awareness}
Limited physical law awareness is a significant challenge in embodied manipulation, particularly in bridging the sim2real gap and leveraging visual generation models. Many current approaches rely on simplified or incomplete representations of physical laws, such as friction, dynamics, and material properties, which leads to poor generalization from simulation to real-world environments. Sim-to-real gap means the difference in performance and behavior when transitioning from simulation (sim) to the real world (real). This gap can lead to models that work well in simulation failing to perform as expected in reality. It is caused by physics simulation limitations, sensor discrepancies between simulation and real world, actuator variability, environmental complexity, and unmodeled dynamics in the real world~\cite{salvato2021crossing, pitkevich2024survey, zhao2020sim}. Some methods have been proposed to address this issue. Domain randomization is a common approach to apply randomizing physical parameters, lighting conditions, sensor noise in the simulation~\cite{peng2018sim, garcia2023robust, huber2024domain}. Sim-to-real transfer learning and domain adaptation techniques are also useful to address the sim-to-real gap~\cite{james2019sim, liu2023digital, chen2022sim}. Privileged distillation via teacher-student learning shows that learning to behave like the teacher model which is exposed to more information is also effective to address the  sim-to-real gap problem~\cite{he2024bridging, chen2020learning, nguyen2022leveraging}. While visual generation models, particularly in the domain of robotic manipulation, have made significant strides in producing realistic images and videos, they often lack a deep understanding of the underlying physical laws that govern real-world interactions~\cite{kang2024far,bansal2024VideoPhy,wang2024VAMP,wu2024Pastnet}. These models primarily focus on generating visually plausible sequences based on learned patterns from large datasets, but they are typically not trained with explicit awareness of concepts like gravity, momentum, collision dynamics, or material properties. As a result, the generated scenes may not always align with the constraints of the physical world, leading to unrealistic or physically infeasible visual cues. This limited physical awareness can degrade the quality of planning in robotic manipulation tasks. Consequently, robots relying on these models may struggle with tasks involving delicate object manipulation, dynamic interactions, or long-horizon planning, where understanding the consequences of physical actions is critical for success. To address this limitation, several avenues for future research could be explored, including 1) physics-augmented training that incorporates physical simulations to fine-tune generative models, 2) physics-constrained generation that incorporates physical constraints into the generation process, and 3) reinforcement learning with physical feedback that utilizes explicit physical feedback to improve the physical law awareness.

\section{Conclusion}
\label{section:conclusion}
Generative models have emerged as powerful tools to address the critical challenges in robotic manipulation, including insufficient data, long-horizon task planning, and generalization across diverse environments. By leveraging paradigms such as GANs, VAEs, probabilistic flow models, diffusion models, and autoregressive models, researchers have made significant strides in enabling robots to generate data, infer states, and synthesize policies for grasping and trajectory planning. This survey categorizes these applications into foundational, intermediate, and policy layers, providing a structured perspective on their roles in data generation, language and visual modeling, and robotic control. Despite the progress, challenges such as improving data efficiency, handling complex tasks, and achieving robust generalization remain open problems. Future research should focus on developing scalable models, integrating multi-modal data sources, and enhancing real-time decision-making capabilities to push the boundaries of robotic intelligence. These efforts will pave the way for more adaptable, versatile, and autonomous robotic systems capable of tackling increasingly complex and dynamic environments.

{\small
\bibliography{ref.bib}

% Generated by IEEEtranN.bst, version: 1.14 (2015/08/26)
\begin{thebibliography}{253}
\providecommand{\natexlab}[1]{#1}
\providecommand{\url}[1]{#1}
\csname url@samestyle\endcsname
\providecommand{\newblock}{\relax}
\providecommand{\bibinfo}[2]{#2}
\providecommand{\BIBentrySTDinterwordspacing}{\spaceskip=0pt\relax}
\providecommand{\BIBentryALTinterwordstretchfactor}{4}
\providecommand{\BIBentryALTinterwordspacing}{\spaceskip=\fontdimen2\font plus
\BIBentryALTinterwordstretchfactor\fontdimen3\font minus \fontdimen4\font\relax}
\providecommand{\BIBforeignlanguage}[2]{{%
\expandafter\ifx\csname l@#1\endcsname\relax
\typeout{** WARNING: IEEEtranN.bst: No hyphenation pattern has been}%
\typeout{** loaded for the language `#1'. Using the pattern for}%
\typeout{** the default language instead.}%
\else
\language=\csname l@#1\endcsname
\fi
#2}}
\providecommand{\BIBdecl}{\relax}
\BIBdecl

\bibitem[Billard and Kragic(2019)]{Billard2019}
A.~Billard and D.~Kragic, ``Trends and challenges in robot manipulation,'' \emph{Science}, vol. 364, no. 6446, p. eaat8414, 2019.

\bibitem[Mason(2018)]{Mason2018}
M.~T. Mason, ``Toward robotic manipulation,'' \emph{Annual Review of Control, Robotics, and Autonomous Systems}, vol.~1, pp. 1--28, 2018.

\bibitem[Firoozi et~al.(2023)Firoozi, Tucker, Tian, Majumdar, Sun, Liu, Zhu, Song, Kapoor, Hausman, et~al.]{firoozi2023foundation}
R.~Firoozi, J.~Tucker, S.~Tian, A.~Majumdar, J.~Sun, W.~Liu, Y.~Zhu, S.~Song, A.~Kapoor, K.~Hausman \emph{et~al.}, ``Foundation models in robotics: Applications, challenges, and the future,'' \emph{The International Journal of Robotics Research}, p. 02783649241281508, 2023.

\bibitem[Tobin et~al.(2017)Tobin, Fong, Ray, Schneider, Zaremba, and Abbeel]{Tobin2017DomainRandomization}
J.~Tobin, R.~Fong, A.~Ray, J.~Schneider, W.~Zaremba, and P.~Abbeel, ``Domain randomization for transferring deep neural networks from simulation to the real world,'' in \emph{2017 IEEE/RSJ International Conference on Intelligent Robots and Systems (IROS)}.\hskip 1em plus 0.5em minus 0.4em\relax IEEE, 2017, pp. 23--30.

\bibitem[Wang et~al.(2023{\natexlab{a}})Wang, Xian, Chen, Wang, Wang, Fragkiadaki, Erickson, Held, and Gan]{wang2023robogen}
Y.~Wang, Z.~Xian, F.~Chen, T.-H. Wang, Y.~Wang, K.~Fragkiadaki, Z.~Erickson, D.~Held, and C.~Gan, ``Robogen: Towards unleashing infinite data for automated robot learning via generative simulation,'' \emph{arXiv preprint arXiv:2311.01455}, 2023.

\bibitem[Chi et~al.(2023)Chi, Xu, Feng, Cousineau, Du, Burchfiel, Tedrake, and Song]{chi2023diffusion}
C.~Chi, Z.~Xu, S.~Feng, E.~Cousineau, Y.~Du, B.~Burchfiel, R.~Tedrake, and S.~Song, ``Diffusion policy: Visuomotor policy learning via action diffusion,'' \emph{The International Journal of Robotics Research}, p. 02783649241273668, 2023.

\bibitem[Ke et~al.(2024)Ke, Gkanatsios, and Fragkiadaki]{3d_diffuser_actor}
T.-W. Ke, N.~Gkanatsios, and K.~Fragkiadaki, ``3d diffuser actor: Policy diffusion with 3d scene representations,'' \emph{Arxiv}, 2024.

\bibitem[Sundermeyer et~al.(2021)Sundermeyer, Mousavian, Triebel, and Fox]{Sundermeyer2021ContactGraspNet}
M.~Sundermeyer, A.~Mousavian, R.~Triebel, and D.~Fox, ``Contact-graspnet: Efficient 6-dof grasp generation in cluttered scenes,'' in \emph{2021 IEEE International Conference on Robotics and Automation (ICRA)}, 2021, pp. 13\,438--13\,444.

\bibitem[Hafner et~al.({\natexlab{a}})Hafner, Lillicrap, Ba, and Norouzi]{hafnerdream}
D.~Hafner, T.~Lillicrap, J.~Ba, and M.~Norouzi, ``Dream to control: Learning behaviors by latent imagination,'' in \emph{International Conference on Learning Representations}.

\bibitem[Bauer et~al.()Bauer, Xu, and Song]{bauer_doughnet_nodate}
D.~Bauer, Z.~Xu, and S.~Song, ``{DoughNet}: A visual predictive model for topological manipulation of deformable objects.''

\bibitem[Levine et~al.(2016)Levine, Finn, Darrell, and Abbeel]{Levine2016EndToEnd}
S.~Levine, C.~Finn, T.~Darrell, and P.~Abbeel, ``End-to-end training of deep visuomotor policies,'' \emph{Journal of Machine Learning Research}, vol.~17, no.~39, pp. 1--40, 2016.

\bibitem[Zhu et~al.(2018)Zhu, Wang, Merel, Rusu, Erez, Cabi, Tunyasuvunakool, Kramár, Hadsell, de~Freitas, and Heess]{Zhu2018Reinforcement}
\BIBentryALTinterwordspacing
Y.~Zhu, Z.~Wang, J.~Merel, A.~Rusu, T.~Erez, S.~Cabi, S.~Tunyasuvunakool, J.~Kramár, R.~Hadsell, N.~de~Freitas, and N.~Heess, ``Reinforcement and imitation learning for diverse visuomotor skills,'' 2018. [Online]. Available: \url{https://arxiv.org/abs/1802.09564}
\BIBentrySTDinterwordspacing

\bibitem[Pinto and Gupta(2016)]{Pinto2016Supersizing}
L.~Pinto and A.~Gupta, ``Supersizing self-supervision: Learning to grasp from 50k tries and 700 robot hours,'' in \emph{2016 IEEE International Conference on Robotics and Automation (ICRA)}.\hskip 1em plus 0.5em minus 0.4em\relax IEEE, 2016, pp. 3406--3413.

\bibitem[James et~al.(2019)James, Wohlhart, Kalakrishnan, Kalashnikov, Irpan, Ibarz, Levine, Hadsell, and Bousmalis]{james2019sim}
S.~James, P.~Wohlhart, M.~Kalakrishnan, D.~Kalashnikov, A.~Irpan, J.~Ibarz, S.~Levine, R.~Hadsell, and K.~Bousmalis, ``Sim-to-real via sim-to-sim: Data-efficient robotic grasping via randomized-to-canonical adaptation networks,'' in \emph{Proceedings of the IEEE/CVF conference on computer vision and pattern recognition}, 2019, pp. 12\,627--12\,637.

\bibitem[Rusu et~al.(2017)Rusu, Vecerik, Roth{\"o}rl, Heess, Pascanu, and Hadsell]{Rusu2017SimToReal}
A.~A. Rusu, M.~Vecerik, T.~Roth{\"o}rl, N.~Heess, R.~Pascanu, and R.~Hadsell, ``Sim-to-real robot learning from pixels with progressive nets,'' in \emph{Conference on Robot Learning}.\hskip 1em plus 0.5em minus 0.4em\relax PMLR, 2017, pp. 262--270.

\bibitem[Zhu et~al.(2017)Zhu, Mottaghi, Kolve, Lim, Gupta, Fei-Fei, and Farhadi]{Zhu2017Target}
Y.~Zhu, R.~Mottaghi, E.~Kolve, J.~J. Lim, A.~Gupta, L.~Fei-Fei, and A.~Farhadi, ``Target-driven visual navigation in indoor scenes using deep reinforcement learning,'' in \emph{2017 IEEE International Conference on Robotics and Automation (ICRA)}.\hskip 1em plus 0.5em minus 0.4em\relax IEEE, 2017, pp. 3357--3364.

\bibitem[Rombach et~al.(2022)Rombach, Blattmann, Lorenz, Esser, and Ommer]{Rombach2022HighResolution}
R.~Rombach, A.~Blattmann, D.~Lorenz, P.~Esser, and B.~Ommer, ``High-resolution image synthesis with latent diffusion models,'' in \emph{Proceedings of the IEEE/CVF Conference on Computer Vision and Pattern Recognition (CVPR)}, June 2022, pp. 10\,684--10\,695.

\bibitem[Brown(2020)]{brown2020language}
T.~B. Brown, ``Language models are few-shot learners,'' \emph{arXiv preprint arXiv:2005.14165}, 2020.

\bibitem[Nikolaidis and Shah(2015)]{Nikolaidis2015Efficient}
S.~Nikolaidis and J.~A. Shah, ``Efficient model learning from joint-action demonstrations for human-robot collaborative tasks,'' in \emph{Tenth Annual ACM/IEEE International Conference on Human-Robot Interaction}.\hskip 1em plus 0.5em minus 0.4em\relax IEEE, 2015, pp. 189--196.

\bibitem[Kaelbling and Lozano-P{\'e}rez(2011)]{Kaelbling2011Hierarchical}
L.~P. Kaelbling and T.~Lozano-P{\'e}rez, ``Hierarchical task and motion planning in the now,'' in \emph{IEEE International Conference on Robotics and Automation}.\hskip 1em plus 0.5em minus 0.4em\relax IEEE, 2011, pp. 1470--1477.

\bibitem[Neuberg(2003)]{Pearl2009Causality}
L.~G. Neuberg, ``Causality: Models, reasoning, and inference, by judea pearl, cambridge university press, 2000,'' \emph{Econometric Theory}, vol.~19, no.~4, p. 675–685, 2003.

\bibitem[Toussaint(2015)]{Toussaint2015Logic}
M.~Toussaint, ``Logic-geometric programming: An optimization-based approach to combined task and motion planning,'' in \emph{Twenty-Fourth International Joint Conference on Artificial Intelligence}, 2015, pp. 1930--1936.

\bibitem[Wei et~al.(2022)Wei, Wang, Schuurmans, Bosma, Xia, Chi, Le, Zhou, et~al.]{wei2022chain}
J.~Wei, X.~Wang, D.~Schuurmans, M.~Bosma, F.~Xia, E.~Chi, Q.~V. Le, D.~Zhou \emph{et~al.}, ``Chain-of-thought prompting elicits reasoning in large language models,'' \emph{Advances in neural information processing systems}, vol.~35, pp. 24\,824--24\,837, 2022.

\bibitem[Ahn et~al.(2022)Ahn, Brohan, Brown, Chebotar, Cortes, David, Finn, Fu, Gopalakrishnan, Hausman, Herzog, Ho, Hsu, Ibarz, Ichter, Irpan, Jang, Ruano, Jeffrey, Jesmonth, Joshi, Julian, Kalashnikov, Kuang, Lee, Levine, Lu, Luu, Parada, Pastor, Quiambao, Rao, Rettinghouse, Reyes, Sermanet, Sievers, Tan, Toshev, Vanhoucke, Xia, Xiao, Xu, Xu, Yan, and Zeng]{Ahn2022DoAsICan}
\BIBentryALTinterwordspacing
M.~Ahn, A.~Brohan, N.~Brown, Y.~Chebotar, O.~Cortes, B.~David, C.~Finn, C.~Fu, K.~Gopalakrishnan, K.~Hausman, A.~Herzog, D.~Ho, J.~Hsu, J.~Ibarz, B.~Ichter, A.~Irpan, E.~Jang, R.~J. Ruano, K.~Jeffrey, S.~Jesmonth, N.~J. Joshi, R.~Julian, D.~Kalashnikov, Y.~Kuang, K.-H. Lee, S.~Levine, Y.~Lu, L.~Luu, C.~Parada, P.~Pastor, J.~Quiambao, K.~Rao, J.~Rettinghouse, D.~Reyes, P.~Sermanet, N.~Sievers, C.~Tan, A.~Toshev, V.~Vanhoucke, F.~Xia, T.~Xiao, P.~Xu, S.~Xu, M.~Yan, and A.~Zeng, ``Do as i can, not as i say: Grounding language in robotic affordances,'' 2022. [Online]. Available: \url{https://arxiv.org/abs/2204.01691}
\BIBentrySTDinterwordspacing

\bibitem[Huang et~al.(2022{\natexlab{a}})Huang, Abbeel, Pathak, and Mordatch]{huang2022language}
W.~Huang, P.~Abbeel, D.~Pathak, and I.~Mordatch, ``Language models as zero-shot planners: Extracting actionable knowledge for embodied agents,'' in \emph{International conference on machine learning}.\hskip 1em plus 0.5em minus 0.4em\relax PMLR, 2022, pp. 9118--9147.

\bibitem[Finn et~al.(2016)Finn, Goodfellow, and Levine]{Finn2016Unsupervised}
C.~Finn, I.~Goodfellow, and S.~Levine, ``Unsupervised learning for physical interaction through video prediction,'' \emph{Advances in neural information processing systems}, vol.~29, 2016.

\bibitem[Ebert et~al.(2018)Ebert, Finn, Dasari, Xie, Lee, and Levine]{Ebert2018VisualForesight}
F.~Ebert, C.~Finn, S.~Dasari, A.~Xie, A.~Lee, and S.~Levine, ``Visual foresight: Model-based deep reinforcement learning for vision-based robotic control,'' \emph{arXiv preprint arXiv:1812.00568}, 2018.

\bibitem[Ha and Schmidhuber(2018)]{Ha2018WorldModels}
D.~Ha and J.~Schmidhuber, ``World models,'' \emph{arXiv preprint arXiv:1803.10122}, 2018.

\bibitem[Urain et~al.(2023)Urain, Funk, Peters, and Chalvatzaki]{urain2023se}
J.~Urain, N.~Funk, J.~Peters, and G.~Chalvatzaki, ``Se (3)-diffusionfields: Learning smooth cost functions for joint grasp and motion optimization through diffusion,'' in \emph{2023 IEEE International Conference on Robotics and Automation (ICRA)}.\hskip 1em plus 0.5em minus 0.4em\relax IEEE, 2023, pp. 5923--5930.

\bibitem[Singh et~al.(2024)Singh, Kalwar, Karim, Sen, Govindan, Sridhar, and Krishna]{singh2024constrained}
G.~Singh, S.~Kalwar, M.~F. Karim, B.~Sen, N.~Govindan, S.~Sridhar, and K.~M. Krishna, ``Constrained 6-dof grasp generation on complex shapes for improved dual-arm manipulation,'' \emph{arXiv preprint arXiv:2404.04643}, 2024.

\bibitem[Fu et~al.(2024{\natexlab{a}})Fu, Zhao, and Finn]{fu2024mobile}
Z.~Fu, T.~Z. Zhao, and C.~Finn, ``Mobile aloha: Learning bimanual mobile manipulation with low-cost whole-body teleoperation,'' in \emph{{Conference on Robot Learning (CoRL)}}, 2024.

\bibitem[Hu et~al.(2024)Hu, Xie, Jain, Francis, Patrikar, Keetha, Kim, Xie, Zhang, Fang, Zhao, Omidshafiei, Kim, akbar Agha-mohammadi, Sycara, Johnson-Roberson, Batra, Wang, Scherer, Wang, Kira, Xia, and Bisk]{hu_toward_2024}
\BIBentryALTinterwordspacing
Y.~Hu, Q.~Xie, V.~Jain, J.~Francis, J.~Patrikar, N.~Keetha, S.~Kim, Y.~Xie, T.~Zhang, H.-S. Fang, S.~Zhao, S.~Omidshafiei, D.-K. Kim, A.~akbar Agha-mohammadi, K.~Sycara, M.~Johnson-Roberson, D.~Batra, X.~Wang, S.~Scherer, C.~Wang, Z.~Kira, F.~Xia, and Y.~Bisk, ``Toward general-purpose robots via foundation models: A survey and meta-analysis,'' 2024. [Online]. Available: \url{https://arxiv.org/abs/2312.08782}
\BIBentrySTDinterwordspacing

\bibitem[McCarthy et~al.(2024)McCarthy, Tan, Schmidt, Acero, Herr, Du, Thuruthel, and Li]{mccarthy_towards_2024}
\BIBentryALTinterwordspacing
R.~McCarthy, D.~C.~H. Tan, D.~Schmidt, F.~Acero, N.~Herr, Y.~Du, T.~G. Thuruthel, and Z.~Li, ``Towards generalist robot learning from internet video: A survey,'' 2024. [Online]. Available: \url{https://arxiv.org/abs/2404.19664}
\BIBentrySTDinterwordspacing

\bibitem[Goodfellow et~al.(2014)Goodfellow, Pouget-Abadie, Mirza, Xu, Warde-Farley, Ozair, Courville, and Bengio]{goodfellow2014generative}
I.~Goodfellow, J.~Pouget-Abadie, M.~Mirza, B.~Xu, D.~Warde-Farley, S.~Ozair, A.~Courville, and Y.~Bengio, ``Generative adversarial nets,'' \emph{Advances in neural information processing systems}, vol.~27, 2014.

\bibitem[Chen et~al.(2023{\natexlab{a}})Chen, Jiang, Lei, Bekiroglu, Chen, and Li]{chen2023graspada}
Y.~Chen, J.~Jiang, R.~Lei, Y.~Bekiroglu, F.~Chen, and M.~Li, ``Graspada: Deep grasp adaptation through domain transfer,'' in \emph{2023 IEEE International Conference on Robotics and Automation (ICRA)}.\hskip 1em plus 0.5em minus 0.4em\relax IEEE, 2023, pp. 10\,268--10\,274.

\bibitem[Kingma(2013)]{kingma2013auto}
D.~P. Kingma, ``Auto-encoding variational bayes,'' \emph{arXiv preprint arXiv:1312.6114}, 2013.

\bibitem[Mousavian et~al.(2019)Mousavian, Eppner, and Fox]{mousavian20196}
A.~Mousavian, C.~Eppner, and D.~Fox, ``6-dof graspnet: Variational grasp generation for object manipulation,'' in \emph{Proceedings of the IEEE/CVF international conference on computer vision}, 2019, pp. 2901--2910.

\bibitem[Ho et~al.(2020)Ho, Jain, and Abbeel]{ho2020denoising}
J.~Ho, A.~Jain, and P.~Abbeel, ``Denoising diffusion probabilistic models,'' \emph{Advances in neural information processing systems}, vol.~33, pp. 6840--6851, 2020.

\bibitem[Song et~al.(2020)Song, Meng, and Ermon]{song2020denoising}
J.~Song, C.~Meng, and S.~Ermon, ``Denoising diffusion implicit models,'' \emph{arXiv preprint arXiv:2010.02502}, 2020.

\bibitem[Rezende and Mohamed(2015)]{rezende2015variational}
D.~Rezende and S.~Mohamed, ``Variational inference with normalizing flows,'' in \emph{International conference on machine learning}.\hskip 1em plus 0.5em minus 0.4em\relax PMLR, 2015, pp. 1530--1538.

\bibitem[Mantegazza et~al.(2022)Mantegazza, Giusti, Gambardella, and Guzzi]{mantegazza2022outlier}
D.~Mantegazza, A.~Giusti, L.~M. Gambardella, and J.~Guzzi, ``An outlier exposure approach to improve visual anomaly detection performance for mobile robots,'' \emph{IEEE Robotics and Automation Letters}, vol.~7, no.~4, pp. 11\,354--11\,361, 2022.

\bibitem[Brockmann et~al.(2023)Brockmann, Rudolph, Rosenhahn, and Wandt]{brockmann2023voraus}
J.~T. Brockmann, M.~Rudolph, B.~Rosenhahn, and B.~Wandt, ``The voraus-ad dataset for anomaly detection in robot applications,'' \emph{IEEE Transactions on Robotics}, 2023.

\bibitem[Wellhausen et~al.(2020)Wellhausen, Ranftl, and Hutter]{wellhausen2020safe}
L.~Wellhausen, R.~Ranftl, and M.~Hutter, ``Safe robot navigation via multi-modal anomaly detection,'' \emph{IEEE Robotics and Automation Letters}, vol.~5, no.~2, pp. 1326--1333, 2020.

\bibitem[Xu et~al.(2023)Xu, Wan, Zhang, Liu, Shan, Shen, Wang, Geng, Weng, Chen, et~al.]{xu2023unidexgrasp}
Y.~Xu, W.~Wan, J.~Zhang, H.~Liu, Z.~Shan, H.~Shen, R.~Wang, H.~Geng, Y.~Weng, J.~Chen \emph{et~al.}, ``Unidexgrasp: Universal robotic dexterous grasping via learning diverse proposal generation and goal-conditioned policy,'' in \emph{Proceedings of the IEEE/CVF Conference on Computer Vision and Pattern Recognition}, 2023, pp. 4737--4746.

\bibitem[Ouyang et~al.(2022)Ouyang, Wu, Jiang, Almeida, Wainwright, Mishkin, Zhang, Agarwal, Slama, Ray, et~al.]{ouyang2022training}
L.~Ouyang, J.~Wu, X.~Jiang, D.~Almeida, C.~Wainwright, P.~Mishkin, C.~Zhang, S.~Agarwal, K.~Slama, A.~Ray \emph{et~al.}, ``Training language models to follow instructions with human feedback,'' \emph{Advances in neural information processing systems}, vol.~35, pp. 27\,730--27\,744, 2022.

\bibitem[Radford(2018)]{radford2018improving}
A.~Radford, ``Improving language understanding by generative pre-training,'' 2018.

\bibitem[Radford et~al.(2019)Radford, Wu, Child, Luan, Amodei, Sutskever, et~al.]{radford2019language}
A.~Radford, J.~Wu, R.~Child, D.~Luan, D.~Amodei, I.~Sutskever \emph{et~al.}, ``Language models are unsupervised multitask learners,'' \emph{OpenAI blog}, vol.~1, no.~8, p.~9, 2019.

\bibitem[Achiam et~al.(2023)Achiam, Adler, Agarwal, Ahmad, Akkaya, Aleman, Almeida, Altenschmidt, Altman, Anadkat, et~al.]{achiam2023gpt}
J.~Achiam, S.~Adler, S.~Agarwal, L.~Ahmad, I.~Akkaya, F.~L. Aleman, D.~Almeida, J.~Altenschmidt, S.~Altman, S.~Anadkat \emph{et~al.}, ``Gpt-4 technical report,'' \emph{arXiv preprint arXiv:2303.08774}, 2023.

\bibitem[Chen et~al.(2020{\natexlab{a}})Chen, Radford, Child, Wu, Jun, Luan, and Sutskever]{chen2020generative}
M.~Chen, A.~Radford, R.~Child, J.~Wu, H.~Jun, D.~Luan, and I.~Sutskever, ``Generative pretraining from pixels,'' in \emph{International conference on machine learning}.\hskip 1em plus 0.5em minus 0.4em\relax PMLR, 2020, pp. 1691--1703.

\bibitem[Ramesh et~al.(2021)Ramesh, Pavlov, Goh, Gray, Voss, Radford, Chen, and Sutskever]{ramesh2021zero}
A.~Ramesh, M.~Pavlov, G.~Goh, S.~Gray, C.~Voss, A.~Radford, M.~Chen, and I.~Sutskever, ``Zero-shot text-to-image generation,'' in \emph{International conference on machine learning}.\hskip 1em plus 0.5em minus 0.4em\relax Pmlr, 2021, pp. 8821--8831.

\bibitem[Sun et~al.(2024)Sun, Jiang, Chen, Zhang, Peng, Luo, and Yuan]{sun2024autoregressive}
P.~Sun, Y.~Jiang, S.~Chen, S.~Zhang, B.~Peng, P.~Luo, and Z.~Yuan, ``Autoregressive model beats diffusion: Llama for scalable image generation,'' \emph{arXiv preprint arXiv:2406.06525}, 2024.

\bibitem[Tian et~al.(2024)Tian, Jiang, Yuan, Peng, and Wang]{tian2024visual}
K.~Tian, Y.~Jiang, Z.~Yuan, B.~Peng, and L.~Wang, ``Visual autoregressive modeling: Scalable image generation via next-scale prediction,'' \emph{arXiv preprint arXiv:2404.02905}, 2024.

\bibitem[Kaplan et~al.(2020)Kaplan, McCandlish, Henighan, Brown, Chess, Child, Gray, Radford, Wu, and Amodei]{kaplan2020scaling}
J.~Kaplan, S.~McCandlish, T.~Henighan, T.~B. Brown, B.~Chess, R.~Child, S.~Gray, A.~Radford, J.~Wu, and D.~Amodei, ``Scaling laws for neural language models,'' \emph{arXiv preprint arXiv:2001.08361}, 2020.

\bibitem[Brohan et~al.(2023{\natexlab{a}})Brohan, Brown, Carbajal, Chebotar, Chen, Choromanski, Ding, Driess, Dubey, Finn, et~al.]{brohan2023rt}
A.~Brohan, N.~Brown, J.~Carbajal, Y.~Chebotar, X.~Chen, K.~Choromanski, T.~Ding, D.~Driess, A.~Dubey, C.~Finn \emph{et~al.}, ``Rt-2: Vision-language-action models transfer web knowledge to robotic control,'' \emph{arXiv preprint arXiv:2307.15818}, 2023.

\bibitem[Kim et~al.(2024)Kim, Pertsch, Karamcheti, Xiao, Balakrishna, Nair, Rafailov, Foster, Lam, Sanketi, et~al.]{kim2024openvla}
M.~J. Kim, K.~Pertsch, S.~Karamcheti, T.~Xiao, A.~Balakrishna, S.~Nair, R.~Rafailov, E.~Foster, G.~Lam, P.~Sanketi \emph{et~al.}, ``Openvla: An open-source vision-language-action model,'' \emph{arXiv preprint arXiv:2406.09246}, 2024.

\bibitem[Coumans and Bai(2016)]{coumans2016pybullet}
E.~Coumans and Y.~Bai, ``Pybullet, a python module for physics simulation for games, robotics and machine learning,'' 2016.

\bibitem[Todorov et~al.(2012)Todorov, Erez, and Tassa]{6386109}
E.~Todorov, T.~Erez, and Y.~Tassa, ``Mujoco: A physics engine for model-based control,'' in \emph{2012 IEEE/RSJ International Conference on Intelligent Robots and Systems}, 2012.

\bibitem[Rohmer et~al.()Rohmer, Singh, and Freese]{rohmerversatile}
E.~Rohmer, S.~Singh, and M.~C. Freese, ``A versatile and scalable robot simulation framework,'' in \emph{Proceedings of the International Conference on Intelligent Robots and Systems (IROS)}.

\bibitem[Zhang et~al.(2023)Zhang, Du, Shan, Zhou, Du, Tenenbaum, Shu, and Gan]{zhang2023building}
H.~Zhang, W.~Du, J.~Shan, Q.~Zhou, Y.~Du, J.~B. Tenenbaum, T.~Shu, and C.~Gan, ``Building cooperative embodied agents modularly with large language models,'' \emph{arXiv preprint arXiv:2307.02485}, 2023.

\bibitem[Ramos et~al.(2019)Ramos, Possas, and Fox]{ramos2019bayessim}
F.~Ramos, R.~C. Possas, and D.~Fox, ``Bayessim: adaptive domain randomization via probabilistic inference for robotics simulators,'' \emph{arXiv preprint arXiv:1906.01728}, 2019.

\bibitem[Lyu et~al.(2024)Lyu, Bai, Yang, Lu, and Li]{lyu2024cross}
J.~Lyu, C.~Bai, J.~Yang, Z.~Lu, and X.~Li, ``Cross-domain policy adaptation by capturing representation mismatch,'' \emph{arXiv preprint arXiv:2405.15369}, 2024.

\bibitem[Li et~al.(2024{\natexlab{a}})Li, Jin, Yu, Shi, Hao, Hao, Liu, Sun, Zhang, Fang, et~al.]{li2024foundation}
D.~Li, Y.~Jin, H.~Yu, J.~Shi, X.~Hao, P.~Hao, H.~Liu, F.~Sun, J.~Zhang, B.~Fang \emph{et~al.}, ``What foundation models can bring for robot learning in manipulation: A survey,'' \emph{arXiv preprint arXiv:2404.18201}, 2024.

\bibitem[Wang et~al.(2024{\natexlab{a}})Wang, Chen, Huang, Ben, Wang, Mi, Huang, Zhao, Chen, Yang, et~al.]{wang2024grutopia}
H.~Wang, J.~Chen, W.~Huang, Q.~Ben, T.~Wang, B.~Mi, T.~Huang, S.~Zhao, Y.~Chen, S.~Yang \emph{et~al.}, ``Grutopia: Dream general robots in a city at scale,'' \emph{arXiv preprint arXiv:2407.10943}, 2024.

\bibitem[Freiberg et~al.(2024)Freiberg, Qualmann, Vien, and Neumann]{freiberg2024diffusion}
R.~Freiberg, A.~Qualmann, N.~A. Vien, and G.~Neumann, ``Diffusion for multi-embodiment grasping,'' \emph{arXiv preprint arXiv:2410.18835}, 2024.

\bibitem[Katara et~al.(2024)Katara, Xian, and Fragkiadaki]{katara2024gen2sim}
P.~Katara, Z.~Xian, and K.~Fragkiadaki, ``Gen2sim: Scaling up robot learning in simulation with generative models,'' in \emph{2024 IEEE International Conference on Robotics and Automation (ICRA)}.\hskip 1em plus 0.5em minus 0.4em\relax IEEE, 2024, pp. 6672--6679.

\bibitem[Yang et~al.(2024{\natexlab{a}})Yang, Sun, Weihs, VanderBilt, Herrasti, Han, Wu, Haber, Krishna, Liu, et~al.]{yang2024holodeck}
Y.~Yang, F.-Y. Sun, L.~Weihs, E.~VanderBilt, A.~Herrasti, W.~Han, J.~Wu, N.~Haber, R.~Krishna, L.~Liu \emph{et~al.}, ``Holodeck: Language guided generation of 3d embodied ai environments,'' in \emph{Proceedings of the IEEE/CVF Conference on Computer Vision and Pattern Recognition}, 2024, pp. 16\,227--16\,237.

\bibitem[Brooks et~al.(2024)Brooks, Peebles, Holmes, DePue, Guo, Jing, Schnurr, Taylor, Luhman, Luhman, et~al.]{brooks2024video}
T.~Brooks, B.~Peebles, C.~Holmes, W.~DePue, Y.~Guo, L.~Jing, D.~Schnurr, J.~Taylor, T.~Luhman, E.~Luhman \emph{et~al.}, ``Video generation models as world simulators,'' 2024.

\bibitem[Yang et~al.(2023)Yang, Du, Ghasemipour, Tompson, Schuurmans, and Abbeel]{yang2023learning}
M.~Yang, Y.~Du, K.~Ghasemipour, J.~Tompson, D.~Schuurmans, and P.~Abbeel, ``Learning interactive real-world simulators,'' \emph{arXiv preprint arXiv:2310.06114}, 2023.

\bibitem[Mandlekar et~al.(2023)Mandlekar, Nasiriany, Wen, Akinola, Narang, Fan, Zhu, and Fox]{mandlekar2023mimicgen}
A.~Mandlekar, S.~Nasiriany, B.~Wen, I.~Akinola, Y.~Narang, L.~Fan, Y.~Zhu, and D.~Fox, ``Mimicgen: A data generation system for scalable robot learning using human demonstrations,'' \emph{arXiv preprint arXiv:2310.17596}, 2023.

\bibitem[Wang et~al.(2024{\natexlab{b}})Wang, Qin, Kuang, Korkmaz, Gurumoorthy, Su, and Wang]{wang2024cyberdemo}
J.~Wang, Y.~Qin, K.~Kuang, Y.~Korkmaz, A.~Gurumoorthy, H.~Su, and X.~Wang, ``Cyberdemo: Augmenting simulated human demonstration for real-world dexterous manipulation,'' in \emph{Proceedings of the IEEE/CVF Conference on Computer Vision and Pattern Recognition}, 2024, pp. 17\,952--17\,963.

\bibitem[Zhang et~al.(2024{\natexlab{a}})Zhang, Chang, Kumar, and Gupta]{zhang2024diffusion}
X.~Zhang, M.~Chang, P.~Kumar, and S.~Gupta, ``Diffusion meets dagger: Supercharging eye-in-hand imitation learning,'' \emph{arXiv preprint arXiv:2402.17768}, 2024.

\bibitem[Jiang et~al.(2024)Jiang, Xie, Lin, Xu, Wan, Mandlekar, Fan, and Zhu]{jiang2024dexmimicgen}
Z.~Jiang, Y.~Xie, K.~Lin, Z.~Xu, W.~Wan, A.~Mandlekar, L.~Fan, and Y.~Zhu, ``Dexmimicgen: Automated data generation for bimanual dexterous manipulation via imitation learning,'' \emph{arXiv preprint arXiv:2410.24185}, 2024.

\bibitem[Hoque et~al.()Hoque, Mandlekar, Garrett, Goldberg, and Fox]{hoque2023interventional}
R.~Hoque, A.~Mandlekar, C.~R. Garrett, K.~Goldberg, and D.~Fox, ``Interventional data generation for robust and data-efficient robot imitation learning,'' in \emph{First Workshop on Out-of-Distribution Generalization in Robotics at CoRL 2023}.

\bibitem[Ha et~al.(2023)Ha, Florence, and Song]{ha2023scaling}
H.~Ha, P.~Florence, and S.~Song, ``Scaling up and distilling down: Language-guided robot skill acquisition,'' in \emph{Conference on Robot Learning}.\hskip 1em plus 0.5em minus 0.4em\relax PMLR, 2023, pp. 3766--3777.

\bibitem[Chen et~al.(2023{\natexlab{b}})Chen, Kiami, Gupta, and Kumar]{chen2023genaug}
Z.~Chen, S.~Kiami, A.~Gupta, and V.~Kumar, ``Genaug: Retargeting behaviors to unseen situations via generative augmentation,'' \emph{arXiv preprint arXiv:2302.06671}, 2023.

\bibitem[Yu et~al.(2023{\natexlab{a}})Yu, Xiao, Stone, Tompson, Brohan, Wang, Singh, Tan, Peralta, Ichter, et~al.]{yu2023scaling}
T.~Yu, T.~Xiao, A.~Stone, J.~Tompson, A.~Brohan, S.~Wang, J.~Singh, C.~Tan, J.~Peralta, B.~Ichter \emph{et~al.}, ``Scaling robot learning with semantically imagined experience,'' \emph{arXiv preprint arXiv:2302.11550}, 2023.

\bibitem[Chen et~al.(2024{\natexlab{a}})Chen, Xu, Dharmarajan, Irshad, Cheng, Keutzer, Tomizuka, Vuong, and Goldberg]{chen2024rovi}
L.~Y. Chen, C.~Xu, K.~Dharmarajan, M.~Z. Irshad, R.~Cheng, K.~Keutzer, M.~Tomizuka, Q.~Vuong, and K.~Goldberg, ``Rovi-aug: Robot and viewpoint augmentation for cross-embodiment robot learning,'' \emph{arXiv preprint arXiv:2409.03403}, 2024.

\bibitem[Chen et~al.(2022{\natexlab{a}})Chen, Van~Wyk, Chao, Yang, Mousavian, Gupta, and Fox]{chen2022learning}
Z.~Q. Chen, K.~Van~Wyk, Y.-W. Chao, W.~Yang, A.~Mousavian, A.~Gupta, and D.~Fox, ``Learning robust real-world dexterous grasping policies via implicit shape augmentation,'' \emph{arXiv preprint arXiv:2210.13638}, 2022.

\bibitem[Kapelyukh et~al.(2023)Kapelyukh, Vosylius, and Johns]{kapelyukh2023dall}
I.~Kapelyukh, V.~Vosylius, and E.~Johns, ``Dall-e-bot: Introducing web-scale diffusion models to robotics,'' \emph{IEEE Robotics and Automation Letters}, vol.~8, no.~7, pp. 3956--3963, 2023.

\bibitem[Lepert et~al.()Lepert, Doshi, and Bohg]{lepertshadow}
M.~Lepert, R.~Doshi, and J.~Bohg, ``Shadow: Leveraging segmentation masks for cross-embodiment policy transfer,'' in \emph{8th Annual Conference on Robot Learning}.

\bibitem[Bahl et~al.(2022)Bahl, Gupta, and Pathak]{bahl2022human}
S.~Bahl, A.~Gupta, and D.~Pathak, ``Human-to-robot imitation in the wild,'' \emph{arXiv preprint arXiv:2207.09450}, 2022.

\bibitem[Chen et~al.(2024{\natexlab{b}})Chen, Hari, Dharmarajan, Xu, Vuong, and Goldberg]{chen2024mirage}
L.~Y. Chen, K.~Hari, K.~Dharmarajan, C.~Xu, Q.~Vuong, and K.~Goldberg, ``Mirage: Cross-embodiment zero-shot policy transfer with cross-painting,'' \emph{arXiv preprint arXiv:2402.19249}, 2024.

\bibitem[Mandi et~al.(2022)Mandi, Bharadhwaj, Moens, Song, Rajeswaran, and Kumar]{mandi2022cacti}
Z.~Mandi, H.~Bharadhwaj, V.~Moens, S.~Song, A.~Rajeswaran, and V.~Kumar, ``Cacti: A framework for scalable multi-task multi-scene visual imitation learning,'' \emph{arXiv preprint arXiv:2212.05711}, 2022.

\bibitem[Bharadhwaj et~al.(2024{\natexlab{a}})Bharadhwaj, Vakil, Sharma, Gupta, Tulsiani, and Kumar]{bharadhwaj2024roboagent}
H.~Bharadhwaj, J.~Vakil, M.~Sharma, A.~Gupta, S.~Tulsiani, and V.~Kumar, ``Roboagent: Generalization and efficiency in robot manipulation via semantic augmentations and action chunking,'' in \emph{2024 IEEE International Conference on Robotics and Automation (ICRA)}.\hskip 1em plus 0.5em minus 0.4em\relax IEEE, 2024, pp. 4788--4795.

\bibitem[Hirose et~al.(2023)Hirose, Shah, Sridhar, and Levine]{hirose2023exaug}
N.~Hirose, D.~Shah, A.~Sridhar, and S.~Levine, ``Exaug: Robot-conditioned navigation policies via geometric experience augmentation,'' in \emph{2023 IEEE International Conference on Robotics and Automation (ICRA)}.\hskip 1em plus 0.5em minus 0.4em\relax IEEE, 2023, pp. 4077--4084.

\bibitem[Dai et~al.(2024)Dai, Lee, Fazeli, and Chai]{dai2024racer}
Y.~Dai, J.~Lee, N.~Fazeli, and J.~Chai, ``Racer: Rich language-guided failure recovery policies for imitation learning,'' \emph{arXiv preprint arXiv:2409.14674}, 2024.

\bibitem[Xiao et~al.(2022)Xiao, Chan, Sermanet, Wahid, Brohan, Hausman, Levine, and Tompson]{xiao2022robotic}
T.~Xiao, H.~Chan, P.~Sermanet, A.~Wahid, A.~Brohan, K.~Hausman, S.~Levine, and J.~Tompson, ``Robotic skill acquisition via instruction augmentation with vision-language models,'' \emph{arXiv preprint arXiv:2211.11736}, 2022.

\bibitem[Ma et~al.(2023{\natexlab{a}})Ma, Liang, Wang, Huang, Bastani, Jayaraman, Zhu, Fan, and Anandkumar]{yechengeureka_jason_ma2023}
Y.~J. Ma, W.~Liang, G.~Wang, D.-A. Huang, O.~Bastani, D.~Jayaraman, Y.~Zhu, L.~Fan, and A.~Anandkumar, ``Eureka: Human-level reward design via coding large language models.'' in \emph{CoRR}, 2023.

\bibitem[Du et~al.(2023)Du, Konyushkova, Denil, Raju, Landon, Hill, de~Freitas, and Cabi]{yuqing_du2023}
Y.~Du, K.~Konyushkova, M.~Denil, A.~Raju, J.~Landon, F.~Hill, N.~de~Freitas, and S.~Cabi, ``Vision-language models as success detectors.'' in \emph{CoRR}, 2023.

\bibitem[Ge et~al.()Ge, Macaluso, Li, Luo, and Wang]{ge_policy_2023}
\BIBentryALTinterwordspacing
Y.~Ge, A.~Macaluso, L.~E. Li, P.~Luo, and X.~Wang, ``Policy adaptation from foundation model feedback,'' in \emph{2023 {IEEE}/{CVF} Conference on Computer Vision and Pattern Recognition ({CVPR})}.\hskip 1em plus 0.5em minus 0.4em\relax {IEEE}, pp. 19\,059--19\,069. [Online]. Available: \url{https://ieeexplore.ieee.org/document/10204201/}
\BIBentrySTDinterwordspacing

\bibitem[He et~al.(2020)He, 0001, Wu, Xie, and Girshick]{kaiming_he2020}
K.~He, H.~F. 0001, Y.~Wu, S.~Xie, and R.~B. Girshick, ``Momentum contrast for unsupervised visual representation learning.'' in \emph{CVPR}, 2020.

\bibitem[Wang et~al.(2021)Wang, Zhang, Shen, Kong, and 0005]{xinlong_wang2021}
X.~Wang, R.~Zhang, C.~Shen, T.~Kong, and L.~L. 0005, ``Dense contrastive learning for self-supervised visual pre-training.'' in \emph{CVPR}, 2021.

\bibitem[Kraus et~al.(2023)Kraus, Kenyon-Dean, Saberian, Fallah, McLean, Leung, Sharma, Khan, Balakrishnan, Celik, Sypetkowski, Cheng, Morse, Makes, Mabey, and Earnshaw]{oren_kraus2023}
O.~Kraus, K.~Kenyon-Dean, S.~Saberian, M.~Fallah, P.~McLean, J.~Leung, V.~Sharma, A.~Khan, J.~Balakrishnan, S.~Celik, M.~Sypetkowski, C.~V. Cheng, K.~Morse, M.~Makes, B.~Mabey, and B.~Earnshaw, ``Masked autoencoders are scalable learners of cellular morphology.'' in \emph{CoRR}, 2023.

\bibitem[Li et~al.(2022)Li, Zhang, Zhang, Yang, Li, Zhong, Wang, Yuan, 0001, Hwang, Chang, and 0001]{liunian_harold_li2022}
L.~H. Li, P.~Zhang, H.~Zhang, J.~Yang, C.~Li, Y.~Zhong, L.~Wang, L.~Yuan, L.~Z. 0001, J.-N. Hwang, K.-W. Chang, and J.~G. 0001, ``Grounded language-image pre-training.'' in \emph{CVPR}, 2022.

\bibitem[Radford et~al.(2021)Radford, Kim, Hallacy, Ramesh, Goh, Agarwal, Sastry, Askell, Mishkin, Clark, et~al.]{radford2021learning}
A.~Radford, J.~W. Kim, C.~Hallacy, A.~Ramesh, G.~Goh, S.~Agarwal, G.~Sastry, A.~Askell, P.~Mishkin, J.~Clark \emph{et~al.}, ``Learning transferable visual models from natural language supervision,'' in \emph{International conference on machine learning}.\hskip 1em plus 0.5em minus 0.4em\relax PMLR, 2021, pp. 8748--8763.

\bibitem[Kamath et~al.(2021)Kamath, Singh, LeCun, Misra, Synnaeve, and Carion]{aishwarya_kamath2021}
A.~Kamath, M.~Singh, Y.~LeCun, I.~Misra, G.~Synnaeve, and N.~Carion, ``Mdetr - modulated detection for end-to-end multi-modal understanding.'' in \emph{CoRR}, 2021.

\bibitem[Vuong et~al.(2024)Vuong, Vu, Huang, Nguyen, Le, Vo, and Nguyen]{vuong2024language}
A.~D. Vuong, M.~N. Vu, B.~Huang, N.~Nguyen, H.~Le, T.~Vo, and A.~Nguyen, ``Language-driven grasp detection,'' in \emph{Proceedings of the IEEE/CVF Conference on Computer Vision and Pattern Recognition}, 2024, pp. 17\,902--17\,912.

\bibitem[Wang et~al.(2022)Wang, Yang, Men, Lin, Bai, Li, Ma, Zhou, Zhou, and Yang]{wang2022ofa}
P.~Wang, A.~Yang, R.~Men, J.~Lin, S.~Bai, Z.~Li, J.~Ma, C.~Zhou, J.~Zhou, and H.~Yang, ``Ofa: Unifying architectures, tasks, and modalities through a simple sequence-to-sequence learning framework,'' in \emph{International conference on machine learning}.\hskip 1em plus 0.5em minus 0.4em\relax PMLR, 2022, pp. 23\,318--23\,340.

\bibitem[Kirillov et~al.(2023)Kirillov, Mintun, Ravi, Mao, Rolland, Gustafson, Xiao, Whitehead, Berg, Lo, et~al.]{kirillov2023segment}
A.~Kirillov, E.~Mintun, N.~Ravi, H.~Mao, C.~Rolland, L.~Gustafson, T.~Xiao, S.~Whitehead, A.~C. Berg, W.-Y. Lo \emph{et~al.}, ``Segment anything,'' in \emph{Proceedings of the IEEE/CVF International Conference on Computer Vision}, 2023, pp. 4015--4026.

\bibitem[Sun et~al.(2023)Sun, Chen, Zhu, Xiao, 0002, Xie, and Yan]{peize_sun2023}
P.~Sun, S.~Chen, C.~Zhu, F.~Xiao, P.~L. 0002, S.~Xie, and Z.~Yan, ``Going denser with open-vocabulary part segmentation.'' in \emph{CoRR}, 2023.

\bibitem[Cao et~al.(2023)Cao, Zhou, Guo, Zhang, Liu, and Tan]{cao2023nbmod}
B.~Cao, X.~Zhou, C.~Guo, B.~Zhang, Y.~Liu, and Q.~Tan, ``Nbmod: Find it and grasp it in noisy background,'' \emph{arXiv preprint arXiv:2306.10265}, 2023.

\bibitem[Blank et~al.(2024)Blank, Reuss, R{\"u}hle, Ya{\u{g}}murlu, Wenzel, Mees, and Lioutikov]{blank2024scaling}
N.~Blank, M.~Reuss, M.~R{\"u}hle, {\"O}.~E. Ya{\u{g}}murlu, F.~Wenzel, O.~Mees, and R.~Lioutikov, ``Scaling robot policy learning via zero-shot labeling with foundation models,'' \emph{arXiv preprint arXiv:2410.17772}, 2024.

\bibitem[Sumers et~al.()Sumers, Marino, Ahuja, Fergus, and Dasgupta]{sumers_distilling_2023}
\BIBentryALTinterwordspacing
T.~Sumers, K.~Marino, A.~Ahuja, R.~Fergus, and I.~Dasgupta, ``Distilling internet-scale vision-language models into embodied agents,'' in \emph{Proceedings of the 40th International Conference on Machine Learning}.\hskip 1em plus 0.5em minus 0.4em\relax {PMLR}, pp. 32\,797--32\,818, {ISSN}: 2640-3498. [Online]. Available: \url{https://proceedings.mlr.press/v202/sumers23a.html}
\BIBentrySTDinterwordspacing

\bibitem[Andrychowicz et~al.(2017)Andrychowicz, Wolski, Ray, Schneider, Fong, Welinder, McGrew, Tobin, Pieter~Abbeel, and Zaremba]{andrychowicz2017hindsight}
M.~Andrychowicz, F.~Wolski, A.~Ray, J.~Schneider, R.~Fong, P.~Welinder, B.~McGrew, J.~Tobin, O.~Pieter~Abbeel, and W.~Zaremba, ``Hindsight experience replay,'' \emph{Advances in neural information processing systems}, vol.~30, 2017.

\bibitem[Alayrac et~al.(2022)Alayrac, Donahue, and Others]{baptiste_alayrac2022}
J.-B. Alayrac, J.~Donahue, and Others, ``Flamingo: a visual language model for few-shot learning.'' in \emph{CoRR}, 2022.

\bibitem[Xie et~al.(2023)Xie, Zhao, Wu, Liu, Luo, Zhong, Yang, and 0009]{tianbao_xie2023}
T.~Xie, S.~Zhao, C.~H. Wu, Y.~Liu, Q.~Luo, V.~Zhong, Y.~Yang, and T.~Y. 0009, ``Text2reward: Automated dense reward function generation for reinforcement learning.'' in \emph{CoRR}, 2023.

\bibitem[Fu et~al.(2024{\natexlab{b}})Fu, Zhang, 0044, Xu, and Boulet]{yuwei_fu2024}
Y.~Fu, H.~Zhang, D.~W. 0044, W.~Xu, and B.~Boulet, ``Furl: Visual-language models as fuzzy rewards for reinforcement learning.'' in \emph{CoRR}, 2024.

\bibitem[Alakuijala et~al.(2022)Alakuijala, Dulac-Arnold, Mairal, Ponce, and Schmid]{minttu_alakuijala2022}
M.~Alakuijala, G.~Dulac-Arnold, J.~Mairal, J.~Ponce, and C.~Schmid, ``Learning reward functions for robotic manipulation by observing humans.'' in \emph{CoRR}, 2022.

\bibitem[Ma et~al.(2023{\natexlab{b}})Ma, Liang, Som, Kumar, 0001, Bastani, and Jayaraman]{yecheng_jason_ma2023}
Y.~J. Ma, W.~Liang, V.~Som, V.~Kumar, A.~Z. 0001, O.~Bastani, and D.~Jayaraman, ``Liv: Language-image representations and rewards for robotic control.'' in \emph{CoRR}, 2023.

\bibitem[Hu et~al.(2023)Hu, Keloth, Raja, 0016, and 0001]{yan_hu2023}
Y.~Hu, V.~K. Keloth, K.~Raja, Y.~C. 0016, and H.~X. 0001, ``Towards precise pico extraction from abstracts of randomized controlled trials using a section-specific learning approach.'' in \emph{Bioinform.}, 2023.

\bibitem[0003 et~al.(2022)0003, Rajeswaran, Kumar, Finn, and 0001]{suraj_nair_00032022}
S.~N. 0003, A.~Rajeswaran, V.~Kumar, C.~Finn, and A.~G. 0001, ``R3m: A universal visual representation for robot manipulation.'' in \emph{CoRL}, 2022.

\bibitem[Shridhar et~al.(2021)Shridhar, Manuelli, and Fox]{mohit_shridhar2021}
M.~Shridhar, L.~Manuelli, and D.~Fox, ``Cliport: What and where pathways for robotic manipulation.'' in \emph{CoRR}, 2021.

\bibitem[Zhao et~al.(2024{\natexlab{a}})Zhao, Chen, Meng, Mao, Song, and Zhang]{zhao2024vlmpc}
W.~Zhao, J.~Chen, Z.~Meng, D.~Mao, R.~Song, and W.~Zhang, ``Vlmpc: Vision-language model predictive control for robotic manipulation,'' \emph{arXiv preprint arXiv:2407.09829}, 2024.

\bibitem[Finn and Levine(2017)]{finn2017deep}
C.~Finn and S.~Levine, ``Deep visual foresight for planning robot motion,'' in \emph{2017 IEEE International Conference on Robotics and Automation (ICRA)}.\hskip 1em plus 0.5em minus 0.4em\relax IEEE, 2017, pp. 2786--2793.

\bibitem[Yu et~al.(2023{\natexlab{b}})Yu, Gileadi, Fu, Kirmani, Lee, Arenas, Chiang, Erez, Hasenclever, Humplik, Ichter, Xiao, Xu, 0001, Zhang, Heess, Sadigh, Tan, Tassa, and Xia]{yu_2023_arxiv}
W.~Yu, N.~Gileadi, C.~Fu, S.~Kirmani, K.-H. Lee, M.~G. Arenas, H.-T.~L. Chiang, T.~Erez, L.~Hasenclever, J.~Humplik, B.~Ichter, T.~Xiao, P.~Xu, A.~Z. 0001, T.~Zhang, N.~Heess, D.~Sadigh, J.~Tan, Y.~Tassa, and F.~Xia, ``Language to rewards for robotic skill synthesis.'' in \emph{CoRR}, 2023.

\bibitem[Zeng et~al.(2024)Zeng, 0001, and 0002]{yuwei_zeng2024}
Y.~Zeng, Y.~M. 0001, and L.~S. 0002, ``Learning reward for robot skills using large language models via self-alignment.'' in \emph{CoRR}, 2024.

\bibitem[Ke et~al.(2021)Ke, Choudhury, Barnes, Sun, Lee, and Srinivasa]{ke2021imitation}
L.~Ke, S.~Choudhury, M.~Barnes, W.~Sun, G.~Lee, and S.~Srinivasa, ``Imitation learning as f-divergence minimization,'' in \emph{Algorithmic Foundations of Robotics XIV: Proceedings of the Fourteenth Workshop on the Algorithmic Foundations of Robotics 14}.\hskip 1em plus 0.5em minus 0.4em\relax Springer, 2021, pp. 313--329.

\bibitem[Zhong et~al.()Zhong, Misra, Yuan, and Côté]{zhong_policy_2024}
\BIBentryALTinterwordspacing
V.~Zhong, D.~Misra, X.~Yuan, and M.-A. Côté, ``Policy improvement using language feedback models.'' [Online]. Available: \url{http://arxiv.org/abs/2402.07876}
\BIBentrySTDinterwordspacing

\bibitem[Dalal et~al.(2024)Dalal, Chiruvolu, Chaplot, and Salakhutdinov]{dalal2024plan}
M.~Dalal, T.~Chiruvolu, D.~Chaplot, and R.~Salakhutdinov, ``Plan-seq-learn: Language model guided rl for solving long horizon robotics tasks,'' \emph{arXiv preprint arXiv:2405.01534}, 2024.

\bibitem[Zhao et~al.(2024{\natexlab{b}})Zhao, Lee, and Hsu]{zhao2024large}
Z.~Zhao, W.~S. Lee, and D.~Hsu, ``Large language models as commonsense knowledge for large-scale task planning,'' \emph{Advances in Neural Information Processing Systems}, vol.~36, 2024.

\bibitem[Liu et~al.(2023{\natexlab{a}})Liu, Bahety, and Song]{liu2023reflect}
Z.~Liu, A.~Bahety, and S.~Song, ``Reflect: Summarizing robot experiences for failure explanation and correction,'' \emph{arXiv preprint arXiv:2306.15724}, 2023.

\bibitem[Lin et~al.(2023{\natexlab{a}})Lin, Cui, Hao, Xia, and Sadigh]{lin2023gesture}
L.-H. Lin, Y.~Cui, Y.~Hao, F.~Xia, and D.~Sadigh, ``Gesture-informed robot assistance via foundation models,'' in \emph{7th Annual Conference on Robot Learning}, 2023.

\bibitem[Wang et~al.(2024{\natexlab{c}})Wang, Wu, Li, Jiang, Shu, Shi, Hu, Ma, Liu, Wang, et~al.]{wang2024large}
J.~Wang, Z.~Wu, Y.~Li, H.~Jiang, P.~Shu, E.~Shi, H.~Hu, C.~Ma, Y.~Liu, X.~Wang \emph{et~al.}, ``Large language models for robotics: Opportunities, challenges, and perspectives,'' \emph{arXiv preprint arXiv:2401.04334}, 2024.

\bibitem[Li et~al.(2024{\natexlab{b}})Li, Zhao, Wang, Wang, Zhou, Srivastava, Gokmen, Lee, Li, Zhang, et~al.]{li2024embodied}
M.~Li, S.~Zhao, Q.~Wang, K.~Wang, Y.~Zhou, S.~Srivastava, C.~Gokmen, T.~Lee, L.~E. Li, R.~Zhang \emph{et~al.}, ``Embodied agent interface: Benchmarking llms for embodied decision making,'' \emph{arXiv preprint arXiv:2410.07166}, 2024.

\bibitem[Puig et~al.(2018)Puig, Ra, Boben, Li, Wang, Fidler, and Torralba]{puig2018virtualhome}
X.~Puig, K.~Ra, M.~Boben, J.~Li, T.~Wang, S.~Fidler, and A.~Torralba, ``Virtualhome: Simulating household activities via programs,'' in \emph{Proceedings of the IEEE conference on computer vision and pattern recognition}, 2018, pp. 8494--8502.

\bibitem[Srivastava et~al.(2022)Srivastava, Li, Lingelbach, Mart{\'\i}n-Mart{\'\i}n, Xia, Vainio, Lian, Gokmen, Buch, Liu, et~al.]{srivastava2022behavior}
S.~Srivastava, C.~Li, M.~Lingelbach, R.~Mart{\'\i}n-Mart{\'\i}n, F.~Xia, K.~E. Vainio, Z.~Lian, C.~Gokmen, S.~Buch, K.~Liu \emph{et~al.}, ``Behavior: Benchmark for everyday household activities in virtual, interactive, and ecological environments,'' in \emph{Conference on robot learning}.\hskip 1em plus 0.5em minus 0.4em\relax PMLR, 2022, pp. 477--490.

\bibitem[Mu et~al.(2024{\natexlab{a}})Mu, Zhang, Hu, Wang, Ding, Jin, Wang, Dai, Qiao, and Luo]{mu2024embodiedgpt}
Y.~Mu, Q.~Zhang, M.~Hu, W.~Wang, M.~Ding, J.~Jin, B.~Wang, J.~Dai, Y.~Qiao, and P.~Luo, ``Embodiedgpt: Vision-language pre-training via embodied chain of thought,'' \emph{Advances in Neural Information Processing Systems}, vol.~36, 2024.

\bibitem[Grauman et~al.(2022)Grauman, Westbury, Byrne, Chavis, Furnari, Girdhar, Hamburger, Jiang, Liu, Liu, et~al.]{grauman2022ego4d}
K.~Grauman, A.~Westbury, E.~Byrne, Z.~Chavis, A.~Furnari, R.~Girdhar, J.~Hamburger, H.~Jiang, M.~Liu, X.~Liu \emph{et~al.}, ``Ego4d: Around the world in 3,000 hours of egocentric video,'' in \emph{Proceedings of the IEEE/CVF Conference on Computer Vision and Pattern Recognition}, 2022, pp. 18\,995--19\,012.

\bibitem[Zhao et~al.(2023{\natexlab{a}})Zhao, Li, Weber, Hafez, and Wermter]{zhao2023chat}
X.~Zhao, M.~Li, C.~Weber, M.~B. Hafez, and S.~Wermter, ``Chat with the environment: Interactive multimodal perception using large language models,'' in \emph{2023 IEEE/RSJ International Conference on Intelligent Robots and Systems (IROS)}.\hskip 1em plus 0.5em minus 0.4em\relax IEEE, 2023, pp. 3590--3596.

\bibitem[Choi et~al.()Choi, Kim, Yoo, and Woo]{choiembodied}
W.~Choi, W.~K. Kim, M.~Yoo, and H.~Woo, ``Embodied cot distillation from llm to off-the-shelf agents,'' in \emph{Forty-first International Conference on Machine Learning}.

\bibitem[Huang et~al.(2024{\natexlab{a}})Huang, Xia, Shah, Driess, Zeng, Lu, Florence, Mordatch, Levine, Hausman, et~al.]{huang2024grounded}
W.~Huang, F.~Xia, D.~Shah, D.~Driess, A.~Zeng, Y.~Lu, P.~Florence, I.~Mordatch, S.~Levine, K.~Hausman \emph{et~al.}, ``Grounded decoding: Guiding text generation with grounded models for embodied agents,'' \emph{Advances in Neural Information Processing Systems}, vol.~36, 2024.

\bibitem[Huang et~al.(2022{\natexlab{b}})Huang, Xia, Xiao, Chan, Liang, Florence, Zeng, Tompson, Mordatch, Chebotar, et~al.]{huang2022inner}
W.~Huang, F.~Xia, T.~Xiao, H.~Chan, J.~Liang, P.~Florence, A.~Zeng, J.~Tompson, I.~Mordatch, Y.~Chebotar \emph{et~al.}, ``Inner monologue: Embodied reasoning through planning with language models,'' \emph{arXiv preprint arXiv:2207.05608}, 2022.

\bibitem[Guo et~al.(2024{\natexlab{a}})Guo, Xiang, Zhao, Zhu, Tomizuka, Ding, and Zhan]{guo2024phygrasp}
D.~Guo, Y.~Xiang, S.~Zhao, X.~Zhu, M.~Tomizuka, M.~Ding, and W.~Zhan, ``Phygrasp: Generalizing robotic grasping with physics-informed large multimodal models,'' \emph{arXiv preprint arXiv:2402.16836}, 2024.

\bibitem[Ren et~al.(2023)Ren, Govil, Yang, Narasimhan, and Majumdar]{ren2023leveraging}
A.~Z. Ren, B.~Govil, T.-Y. Yang, K.~R. Narasimhan, and A.~Majumdar, ``Leveraging language for accelerated learning of tool manipulation,'' in \emph{Conference on Robot Learning}.\hskip 1em plus 0.5em minus 0.4em\relax PMLR, 2023, pp. 1531--1541.

\bibitem[Rana et~al.(2023)Rana, Haviland, Garg, Abou-Chakra, Reid, and Suenderhauf]{rana2023sayplan}
K.~Rana, J.~Haviland, S.~Garg, J.~Abou-Chakra, I.~Reid, and N.~Suenderhauf, ``Sayplan: Grounding large language models using 3d scene graphs for scalable robot task planning,'' in \emph{7th Annual Conference on Robot Learning}, 2023.

\bibitem[Lv et~al.(2024)Lv, Li, Deng, Shao, Wang, and Nie]{lv2024robomp}
Q.~Lv, H.~Li, X.~Deng, R.~Shao, M.~Y. Wang, and L.~Nie, ``Robomp$^{2}$: A robotic multimodal perception-planning framework with multimodal large language models,'' \emph{arXiv preprint arXiv:2404.04929}, 2024.

\bibitem[Lewis et~al.(2020)Lewis, Perez, Piktus, Petroni, Karpukhin, Goyal, K{\"u}ttler, Lewis, Yih, Rockt{\"a}schel, et~al.]{lewis2020retrieval}
P.~Lewis, E.~Perez, A.~Piktus, F.~Petroni, V.~Karpukhin, N.~Goyal, H.~K{\"u}ttler, M.~Lewis, W.-t. Yih, T.~Rockt{\"a}schel \emph{et~al.}, ``Retrieval-augmented generation for knowledge-intensive nlp tasks,'' \emph{Advances in Neural Information Processing Systems}, vol.~33, pp. 9459--9474, 2020.

\bibitem[Liu et~al.(2021)Liu, Shen, Zhang, Dolan, Carin, and Chen]{liu2021makes}
J.~Liu, D.~Shen, Y.~Zhang, B.~Dolan, L.~Carin, and W.~Chen, ``What makes good in-context examples for gpt-$3 $?'' \emph{arXiv preprint arXiv:2101.06804}, 2021.

\bibitem[Lin et~al.(2023{\natexlab{b}})Lin, Agia, Migimatsu, Pavone, and Bohg]{lin2023text2motion}
K.~Lin, C.~Agia, T.~Migimatsu, M.~Pavone, and J.~Bohg, ``Text2motion: From natural language instructions to feasible plans,'' \emph{Autonomous Robots}, vol.~47, no.~8, pp. 1345--1365, 2023.

\bibitem[Agia et~al.(2023)Agia, Migimatsu, Wu, and Bohg]{agia2023stap}
C.~Agia, T.~Migimatsu, J.~Wu, and J.~Bohg, ``Stap: Sequencing task-agnostic policies,'' in \emph{2023 IEEE International Conference on Robotics and Automation (ICRA)}.\hskip 1em plus 0.5em minus 0.4em\relax IEEE, 2023, pp. 7951--7958.

\bibitem[Liang et~al.(2023)Liang, Huang, Xia, Xu, Hausman, Ichter, Florence, and Zeng]{liang2023code}
J.~Liang, W.~Huang, F.~Xia, P.~Xu, K.~Hausman, B.~Ichter, P.~Florence, and A.~Zeng, ``Code as policies: Language model programs for embodied control,'' in \emph{2023 IEEE International Conference on Robotics and Automation (ICRA)}.\hskip 1em plus 0.5em minus 0.4em\relax IEEE, 2023, pp. 9493--9500.

\bibitem[Zhao et~al.(2023{\natexlab{b}})Zhao, Chai, Wang, Boyi, Hao, Cao, Ye, and Wang]{zhao2023see}
Z.~Zhao, W.~Chai, X.~Wang, L.~Boyi, S.~Hao, S.~Cao, T.~Ye, and G.~Wang, ``See and think: Embodied agent in virtual environment,'' \emph{arXiv preprint arXiv:2311.15209}, 2023.

\bibitem[Chen et~al.(2024{\natexlab{c}})Chen, Mu, Yu, Wei, Wu, Yuan, Liang, Yang, Zhang, Shao, et~al.]{chen2024roboscript}
J.~Chen, Y.~Mu, Q.~Yu, T.~Wei, S.~Wu, Z.~Yuan, Z.~Liang, C.~Yang, K.~Zhang, W.~Shao \emph{et~al.}, ``Roboscript: Code generation for free-form manipulation tasks across real and simulation,'' \emph{arXiv preprint arXiv:2402.16117}, 2024.

\bibitem[Jin et~al.(2024)Jin, Li, A, Shi, Hao, Sun, Zhang, and Fang]{jin2024robotgpt}
Y.~Jin, D.~Li, Y.~A, J.~Shi, P.~Hao, F.~Sun, J.~Zhang, and B.~Fang, ``Robotgpt: Robot manipulation learning from chatgpt,'' \emph{IEEE Robotics and Automation Letters}, vol.~9, no.~3, pp. 2543--2550, 2024.

\bibitem[Huang et~al.(2023{\natexlab{a}})Huang, Jiang, Dong, Qiao, Gao, and Li]{huang2023instruct}
S.~Huang, Z.~Jiang, H.~Dong, Y.~Qiao, P.~Gao, and H.~Li, ``Instruct2act: Mapping multi-modality instructions to robotic actions with large language model,'' \emph{arXiv preprint arXiv:2305.11176}, 2023.

\bibitem[Zhi et~al.(2024)Zhi, Zhang, Han, Zhang, Li, Jiao, Jia, and Huang]{zhi2024closed}
P.~Zhi, Z.~Zhang, M.~Han, Z.~Zhang, Z.~Li, Z.~Jiao, B.~Jia, and S.~Huang, ``Closed-loop open-vocabulary mobile manipulation with gpt-4v,'' \emph{arXiv preprint arXiv:2404.10220}, 2024.

\bibitem[Singh et~al.(2023)Singh, Blukis, Mousavian, Goyal, Xu, Tremblay, Fox, Thomason, and Garg]{singh2023progprompt}
I.~Singh, V.~Blukis, A.~Mousavian, A.~Goyal, D.~Xu, J.~Tremblay, D.~Fox, J.~Thomason, and A.~Garg, ``Progprompt: Program generation for situated robot task planning using large language models,'' in \emph{Autonomous Robots}, vol.~47, 2023, pp. 999--1012.

\bibitem[Mu et~al.(2024{\natexlab{b}})Mu, Chen, Zhang, Chen, Yu, Ge, Chen, Liang, Hu, Tao, et~al.]{mu2024robocodex}
Y.~Mu, J.~Chen, Q.~Zhang, S.~Chen, Q.~Yu, C.~Ge, R.~Chen, Z.~Liang, M.~Hu, C.~Tao \emph{et~al.}, ``Robocodex: Multimodal code generation for robotic behavior synthesis,'' \emph{arXiv preprint arXiv:2402.16117}, 2024.

\bibitem[Wang et~al.(2024{\natexlab{d}})Wang, Ling, Yuan, Shridhar, Bao, Qin, Wang, Xu, and Wang]{wang2024gensim}
L.~Wang, Y.~Ling, Z.~Yuan, M.~Shridhar, C.~Bao, Y.~Qin, B.~Wang, H.~Xu, and X.~Wang, ``Gensim: Generating robotic simulation tasks via large language models,'' \emph{International Conference on Learning Representations}, 2024.

\bibitem[Yang et~al.(2024{\natexlab{b}})Yang, Dong, Liu, Li, Wang, Tan, Jiang, Kang, Zhang, Zhou, et~al.]{yang2024octopus}
J.~Yang, Y.~Dong, S.~Liu, B.~Li, Z.~Wang, H.~Tan, C.~Jiang, J.~Kang, Y.~Zhang, K.~Zhou \emph{et~al.}, ``Octopus: Embodied vision-language programmer from environmental feedback,'' \emph{arXiv preprint arXiv:2402.14623}, 2024.

\bibitem[Karli et~al.(2024)Karli, Chen, Antony, and Huang]{karli2024alchemist}
U.~B. Karli, J.-T. Chen, V.~N. Antony, and C.-M. Huang, ``Alchemist: Llm-aided end-user development of robot applications,'' pp. 361--370, 2024.

\bibitem[Huang et~al.(2024{\natexlab{b}})Huang, Wang, Li, Zhang, and Fei-Fei]{huang2024rekep}
W.~Huang, C.~Wang, Y.~Li, R.~Zhang, and L.~Fei-Fei, ``Rekep: Spatio-temporal reasoning of relational keypoint constraints for robotic manipulation,'' \emph{arXiv preprint arXiv:2409.01652}, 2024.

\bibitem[Huang et~al.(2023{\natexlab{b}})Huang, Wang, Zhang, Li, Wu, and Fei-Fei]{huang2023voxposer}
W.~Huang, C.~Wang, R.~Zhang, Y.~Li, J.~Wu, and L.~Fei-Fei, ``Voxposer: Composable 3d value maps for robotic manipulation with language models,'' \emph{arXiv preprint arXiv:2307.05973}, 2023.

\bibitem[Saccon et~al.(2024)Saccon, Tikna, De~Martini, Lamon, Palopoli, and Roveri]{saccon2024prolog}
E.~Saccon, A.~Tikna, D.~De~Martini, E.~Lamon, L.~Palopoli, and M.~Roveri, ``When prolog meets generative models: a new approach for managing knowledge and planning in robotic applications,'' in \emph{2024 IEEE International Conference on Robotics and Automation (ICRA)}.\hskip 1em plus 0.5em minus 0.4em\relax IEEE, 2024, pp. 17\,065--17\,071.

\bibitem[Jiang et~al.(2023)Jiang, Gupta, Zhang, Wang, Dou, Chen, Fei-Fei, Anandkumar, Zhu, and Fan]{jiang2023vima}
Y.~Jiang, A.~Gupta, Z.~Zhang, G.~Wang, Y.~Dou, Y.~Chen, L.~Fei-Fei, A.~Anandkumar, Y.~Zhu, and L.~Fan, ``Vima: General robot manipulation with multimodal prompts,'' in \emph{Fortieth International Conference on Machine Learning}, 2023.

\bibitem[Ze et~al.(2023)Ze, Yan, Wu, Macaluso, Ge, Ye, Hansen, Li, and Wang]{ze2023gn-factor}
Y.~Ze, G.~Yan, Y.-H. Wu, A.~Macaluso, Y.~Ge, J.~Ye, N.~Hansen, L.~E. Li, and X.~Wang, ``Gnfactor: Multi-task real robot learning with generalizable neural feature fields,'' in \emph{Conference on Robot Learning}.\hskip 1em plus 0.5em minus 0.4em\relax PMLR, 2023, pp. 284--301.

\bibitem[Karras et~al.(2020)Karras, Laine, Aittala, Hellsten, Lehtinen, and Aila]{karras2020StyleGAN}
T.~Karras, S.~Laine, M.~Aittala, J.~Hellsten, J.~Lehtinen, and T.~Aila, ``Analyzing and improving the image quality of stylegan,'' in \emph{Proceedings of the IEEE/CVF conference on computer vision and pattern recognition}, 2020, pp. 8110--8119.

\bibitem[Wang et~al.(2023{\natexlab{b}})Wang, Yuan, Chen, Zhang, Wang, and Zhang]{wang2023modelscopeT2V}
J.~Wang, H.~Yuan, D.~Chen, Y.~Zhang, X.~Wang, and S.~Zhang, ``Modelscope text-to-video technical report,'' \emph{arXiv preprint arXiv:2308.06571}, 2023.

\bibitem[Ho et~al.(2022)Ho, Salimans, Gritsenko, Chan, Norouzi, and Fleet]{ho2022video_diffusion_models}
J.~Ho, T.~Salimans, A.~Gritsenko, W.~Chan, M.~Norouzi, and D.~J. Fleet, ``Video diffusion models,'' \emph{arXiv preprint arXiv:2204.03458}, 2022.

\bibitem[Du et~al.(2024)Du, Yang, Dai, Dai, Nachum, Tenenbaum, Schuurmans, and Abbeel]{du2024UniPI}
Y.~Du, S.~Yang, B.~Dai, H.~Dai, O.~Nachum, J.~Tenenbaum, D.~Schuurmans, and P.~Abbeel, ``Learning universal policies via text-guided video generation,'' \emph{Advances in Neural Information Processing Systems}, vol.~36, 2024.

\bibitem[Hong et~al.(2024)Hong, Liu, Wu, Zhai, Chang, Li, Lin, Lin, Wang, Yang, et~al.]{hong2024SlowFast-VGen}
Y.~Hong, B.~Liu, M.~Wu, Y.~Zhai, K.-W. Chang, L.~Li, K.~Lin, C.-C. Lin, J.~Wang, Z.~Yang \emph{et~al.}, ``Slowfast-vgen: Slow-fast learning for action-driven long video generation,'' \emph{arXiv preprint arXiv:2410.23277}, 2024.

\bibitem[Cen et~al.(2024)Cen, Wu, Liu, Yin, Pei, Yang, Chen, Duan, and Zhang]{cen2024left_right}
J.~Cen, C.~Wu, X.~Liu, S.~Yin, Y.~Pei, J.~Yang, Q.~Chen, N.~Duan, and J.~Zhang, ``Using left and right brains together: Towards vision and language planning,'' \emph{arXiv preprint arXiv:2402.10534}, 2024.

\bibitem[Blattmann et~al.(2023)Blattmann, Dockhorn, Kulal, Mendelevitch, Kilian, Lorenz, Levi, English, Voleti, Letts, et~al.]{blattmann2023stablevideodiffusion}
A.~Blattmann, T.~Dockhorn, S.~Kulal, D.~Mendelevitch, M.~Kilian, D.~Lorenz, Y.~Levi, Z.~English, V.~Voleti, A.~Letts \emph{et~al.}, ``Stable video diffusion: Scaling latent video diffusion models to large datasets,'' \emph{arXiv preprint arXiv:2311.15127}, 2023.

\bibitem[Ajay et~al.(2024)Ajay, Han, Du, Li, Gupta, Jaakkola, Tenenbaum, Kaelbling, Srivastava, and Agrawal]{ajay2024HiP}
A.~Ajay, S.~Han, Y.~Du, S.~Li, A.~Gupta, T.~Jaakkola, J.~Tenenbaum, L.~Kaelbling, A.~Srivastava, and P.~Agrawal, ``Compositional foundation models for hierarchical planning,'' \emph{Advances in Neural Information Processing Systems}, vol.~36, 2024.

\bibitem[Bu et~al.(2024)Bu, Zeng, Chen, Yang, Zhou, Yan, Luo, Cui, Ma, and Li]{bu2024CLOVER}
Q.~Bu, J.~Zeng, L.~Chen, Y.~Yang, G.~Zhou, J.~Yan, P.~Luo, H.~Cui, Y.~Ma, and H.~Li, ``Closed-loop visuomotor control with generative expectation for robotic manipulation,'' \emph{arXiv preprint arXiv:2409.09016}, 2024.

\bibitem[Wu et~al.(2024{\natexlab{a}})Wu, Jing, Cheang, Chen, Xu, Li, Liu, Li, and Kong]{wu2023unleashing}
H.~Wu, Y.~Jing, C.~Cheang, G.~Chen, J.~Xu, X.~Li, M.~Liu, H.~Li, and T.~Kong, ``Unleashing large-scale video generative pre-training for visual robot manipulation,'' in \emph{International Conference on Learning Representations}, 2024.

\bibitem[Cheang et~al.(2024)Cheang, Chen, Jing, Kong, Li, Li, Liu, Wu, Xu, Yang, et~al.]{cheang2024gr-2}
C.-L. Cheang, G.~Chen, Y.~Jing, T.~Kong, H.~Li, Y.~Li, Y.~Liu, H.~Wu, J.~Xu, Y.~Yang \emph{et~al.}, ``Gr-2: A generative video-language-action model with web-scale knowledge for robot manipulation,'' \emph{arXiv preprint arXiv:2410.06158}, 2024.

\bibitem[Black et~al.(2023)Black, Nakamoto, Atreya, Walke, Finn, Kumar, and Levine]{black2023SuSIE}
K.~Black, M.~Nakamoto, P.~Atreya, H.~Walke, C.~Finn, A.~Kumar, and S.~Levine, ``Zero-shot robotic manipulation with pretrained image-editing diffusion models,'' \emph{arXiv preprint arXiv:2310.10639}, 2023.

\bibitem[Brooks et~al.(2023)Brooks, Holynski, and Efros]{brooks2023instructpix2pix}
T.~Brooks, A.~Holynski, and A.~A. Efros, ``Instructpix2pix: Learning to follow image editing instructions,'' in \emph{Proceedings of the IEEE/CVF Conference on Computer Vision and Pattern Recognition}, 2023, pp. 18\,392--18\,402.

\bibitem[Ni et~al.(2024)Ni, Hao, Wu, Kou, Liu, Zheng, Wang, and Zhuang]{ni2024CoTDiffusion}
F.~Ni, J.~Hao, S.~Wu, L.~Kou, J.~Liu, Y.~Zheng, B.~Wang, and Y.~Zhuang, ``Generate subgoal images before act: Unlocking the chain-of-thought reasoning in diffusion model for robot manipulation with multimodal prompts,'' in \emph{Proceedings of the IEEE/CVF Conference on Computer Vision and Pattern Recognition}, 2024, pp. 13\,991--14\,000.

\bibitem[Ren et~al.(2024)Ren, Zhang, Zheng, Li, Wen, Zhu, Ma, and Liang]{ren2024surferprogressivereasoningworld}
\BIBentryALTinterwordspacing
P.~Ren, K.~Zhang, H.~Zheng, Z.~Li, Y.~Wen, F.~Zhu, M.~Ma, and X.~Liang, ``Surfer: Progressive reasoning with world models for robotic manipulation,'' 2024. [Online]. Available: \url{https://arxiv.org/abs/2306.11335}
\BIBentrySTDinterwordspacing

\bibitem[Assran et~al.(2023)Assran, Duval, Misra, Bojanowski, Vincent, Rabbat, LeCun, and Ballas]{assran2023I-JEPA}
M.~Assran, Q.~Duval, I.~Misra, P.~Bojanowski, P.~Vincent, M.~Rabbat, Y.~LeCun, and N.~Ballas, ``Self-supervised learning from images with a joint-embedding predictive architecture,'' in \emph{Proceedings of the IEEE/CVF Conference on Computer Vision and Pattern Recognition}, 2023, pp. 15\,619--15\,629.

\bibitem[Pan et~al.(2023)Pan, Okorn, Zhang, Eisner, and Held]{pan2023tax-pose}
C.~Pan, B.~Okorn, H.~Zhang, B.~Eisner, and D.~Held, ``Tax-pose: Task-specific cross-pose estimation for robot manipulation,'' in \emph{Conference on Robot Learning}.\hskip 1em plus 0.5em minus 0.4em\relax PMLR, 2023, pp. 1783--1792.

\bibitem[Huang et~al.(2024{\natexlab{c}})Huang, Schmeckpeper, Wang, Biza, Qian, Liu, Jia, Platt, and Walters]{huang2024imagination-policy}
H.~Huang, K.~Schmeckpeper, D.~Wang, O.~Biza, Y.~Qian, H.~Liu, M.~Jia, R.~Platt, and R.~Walters, ``Imagination policy: Using generative point cloud models for learning manipulation policies,'' \emph{arXiv preprint arXiv:2406.11740}, 2024.

\bibitem[Dasgupta et~al.()Dasgupta, Gupta, Tuli, and Paul]{dasgupta_uncertainty-aware_2024}
\BIBentryALTinterwordspacing
S.~Dasgupta, A.~Gupta, S.~Tuli, and R.~Paul, ``Uncertainty-aware active learning of {NeRF}-based object models for robot manipulators using visual and re-orientation actions.'' [Online]. Available: \url{http://arxiv.org/abs/2404.01812}
\BIBentrySTDinterwordspacing

\bibitem[Shridhar et~al.(2022)Shridhar, Manuelli, and Fox]{shridhar2022peract}
M.~Shridhar, L.~Manuelli, and D.~Fox, ``Perceiver-actor: A multi-task transformer for robotic manipulation,'' in \emph{Proceedings of the 6th Conference on Robot Learning (CoRL)}, 2022.

\bibitem[Lu et~al.(2025)Lu, Zhang, Wang, Liu, Lu, and Tang]{lu2025manigaussian}
G.~Lu, S.~Zhang, Z.~Wang, C.~Liu, J.~Lu, and Y.~Tang, ``Manigaussian: Dynamic gaussian splatting for multi-task robotic manipulation,'' in \emph{European Conference on Computer Vision}.\hskip 1em plus 0.5em minus 0.4em\relax Springer, 2025, pp. 349--366.

\bibitem[LeCun(2022)]{lecun2022path}
Y.~LeCun, ``A path towards autonomous machine intelligence version 0.9. 2, 2022-06-27,'' \emph{Open Review}, vol.~62, no.~1, pp. 1--62, 2022.

\bibitem[Ma et~al.()Ma, Sodhani, Jayaraman, Bastani, Kumar, and Zhang]{ma_vip_2023}
\BIBentryALTinterwordspacing
Y.~J. Ma, S.~Sodhani, D.~Jayaraman, O.~Bastani, V.~Kumar, and A.~Zhang, ``{VIP}: Towards universal visual reward and representation via value-implicit pre-training.'' [Online]. Available: \url{http://arxiv.org/abs/2210.00030}
\BIBentrySTDinterwordspacing

\bibitem[Mu et~al.(2023)Mu, Yao, Ding, Luo, and Gan]{mu2023ec2}
Y.~Mu, S.~Yao, M.~Ding, P.~Luo, and C.~Gan, ``Ec2: Emergent communication for embodied control,'' in \emph{Proceedings of the IEEE/CVF Conference on Computer Vision and Pattern Recognition}, 2023, pp. 6704--6714.

\bibitem[Hafner et~al.({\natexlab{b}})Hafner, Lillicrap, Norouzi, and Ba]{hafnermastering}
D.~Hafner, T.~P. Lillicrap, M.~Norouzi, and J.~Ba, ``Mastering atari with discrete world models,'' in \emph{International Conference on Learning Representations}.

\bibitem[Hafner et~al.(2023)Hafner, Pasukonis, Ba, and Lillicrap]{hafner2023mastering}
D.~Hafner, J.~Pasukonis, J.~Ba, and T.~Lillicrap, ``Mastering diverse domains through world models,'' \emph{arXiv preprint arXiv:2301.04104}, 2023.

\bibitem[Seo et~al.(2022)Seo, Lee, James, and Abbeel]{seo2022reinforcement}
Y.~Seo, K.~Lee, S.~L. James, and P.~Abbeel, ``Reinforcement learning with action-free pre-training from videos,'' in \emph{International Conference on Machine Learning}.\hskip 1em plus 0.5em minus 0.4em\relax PMLR, 2022, pp. 19\,561--19\,579.

\bibitem[Zhang et~al.()Zhang, Kan, Shan, and Chen]{zhang_prelar_nodate}
L.~Zhang, M.~Kan, S.~Shan, and X.~Chen, ``{PreLAR}: World model pre-training with learnable action representation.''

\bibitem[Wang et~al.(2024{\natexlab{e}})Wang, Pan, Peng, Liu, Xu, Miao, Zhan, Tomizuka, Pollefeys, and Wang]{wang2024nerf}
G.~Wang, L.~Pan, S.~Peng, S.~Liu, C.~Xu, Y.~Miao, W.~Zhan, M.~Tomizuka, M.~Pollefeys, and H.~Wang, ``Nerf in robotics: A survey,'' \emph{arXiv preprint arXiv:2405.01333}, 2024.

\bibitem[Ryu et~al.(2024)Ryu, Kim, An, Chang, Seo, Kim, Kim, Hwang, Choi, and Horowitz]{ryu2024diffusion}
H.~Ryu, J.~Kim, H.~An, J.~Chang, J.~Seo, T.~Kim, Y.~Kim, C.~Hwang, J.~Choi, and R.~Horowitz, ``Diffusion-edfs: Bi-equivariant denoising generative modeling on se (3) for visual robotic manipulation,'' in \emph{Proceedings of the IEEE/CVF Conference on Computer Vision and Pattern Recognition}, 2024, pp. 18\,007--18\,018.

\bibitem[Guo et~al.(2024{\natexlab{b}})Guo, Hsiao, Liu, and Lee]{guo2024precise}
S.-W. Guo, T.-C. Hsiao, Y.-L. Liu, and C.-Y. Lee, ``Precise pick-and-place using score-based diffusion networks,'' \emph{arXiv preprint arXiv:2409.09725}, 2024.

\bibitem[Weng et~al.(2024)Weng, Lu, Kragic, and Lundell]{weng2024dexdiffuser}
Z.~Weng, H.~Lu, D.~Kragic, and J.~Lundell, ``Dexdiffuser: Generating dexterous grasps with diffusion models,'' \emph{arXiv preprint arXiv:2402.02989}, 2024.

\bibitem[Zhang et~al.(2024{\natexlab{b}})Zhang, Liu, Li, Yu, Geng, Ding, Chen, and Wang]{zhang2024dexgraspnet}
J.~Zhang, H.~Liu, D.~Li, X.~Yu, H.~Geng, Y.~Ding, J.~Chen, and H.~Wang, ``Dexgraspnet 2.0: Learning generative dexterous grasping in large-scale synthetic cluttered scenes,'' in \emph{8th Annual Conference on Robot Learning}, 2024.

\bibitem[Wang et~al.(2024{\natexlab{f}})Wang, Xing, Wei, Wu, and Zheng]{wang2024single}
Y.-K. Wang, C.~Xing, Y.-L. Wei, X.-M. Wu, and W.-S. Zheng, ``Single-view scene point cloud human grasp generation,'' in \emph{Proceedings of the IEEE/CVF Conference on Computer Vision and Pattern Recognition}, 2024, pp. 831--841.

\bibitem[Ye et~al.(2024)Ye, Gupta, Kitani, and Tulsiani]{ye2024g}
Y.~Ye, A.~Gupta, K.~Kitani, and S.~Tulsiani, ``G-hop: Generative hand-object prior for interaction reconstruction and grasp synthesis,'' in \emph{Proceedings of the IEEE/CVF Conference on Computer Vision and Pattern Recognition}, 2024, pp. 1911--1920.

\bibitem[Wu et~al.(2022{\natexlab{a}})Wu, Wang, Zhang, Zhang, Hilliges, Yu, and Tang]{wu2022saga}
Y.~Wu, J.~Wang, Y.~Zhang, S.~Zhang, O.~Hilliges, F.~Yu, and S.~Tang, ``Saga: Stochastic whole-body grasping with contact,'' in \emph{European Conference on Computer Vision}.\hskip 1em plus 0.5em minus 0.4em\relax Springer, 2022, pp. 257--274.

\bibitem[Jiang et~al.(2021)Jiang, Liu, Wang, and Wang]{jiang2021hand}
H.~Jiang, S.~Liu, J.~Wang, and X.~Wang, ``Hand-object contact consistency reasoning for human grasps generation,'' in \emph{Proceedings of the IEEE/CVF international conference on computer vision}, 2021, pp. 11\,107--11\,116.

\bibitem[Zhao et~al.(2024{\natexlab{c}})Zhao, Tsetserukou, and Liu]{zhao2024graingrasp}
F.~Zhao, D.~Tsetserukou, and Q.~Liu, ``Graingrasp: Dexterous grasp generation with fine-grained contact guidance,'' \emph{arXiv preprint arXiv:2405.09310}, 2024.

\bibitem[Wu et~al.(2022{\natexlab{b}})Wu, Guo, and Liu]{wu2022learning}
A.~Wu, M.~Guo, and C.~K. Liu, ``Learning diverse and physically feasible dexterous grasps with generative model and bilevel optimization,'' \emph{arXiv preprint arXiv:2207.00195}, 2022.

\bibitem[Bohg et~al.(2013)Bohg, Morales, Asfour, and Kragic]{bohg2013data}
J.~Bohg, A.~Morales, T.~Asfour, and D.~Kragic, ``Data-driven grasp synthesis—a survey,'' \emph{IEEE Transactions on robotics}, vol.~30, no.~2, pp. 289--309, 2013.

\bibitem[Du et~al.(2021)Du, Wang, Lian, and Zhao]{du2021vision}
G.~Du, K.~Wang, S.~Lian, and K.~Zhao, ``Vision-based robotic grasping from object localization, object pose estimation to grasp estimation for parallel grippers: a review,'' \emph{Artificial Intelligence Review}, vol.~54, no.~3, pp. 1677--1734, 2021.

\bibitem[Newbury et~al.(2023)Newbury, Gu, Chumbley, Mousavian, Eppner, Leitner, Bohg, Morales, Asfour, Kragic, et~al.]{newbury2023deep}
R.~Newbury, M.~Gu, L.~Chumbley, A.~Mousavian, C.~Eppner, J.~Leitner, J.~Bohg, A.~Morales, T.~Asfour, D.~Kragic \emph{et~al.}, ``Deep learning approaches to grasp synthesis: A review,'' \emph{IEEE Transactions on Robotics}, vol.~39, no.~5, pp. 3994--4015, 2023.

\bibitem[Qin et~al.(2020)Qin, Chen, Zhu, Song, Xu, and Su]{qin2020s4g}
Y.~Qin, R.~Chen, H.~Zhu, M.~Song, J.~Xu, and H.~Su, ``S4g: Amodal single-view single-shot se (3) grasp detection in cluttered scenes,'' in \emph{Conference on robot learning}.\hskip 1em plus 0.5em minus 0.4em\relax PMLR, 2020, pp. 53--65.

\bibitem[Mayer et~al.(2022)Mayer, Feng, Deng, Shi, Chen, and Knoll]{mayer2022ffhnet}
V.~Mayer, Q.~Feng, J.~Deng, Y.~Shi, Z.~Chen, and A.~Knoll, ``Ffhnet: Generating multi-fingered robotic grasps for unknown objects in real-time,'' in \emph{2022 International Conference on Robotics and Automation (ICRA)}.\hskip 1em plus 0.5em minus 0.4em\relax IEEE, 2022, pp. 762--769.

\bibitem[Murphy et~al.(2021)Murphy, Esteves, Jampani, Ramalingam, and Makadia]{murphy2021implicit}
K.~A. Murphy, C.~Esteves, V.~Jampani, S.~Ramalingam, and A.~Makadia, ``Implicit-pdf: Non-parametric representation of probability distributions on the rotation manifold,'' in \emph{International Conference on Machine Learning}.\hskip 1em plus 0.5em minus 0.4em\relax PMLR, 2021, pp. 7882--7893.

\bibitem[Kingma and Dhariwal(2018)]{kingma2018glow}
D.~P. Kingma and P.~Dhariwal, ``Glow: Generative flow with invertible 1x1 convolutions,'' \emph{Advances in neural information processing systems}, vol.~31, 2018.

\bibitem[Mirza(2014)]{mirza2014conditional}
M.~Mirza, ``Conditional generative adversarial nets,'' \emph{arXiv preprint arXiv:1411.1784}, 2014.

\bibitem[Yang et~al.(2025)Yang, Dong, Liu, Li, Wang, Tan, Jiang, Kang, Zhang, Zhou, et~al.]{yang2025octopus}
J.~Yang, Y.~Dong, S.~Liu, B.~Li, Z.~Wang, H.~Tan, C.~Jiang, J.~Kang, Y.~Zhang, K.~Zhou \emph{et~al.}, ``Octopus: Embodied vision-language programmer from environmental feedback,'' in \emph{European Conference on Computer Vision}.\hskip 1em plus 0.5em minus 0.4em\relax Springer, 2025, pp. 20--38.

\bibitem[Black et~al.(2024)Black, Brown, Driess, Esmail, Equi, Finn, Fusai, Groom, Hausman, Ichter, et~al.]{black2024pi_0}
K.~Black, N.~Brown, D.~Driess, A.~Esmail, M.~Equi, C.~Finn, N.~Fusai, L.~Groom, K.~Hausman, B.~Ichter \emph{et~al.}, ``$\pi_0 $: A vision-language-action flow model for general robot control,'' \emph{arXiv preprint arXiv:2410.24164}, 2024.

\bibitem[Zhao et~al.(2024{\natexlab{d}})Zhao, Tompson, Driess, Florence, Ghasemipour, Finn, and Wahid]{zhao2024alohaunleashedsimplerecipe}
\BIBentryALTinterwordspacing
T.~Z. Zhao, J.~Tompson, D.~Driess, P.~Florence, K.~Ghasemipour, C.~Finn, and A.~Wahid, ``Aloha unleashed: A simple recipe for robot dexterity,'' 2024. [Online]. Available: \url{https://arxiv.org/abs/2410.13126}
\BIBentrySTDinterwordspacing

\bibitem[Zhang et~al.(2024{\natexlab{c}})Zhang, Liu, Chang, Schramm, and Boularias]{zhang2024autoregressiveactionsequencelearning}
\BIBentryALTinterwordspacing
X.~Zhang, Y.~Liu, H.~Chang, L.~Schramm, and A.~Boularias, ``Autoregressive action sequence learning for robotic manipulation,'' 2024. [Online]. Available: \url{https://arxiv.org/abs/2410.03132}
\BIBentrySTDinterwordspacing

\bibitem[James et~al.(2020)James, Ma, Rovick~Arrojo, and Davison]{james2019rlbench}
S.~James, Z.~Ma, D.~Rovick~Arrojo, and A.~J. Davison, ``Rlbench: The robot learning benchmark \& learning environment,'' \emph{IEEE Robotics and Automation Letters}, 2020.

\bibitem[Mees et~al.(2022)Mees, Hermann, Rosete-Beas, and Burgard]{mees2022calvin}
O.~Mees, L.~Hermann, E.~Rosete-Beas, and W.~Burgard, ``Calvin: A benchmark for language-conditioned policy learning for long-horizon robot manipulation tasks,'' \emph{IEEE Robotics and Automation Letters (RA-L)}, vol.~7, no.~3, pp. 7327--7334, 2022.

\bibitem[Wang et~al.(2024{\natexlab{g}})Wang, Hart, Surovik, Kelestemur, Huang, Zhao, Yeatman, Wang, Walters, and Platt]{wang2024equivariantdiffusionpolicy}
\BIBentryALTinterwordspacing
D.~Wang, S.~Hart, D.~Surovik, T.~Kelestemur, H.~Huang, H.~Zhao, M.~Yeatman, J.~Wang, R.~Walters, and R.~Platt, ``Equivariant diffusion policy,'' 2024. [Online]. Available: \url{https://arxiv.org/abs/2407.01812}
\BIBentrySTDinterwordspacing

\bibitem[Yang et~al.(2024{\natexlab{c}})Yang, ang Cao, Deng, Antonova, Song, and Bohg]{yang2024equibotsim3equivariantdiffusionpolicy}
\BIBentryALTinterwordspacing
J.~Yang, Z.~ang Cao, C.~Deng, R.~Antonova, S.~Song, and J.~Bohg, ``Equibot: Sim(3)-equivariant diffusion policy for generalizable and data efficient learning,'' 2024. [Online]. Available: \url{https://arxiv.org/abs/2407.01479}
\BIBentrySTDinterwordspacing

\bibitem[Li et~al.(2024{\natexlab{c}})Li, Belagali, Shang, and Ryoo]{li2024crosswaydiffusionimprovingdiffusionbased}
\BIBentryALTinterwordspacing
X.~Li, V.~Belagali, J.~Shang, and M.~S. Ryoo, ``Crossway diffusion: Improving diffusion-based visuomotor policy via self-supervised learning,'' 2024. [Online]. Available: \url{https://arxiv.org/abs/2307.01849}
\BIBentrySTDinterwordspacing

\bibitem[Prasad et~al.(2024)Prasad, Lin, Wu, Zhou, and Bohg]{prasad2024consistency}
A.~Prasad, K.~Lin, J.~Wu, L.~Zhou, and J.~Bohg, ``Consistency policy: Accelerated visuomotor policies via consistency distillation,'' \emph{arXiv preprint arXiv:2405.07503}, 2024.

\bibitem[Yu et~al.(2024)Yu, Xu, Chen, Ren, and Pan]{yu2024bikckeyposeconditionedconsistencypolicy}
\BIBentryALTinterwordspacing
D.~Yu, H.~Xu, Y.~Chen, Y.~Ren, and J.~Pan, ``Bikc: Keypose-conditioned consistency policy for bimanual robotic manipulation,'' 2024. [Online]. Available: \url{https://arxiv.org/abs/2406.10093}
\BIBentrySTDinterwordspacing

\bibitem[Chen et~al.(2023{\natexlab{c}})Chen, Djolonga, Padlewski, Mustafa, Changpinyo, Wu, Ruiz, Goodman, Wang, Tay, et~al.]{chen2023pali}
X.~Chen, J.~Djolonga, P.~Padlewski, B.~Mustafa, S.~Changpinyo, J.~Wu, C.~R. Ruiz, S.~Goodman, X.~Wang, Y.~Tay \emph{et~al.}, ``Pali-x: On scaling up a multilingual vision and language model,'' \emph{arXiv preprint arXiv:2305.18565}, 2023.

\bibitem[Driess et~al.(2023)Driess, Xia, Sajjadi, Lynch, Chowdhery, Ichter, Wahid, Tompson, Vuong, Yu, et~al.]{driess2023palm}
D.~Driess, F.~Xia, M.~S. Sajjadi, C.~Lynch, A.~Chowdhery, B.~Ichter, A.~Wahid, J.~Tompson, Q.~Vuong, T.~Yu \emph{et~al.}, ``Palm-e: An embodied multimodal language model,'' \emph{arXiv preprint arXiv:2303.03378}, 2023.

\bibitem[Brohan et~al.(2023{\natexlab{b}})Brohan, Brown, Carbajal, Chebotar, Dabis, Finn, Gopalakrishnan, Hausman, Herzog, Hsu, Ibarz, Ichter, Irpan, Jackson, Jesmonth, Joshi, Julian, Kalashnikov, Kuang, Leal, Lee, Levine, Lu, Malla, Manjunath, Mordatch, Nachum, Parada, Peralta, Perez, Pertsch, Quiambao, Rao, Ryoo, Salazar, Sanketi, Sayed, Singh, Sontakke, Stone, Tan, Tran, Vanhoucke, Vega, Vuong, Xia, Xiao, Xu, Xu, Yu, and Zitkovich]{brohan2023rt1roboticstransformerrealworld}
\BIBentryALTinterwordspacing
A.~Brohan, N.~Brown, J.~Carbajal, Y.~Chebotar, J.~Dabis, C.~Finn, K.~Gopalakrishnan, K.~Hausman, A.~Herzog, J.~Hsu, J.~Ibarz, B.~Ichter, A.~Irpan, T.~Jackson, S.~Jesmonth, N.~J. Joshi, R.~Julian, D.~Kalashnikov, Y.~Kuang, I.~Leal, K.-H. Lee, S.~Levine, Y.~Lu, U.~Malla, D.~Manjunath, I.~Mordatch, O.~Nachum, C.~Parada, J.~Peralta, E.~Perez, K.~Pertsch, J.~Quiambao, K.~Rao, M.~Ryoo, G.~Salazar, P.~Sanketi, K.~Sayed, J.~Singh, S.~Sontakke, A.~Stone, C.~Tan, H.~Tran, V.~Vanhoucke, S.~Vega, Q.~Vuong, F.~Xia, T.~Xiao, P.~Xu, S.~Xu, T.~Yu, and B.~Zitkovich, ``Rt-1: Robotics transformer for real-world control at scale,'' 2023. [Online]. Available: \url{https://arxiv.org/abs/2212.06817}
\BIBentrySTDinterwordspacing

\bibitem[Goyal et~al.(2023)Goyal, Xu, Guo, Blukis, Chao, and Fox]{goyal2023rvtroboticviewtransformer}
\BIBentryALTinterwordspacing
A.~Goyal, J.~Xu, Y.~Guo, V.~Blukis, Y.-W. Chao, and D.~Fox, ``Rvt: Robotic view transformer for 3d object manipulation,'' 2023. [Online]. Available: \url{https://arxiv.org/abs/2306.14896}
\BIBentrySTDinterwordspacing

\bibitem[Goyal et~al.(2024)Goyal, Blukis, Xu, Guo, Chao, and Fox]{goyal2024rvt2learningprecisemanipulation}
\BIBentryALTinterwordspacing
A.~Goyal, V.~Blukis, J.~Xu, Y.~Guo, Y.-W. Chao, and D.~Fox, ``Rvt-2: Learning precise manipulation from few demonstrations,'' 2024. [Online]. Available: \url{https://arxiv.org/abs/2406.08545}
\BIBentrySTDinterwordspacing

\bibitem[Karamcheti et~al.(2024)Karamcheti, Nair, Balakrishna, Liang, Kollar, and Sadigh]{karamcheti2024prismatic}
S.~Karamcheti, S.~Nair, A.~Balakrishna, P.~Liang, T.~Kollar, and D.~Sadigh, ``Prismatic vlms: Investigating the design space of visually-conditioned language models,'' \emph{arXiv preprint arXiv:2402.07865}, 2024.

\bibitem[Oquab et~al.(2023)Oquab, Darcet, Moutakanni, Vo, Szafraniec, Khalidov, Fernandez, Haziza, Massa, El-Nouby, et~al.]{oquab2023dinov2}
M.~Oquab, T.~Darcet, T.~Moutakanni, H.~Vo, M.~Szafraniec, V.~Khalidov, P.~Fernandez, D.~Haziza, F.~Massa, A.~El-Nouby \emph{et~al.}, ``Dinov2: Learning robust visual features without supervision,'' \emph{arXiv preprint arXiv:2304.07193}, 2023.

\bibitem[Zhai et~al.(2023)Zhai, Mustafa, Kolesnikov, and Beyer]{zhai2023sigmoid}
X.~Zhai, B.~Mustafa, A.~Kolesnikov, and L.~Beyer, ``Sigmoid loss for language image pre-training,'' in \emph{Proceedings of the IEEE/CVF International Conference on Computer Vision}, 2023, pp. 11\,975--11\,986.

\bibitem[O'Neill et~al.(2023)O'Neill, Rehman, Gupta, Maddukuri, Gupta, Padalkar, Lee, Pooley, Gupta, Mandlekar, et~al.]{o2023open}
A.~O'Neill, A.~Rehman, A.~Gupta, A.~Maddukuri, A.~Gupta, A.~Padalkar, A.~Lee, A.~Pooley, A.~Gupta, A.~Mandlekar \emph{et~al.}, ``Open x-embodiment: Robotic learning datasets and rt-x models,'' \emph{arXiv preprint arXiv:2310.08864}, 2023.

\bibitem[Yu and Mooney(2023)]{yu2023usingdemonstrationslanguageinstructions}
\BIBentryALTinterwordspacing
A.~Yu and R.~J. Mooney, ``Using both demonstrations and language instructions to efficiently learn robotic tasks,'' 2023. [Online]. Available: \url{https://arxiv.org/abs/2210.04476}
\BIBentrySTDinterwordspacing

\bibitem[Chen et~al.(2023{\natexlab{d}})Chen, Garcia, Schmid, and Laptev]{chen2023polarnet3dpointclouds}
\BIBentryALTinterwordspacing
S.~Chen, R.~Garcia, C.~Schmid, and I.~Laptev, ``Polarnet: 3d point clouds for language-guided robotic manipulation,'' 2023. [Online]. Available: \url{https://arxiv.org/abs/2309.15596}
\BIBentrySTDinterwordspacing

\bibitem[Xu et~al.(2024)Xu, Xu, Xu, Chi, Wetzstein, Veloso, and Song]{xu2024flow}
\BIBentryALTinterwordspacing
M.~Xu, Z.~Xu, Y.~Xu, C.~Chi, G.~Wetzstein, M.~Veloso, and S.~Song, ``Flow as the cross-domain manipulation interface,'' in \emph{8th Annual Conference on Robot Learning}, 2024. [Online]. Available: \url{https://openreview.net/forum?id=cNI0ZkK1yC}
\BIBentrySTDinterwordspacing

\bibitem[Liang et~al.(2024)Liang, Liu, Ozguroglu, Sudhakar, Dave, Tokmakov, Song, and Vondrick]{liang2024dreamitate}
J.~Liang, R.~Liu, E.~Ozguroglu, S.~Sudhakar, A.~Dave, P.~Tokmakov, S.~Song, and C.~Vondrick, ``Dreamitate: Real-world visuomotor policy learning via video generation,'' 2024.

\bibitem[Bharadhwaj et~al.(2024{\natexlab{b}})Bharadhwaj, Dwibedi, Gupta, Tulsiani, Doersch, Xiao, Shah, Xia, Sadigh, and Kirmani]{bharadhwaj2024gen2act}
H.~Bharadhwaj, D.~Dwibedi, A.~Gupta, S.~Tulsiani, C.~Doersch, T.~Xiao, D.~Shah, F.~Xia, D.~Sadigh, and S.~Kirmani, ``Gen2act: Human video generation in novel scenarios enables generalizable robot manipulation,'' \emph{arXiv preprint arXiv:2409.16283}, 2024.

\bibitem[Garcia et~al.(2024)Garcia, Chen, and Schmid]{garcia2024generalizablevisionlanguageroboticmanipulation}
\BIBentryALTinterwordspacing
R.~Garcia, S.~Chen, and C.~Schmid, ``Towards generalizable vision-language robotic manipulation: A benchmark and llm-guided 3d policy,'' 2024. [Online]. Available: \url{https://arxiv.org/abs/2410.01345}
\BIBentrySTDinterwordspacing

\bibitem[Wang et~al.(2024{\natexlab{h}})Wang, Yin, Huang, Kelestemur, Wang, and Li]{wang2024gendp3dsemanticfields}
\BIBentryALTinterwordspacing
Y.~Wang, G.~Yin, B.~Huang, T.~Kelestemur, J.~Wang, and Y.~Li, ``Gendp: 3d semantic fields for category-level generalizable diffusion policy,'' 2024. [Online]. Available: \url{https://arxiv.org/abs/2410.17488}
\BIBentrySTDinterwordspacing

\bibitem[Ma et~al.(2024)Ma, Patidar, Haughton, and James]{ma2024hierarchical}
X.~Ma, S.~Patidar, I.~Haughton, and S.~James, ``Hierarchical diffusion policy for kinematics-aware multi-task robotic manipulation,'' \emph{CVPR}, 2024.

\bibitem[Szot et~al.(2024)Szot, Schwarzer, Agrawal, Mazoure, Talbott, Metcalf, Mackraz, Hjelm, and Toshev]{szot2024largelanguagemodelsgeneralizable}
\BIBentryALTinterwordspacing
A.~Szot, M.~Schwarzer, H.~Agrawal, B.~Mazoure, W.~Talbott, K.~Metcalf, N.~Mackraz, D.~Hjelm, and A.~Toshev, ``Large language models as generalizable policies for embodied tasks,'' 2024. [Online]. Available: \url{https://arxiv.org/abs/2310.17722}
\BIBentrySTDinterwordspacing

\bibitem[Liu et~al.(2024)Liu, Wu, Li, Tan, Chen, Wang, Xu, Su, and Zhu]{liu2024rdt1bdiffusionfoundationmodel}
\BIBentryALTinterwordspacing
S.~Liu, L.~Wu, B.~Li, H.~Tan, H.~Chen, Z.~Wang, K.~Xu, H.~Su, and J.~Zhu, ``Rdt-1b: a diffusion foundation model for bimanual manipulation,'' 2024. [Online]. Available: \url{https://arxiv.org/abs/2410.07864}
\BIBentrySTDinterwordspacing

\bibitem[Gervet et~al.(2023)Gervet, Xian, Gkanatsios, and Fragkiadaki]{gervet2023act3d}
T.~Gervet, Z.~Xian, N.~Gkanatsios, and K.~Fragkiadaki, ``Act3d: 3d feature field transformers for multi-task robotic manipulation,'' in \emph{7th Annual Conference on Robot Learning}, 2023.

\bibitem[James et~al.(2022)James, Wada, Laidlow, and Davison]{james2022coarse}
S.~James, K.~Wada, T.~Laidlow, and A.~J. Davison, ``Coarse-to-fine q-attention: Efficient learning for visual robotic manipulation via discretisation,'' in \emph{Proceedings of the IEEE/CVF Conference on Computer Vision and Pattern Recognition}, 2022, pp. 13\,739--13\,748.

\bibitem[Nguyen et~al.(2024)Nguyen, Vu, Ta, Huang, Vo, Le, and Nguyen]{nguyen2024robotic}
N.~Nguyen, M.~N. Vu, T.~D. Ta, B.~Huang, T.~Vo, N.~Le, and A.~Nguyen, ``Robotic-clip: Fine-tuning clip on action data for robotic applications,'' \emph{arXiv preprint arXiv:2409.17727}, 2024.

\bibitem[Zhang et~al.(2024{\natexlab{d}})Zhang, Wang, Cao, Yuan, Shan, Chen, and Gao]{zhang2024vlbiasbench}
J.~Zhang, S.~Wang, X.~Cao, Z.~Yuan, S.~Shan, X.~Chen, and W.~Gao, ``Vlbiasbench: A comprehensive benchmark for evaluating bias in large vision-language model,'' \emph{arXiv preprint arXiv:2406.14194}, 2024.

\bibitem[Makoviychuk et~al.(2021)Makoviychuk, Wawrzyniak, Guo, Lu, Storey, Macklin, Hoeller, Rudin, Allshire, Handa, et~al.]{makoviychuk2021isaac}
V.~Makoviychuk, L.~Wawrzyniak, Y.~Guo, M.~Lu, K.~Storey, M.~Macklin, D.~Hoeller, N.~Rudin, A.~Allshire, A.~Handa \emph{et~al.}, ``Isaac gym: High performance gpu-based physics simulation for robot learning,'' \emph{arXiv preprint arXiv:2108.10470}, 2021.

\bibitem[Salvato et~al.(2021)Salvato, Fenu, Medvet, and Pellegrino]{salvato2021crossing}
E.~Salvato, G.~Fenu, E.~Medvet, and F.~A. Pellegrino, ``Crossing the reality gap: A survey on sim-to-real transferability of robot controllers in reinforcement learning,'' \emph{IEEE Access}, vol.~9, pp. 153\,171--153\,187, 2021.

\bibitem[Pitkevich and Makarov(2024)]{pitkevich2024survey}
A.~Pitkevich and I.~Makarov, ``A survey on sim-to-real transfer methods for robotic manipulation,'' in \emph{2024 IEEE 22nd Jubilee International Symposium on Intelligent Systems and Informatics (SISY)}.\hskip 1em plus 0.5em minus 0.4em\relax IEEE, 2024, pp. 000\,259--000\,266.

\bibitem[Zhao et~al.(2020)Zhao, Queralta, and Westerlund]{zhao2020sim}
W.~Zhao, J.~P. Queralta, and T.~Westerlund, ``Sim-to-real transfer in deep reinforcement learning for robotics: a survey,'' in \emph{2020 IEEE symposium series on computational intelligence (SSCI)}.\hskip 1em plus 0.5em minus 0.4em\relax IEEE, 2020, pp. 737--744.

\bibitem[Peng et~al.(2018)Peng, Andrychowicz, Zaremba, and Abbeel]{peng2018sim}
X.~B. Peng, M.~Andrychowicz, W.~Zaremba, and P.~Abbeel, ``Sim-to-real transfer of robotic control with dynamics randomization,'' in \emph{2018 IEEE international conference on robotics and automation (ICRA)}.\hskip 1em plus 0.5em minus 0.4em\relax IEEE, 2018, pp. 3803--3810.

\bibitem[Garcia et~al.(2023)Garcia, Strudel, Chen, Arlaud, Laptev, and Schmid]{garcia2023robust}
R.~Garcia, R.~Strudel, S.~Chen, E.~Arlaud, I.~Laptev, and C.~Schmid, ``Robust visual sim-to-real transfer for robotic manipulation,'' in \emph{2023 IEEE/RSJ International Conference on Intelligent Robots and Systems (IROS)}.\hskip 1em plus 0.5em minus 0.4em\relax ieee, 2023, pp. 992--999.

\bibitem[Huber et~al.(2024)Huber, H{\'e}l{\'e}non, Watrelot, Amar, and Doncieux]{huber2024domain}
J.~Huber, F.~H{\'e}l{\'e}non, H.~Watrelot, F.~B. Amar, and S.~Doncieux, ``Domain randomization for sim2real transfer of automatically generated grasping datasets,'' in \emph{2024 IEEE International Conference on Robotics and Automation (ICRA)}.\hskip 1em plus 0.5em minus 0.4em\relax IEEE, 2024, pp. 4112--4118.

\bibitem[Liu et~al.(2023{\natexlab{b}})Liu, Chen, and Wu]{liu2023digital}
D.~Liu, Y.~Chen, and Z.~Wu, ``Digital twin (dt)-cyclegan: Enabling zero-shot sim-to-real transfer of visual grasping models,'' \emph{IEEE Robotics and Automation Letters}, vol.~8, no.~5, pp. 2421--2428, 2023.

\bibitem[Chen et~al.(2022{\natexlab{b}})Chen, Cao, James, Li, Liu, Abbeel, and Dou]{chen2022sim}
K.~Chen, R.~Cao, S.~James, Y.~Li, Y.-H. Liu, P.~Abbeel, and Q.~Dou, ``Sim-to-real 6d object pose estimation via iterative self-training for robotic bin picking,'' in \emph{European Conference on Computer Vision}.\hskip 1em plus 0.5em minus 0.4em\relax Springer, 2022, pp. 533--550.

\bibitem[He et~al.(2024)He, Wu, Bai, Lai, Wang, Pan, Hu, and Zhang]{he2024bridging}
H.~He, P.~Wu, C.~Bai, H.~Lai, L.~Wang, L.~Pan, X.~Hu, and W.~Zhang, ``Bridging the sim-to-real gap from the information bottleneck perspective,'' in \emph{8th Annual Conference on Robot Learning}, 2024.

\bibitem[Chen et~al.(2020{\natexlab{b}})Chen, Zhou, Koltun, and Kr{\"a}henb{\"u}hl]{chen2020learning}
D.~Chen, B.~Zhou, V.~Koltun, and P.~Kr{\"a}henb{\"u}hl, ``Learning by cheating,'' in \emph{Conference on Robot Learning}.\hskip 1em plus 0.5em minus 0.4em\relax PMLR, 2020, pp. 66--75.

\bibitem[Nguyen et~al.(2022)Nguyen, Baisero, Wang, Amato, and Platt]{nguyen2022leveraging}
H.~Nguyen, A.~Baisero, D.~Wang, C.~Amato, and R.~Platt, ``Leveraging fully observable policies for learning under partial observability,'' \emph{arXiv preprint arXiv:2211.01991}, 2022.

\bibitem[Kang et~al.(2024)Kang, Yue, Lu, Lin, Zhao, Wang, Huang, and Feng]{kang2024far}
B.~Kang, Y.~Yue, R.~Lu, Z.~Lin, Y.~Zhao, K.~Wang, G.~Huang, and J.~Feng, ``How far is video generation from world model: A physical law perspective,'' \emph{arXiv preprint arXiv:2411.02385}, 2024.

\bibitem[Bansal et~al.(2024)Bansal, Lin, Xie, Zong, Yarom, Bitton, Jiang, Sun, Chang, and Grover]{bansal2024VideoPhy}
H.~Bansal, Z.~Lin, T.~Xie, Z.~Zong, M.~Yarom, Y.~Bitton, C.~Jiang, Y.~Sun, K.-W. Chang, and A.~Grover, ``Videophy: Evaluating physical commonsense for video generation,'' \emph{arXiv preprint arXiv:2406.03520}, 2024.

\bibitem[Wang et~al.(2024{\natexlab{i}})Wang, Li, Hao, Song, and Hu]{wang2024VAMP}
Z.~Wang, S.~Li, L.~Hao, B.~Song, and X.~Hu, ``What you see is what matters: A novel visual and physics-based metric for evaluating video generation quality,'' \emph{arXiv preprint arXiv:2411.13609}, 2024.

\bibitem[Wu et~al.(2024{\natexlab{b}})Wu, Xu, Chen, Hua, Luo, and Wang]{wu2024Pastnet}
H.~Wu, F.~Xu, C.~Chen, X.-S. Hua, X.~Luo, and H.~Wang, ``Pastnet: Introducing physical inductive biases for spatio-temporal video prediction,'' in \emph{Proceedings of the 32nd ACM International Conference on Multimedia}, 2024, pp. 2917--2926.

\end{thebibliography}
}

\end{document}